\documentclass{article} 
\usepackage{iclr2027_conference,times}

\usepackage{amsmath,amsfonts,bm}

\def\eqref#1{equation~\ref{#1}}

\def\1{\bm{1}}

\def\vx{{\bm{x}}}

\DeclareMathAlphabet{\mathsfit}{\encodingdefault}{\sfdefault}{m}{sl}
\SetMathAlphabet{\mathsfit}{bold}{\encodingdefault}{\sfdefault}{bx}{n}

\usepackage{hyperref}
\usepackage{url}

\title{CIDER-FM: Foundation Models for Causal Inference from Diverse Experimental Regimes}

\author{%
\makebox[\dimexpr\textwidth-2\tabcolsep\relax][c]{%
\begin{tabular}[t]{@{}c@{}}
{\bfseries
Yuche Gao\textsuperscript{1} \quad
Arik Reuter\textsuperscript{1,2} \quad
Siyuan Guo\textsuperscript{3}}
\\[0.25em]
{\bfseries
Anish Dhir\textsuperscript{4} \quad
Bernhard Sch{\"o}lkopf\textsuperscript{5,2} \quad
Adrian Weller\textsuperscript{6,1}}
\\[0.75em]
{\normalfont\small
\textsuperscript{1}University of Cambridge, Cambridge, United Kingdom}
\\
{\normalfont\small
\textsuperscript{2}Max Planck Institute for Intelligent Systems,
T{\"u}bingen, Germany}
\\
{\normalfont\small
\textsuperscript{3}Prior Labs}
\\
{\normalfont\small
\textsuperscript{4}Gatsby Computational Neuroscience Unit,
University College London, London, United Kingdom}
\\
{\normalfont\small
\textsuperscript{5}ELLIS Institute, T{\"u}bingen, Germany}
\\
{\normalfont\small
\textsuperscript{6}The Alan Turing Institute, London, United Kingdom}
\end{tabular}%
}%
}

\usepackage[utf8]{inputenc} 
\usepackage[T1]{fontenc}    
\usepackage{hyperref}       
\usepackage{url}            
\usepackage{amsmath}
\usepackage{amsfonts}
\usepackage{amssymb}
\usepackage{nicefrac}       
\usepackage{graphicx}
\usepackage{microtype}      
\usepackage{xcolor}         

\usepackage{enumitem}

\usepackage{booktabs}
\usepackage{tabularx}
\usepackage{array}
\usepackage{multirow}

\usepackage{mathtools}
\usepackage{amsthm}

\usepackage{wrapfig2} 
\usepackage{subcaption}
\DeclareRobustCommand{\dataset}{\mathcal{D}}
\DeclareRobustCommand{\regimedata}[1]{\mathcal{D}_{#1}}
\DeclareRobustCommand{\regimedatasets}{\mathcal{D}_1,\ldots,\mathcal{D}_R}
\DeclareRobustCommand{\obsdata}{\mathcal{D}_{\mathrm{obs}}}
\DeclareRobustCommand{\intdata}{\mathcal{D}_{\mathrm{int}}}

\DeclareRobustCommand{\doop}[1]{%
  \operatorname{do}\!\left(#1\right)}

\usepackage{algorithm}
\usepackage{algpseudocode}

\usepackage{tikz}
\usetikzlibrary{arrows.meta,positioning}

\definecolor{treatorange}{RGB}{232,126,48}
\definecolor{covred}{RGB}{183,45,38}
\definecolor{outblue}{RGB}{36,64,174}
\definecolor{unobsgray}{RGB}{160,160,160}
\definecolor{surrogateyellow}{RGB}{0,191,0}

\theoremstyle{definition}
\newtheorem{assumption}{Assumption}

\usepackage[nameinlink,capitalize]{cleveref}
\crefname{assumption}{Assumption}{Assumptions}
\Crefname{assumption}{Assumption}{Assumptions}

\hypersetup{
  colorlinks=true,
  linkcolor=blue!50!black,
  citecolor=blue!50!black,
  urlcolor=blue!50!black
}

\usepackage[most]{tcolorbox}
\newtcolorbox{contributionbox}{
  enhanced,
  breakable,
  colback=black!4,
  colframe=black!15,
  boxrule=0.4pt,
  arc=1mm,
  left=5pt,
  right=5pt,
  top=5pt,
  bottom=10pt,
  before skip=10pt,
  after skip=10pt
}
\iclrfinalcopy 

\begin{document}

\maketitle
\lhead{Preprint.}

\begin{abstract}
Causal foundation models (CFMs) amortise causal inference over priors of synthetic structural causal models (SCMs), predicting the effect of an experiment on a specific variable. However, observational data alone may leave multiple causal models compatible with available evidence, while experimental data with interventions on exactly the variable of interest might be unavailable. This work studies CFMs as a method to combine finite observational and surrogate-interventional datasets in order to predict a target conditional interventional distribution (CID) more accurately than with observational data alone. We first formalise the conceptual benefits of surrogate experiments. Building on this analysis, we introduce \textsc{Foundation Models for Causal Inference from Diverse Experimental Regimes} (\emph{CIDER-FM}), a causal foundation model that uses an intervention-aware representation and hierarchical three-axis attention to exchange information across variables, samples, and experimental regimes. We evaluate CIDER-FM against a wide range of baselines across
diverse synthetic graph and mechanism families, as well as
on both simulated and real-world data from Causal Chambers.
Our results demonstrate strong CID prediction performance
and show that incorporating experimental context can improve
predictions over observational data alone.
\end{abstract}

\section{Introduction}

Causal questions are central to scientific inquiry and decision making. When a variable $T$ can be experimentally manipulated and an outcome $Y$ measured, the causal effect of $T$ on $Y$ can be estimated directly. When such a targeted experiment is infeasible and, for instance, only observational data can be collected, multiple causal models may explain the same observational distribution yet lead to different causal effects \citep{pearl2009causality,peters2017elements}. Therefore, traditional causal inference pairs explicit structural assumptions with bespoke estimators tailored to a particular identification strategy. However, explicit structural assumptions, such as knowledge of the entire set of causal relationships between all variables in a system, might not be justified, especially in complex scientific setups. Even worse, having to make an assumption without the necessary grounds can lead to wrong yet confident predictions.

\begin{figure*}[t]
  \centering
  \includegraphics[width=0.8\textwidth]{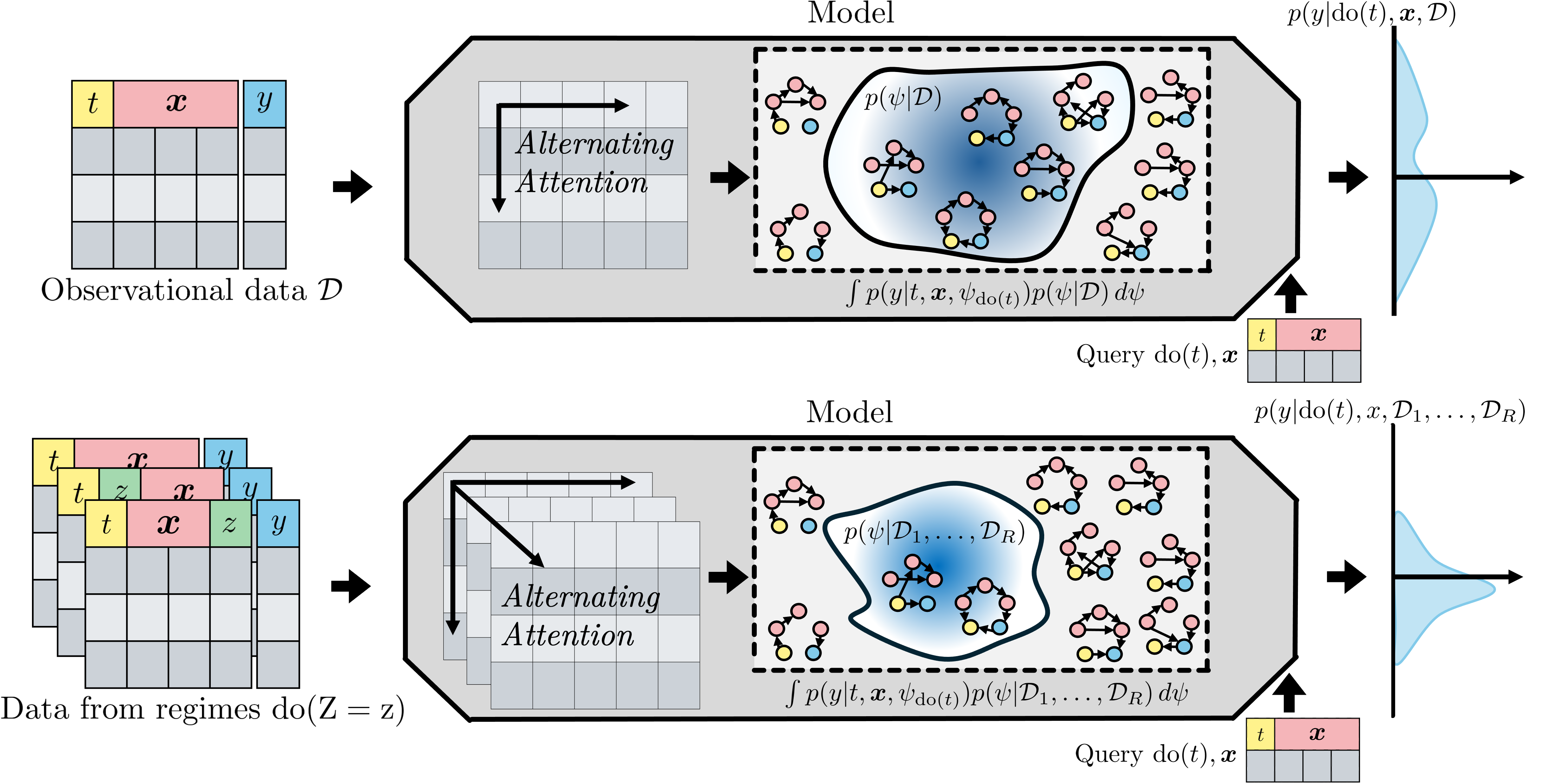}
  \caption[Conceptual comparison of observational data and multi-regime interventional data fusion.] {\textbf{Conceptual comparison: }observational data (top) and multi-regime experimental data  (bottom) in Causal Foundation Models (CFMs). With observational data
alone, multiple compatible SCMs may imply different CIDs, producing a broad or
multimodal posterior predictive distribution. Adding data from different experimental regimes can
exclude or downweight incompatible SCMs and thereby refine the predictive distribution. Compared to existing TFMs, the stacked tables in the
lower panel make explicit a third structural axis, namely the different experimental
regime, alongside variables and samples along which CIDER-FM performs the attention operation.}
  \label{fig:contributions}
\end{figure*}

Recently proposed Causal foundation models (CFMs) offer a complementary approach: pretrained on a distribution of synthetic SCMs, a single model takes as input observational data and amortises causal discovery followed by  causal inference using a prior over causal structures \citep{robertson2026pfn,dhir2026estimating}. A CFM approximates the posterior
predictive conditional interventional distribution (CID)
\begin{equation}
\begin{aligned}
&p\!\left(
    y
    \mid
    \doop{T=t},
    \mathbf{x},
    \mathcal{D}
\right) =
\int
p\!\left(
    y
    \mid
    \mathbf{x},
    \psi_{\mathrm{do}(T=t)}
\right)
p\!\left(
    \psi
    \mid
    \mathcal{D}, \vx
\right)
\,\mathrm{d}\psi ,
\end{aligned}
\label{eq:intro-posterior-predictive}
\end{equation}
where \(\psi\) denotes an SCM and
$\mathcal{D}$ the available data. This
posterior-predictive view allows uncertainty about both the graph and its
mechanisms (from the implicit causal-discovery step) to be propagated into the causal prediction, rather than replaced immediately by a single
fixed causal model \citep{dhir2024bivariate, dhir2026estimating, robertson2026pfn}.
However, the cost of this approach is that, since existing CFMs only use observational data, a broad SCM prior leaves several
causal explanations unresolved \citep{robertson2026pfn, dhir2026estimating}. Mathematically, this corresponds to an uncertain $p(\psi|\mathcal{D}_{obs}, \vx)$. Since the final CID integrates over the posterior belief over SCMs, using only an observational dataset in $p(\psi|\mathcal{D}_{obs}, \vx)$ may lead the CID failing to concentrate and instead remaining diffuse or multimodal (the first panel in \cref{fig:contributions}).

\begin{figure*}[t]
  \centering
  \includegraphics[width=0.8\textwidth]{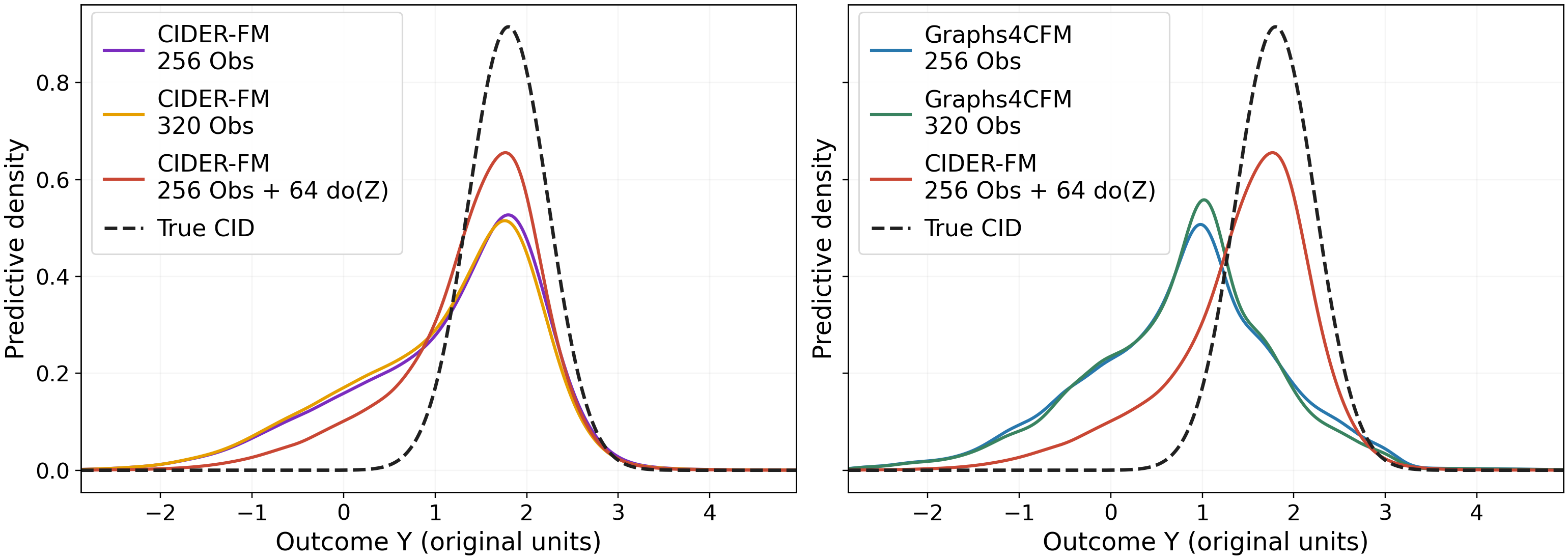}
  \caption{\textbf{Case study:} surrogate experiments sharpen conditional interventional predictions.
Model predictive densities are visualised using Gaussian KDEs.
Left: the same CIDER-FM checkpoint receives 256 observational samples, 320 observational samples, or 256 observational samples plus 64 randomized $\mathrm{do}(Z)$ samples.
Right: comparison with observational-only Graphs4CFM \citep{reuter2026use} without graph input.
Details of the task construction and quantitative results are in \cref{app:cid_concentration}.}
  \label{fig:case_converge}
\end{figure*}

This work studies \emph{interventional data as a complementary source of
causal information in CFMs} to overcome this issue by reducing or even fully collapsing the posterior over SCMs compatible with a scenario, over which we need to amortise. More specifically, we replace $\mathcal{D}_{obs}$ by multiple datasets $\regimedatasets$ that can be observational or interventional data from different experiments. In many scientific settings, observational data
coexist with datasets collected under controlled interventions. 

Direct experimentation on the treatment variable of interest \(T\)
may be unavailable, while surrogate experiments \(\doop{Z=z}\)
on other variables are feasible. This motivates exploiting
multi-environment data from diverse experimental regimes,
including observational data, data from surrogate experiments,
and, when available, interventional data on the treatment
variable \(T\), to constrain the set of SCMs compatible with the available evidence \citep{hauser2012characterization,kivva2023identifiability}, as illustrated in the second panel of
\cref{fig:contributions} and the case study in \cref{fig:case_converge}.
Following the prior-data fitted network (PFN) paradigm of
synthetic pretraining and in-context learning
\citep{hollmann2025accurate}, we introduce
\textsc{Foundation Models for Causal Inference from Diverse
Experimental Regimes} (\emph{CIDER-FM}).

\begin{contributionbox}
\noindent\textbf{Our contributions.}
\begin{enumerate}
    \setlength{\itemsep}{4pt}

    \item \textbf{Theoretical analysis.}
    We formalise how surrogate experiments can enable point
    identification or narrow identified sets, and provide
    complementary finite-sample and asymptotic analyses of
    how they can improve prediction accuracy.

    \item \textbf{Model architecture.}
    CIDER-FM uses intervention-aware representations of continuous,
    row-specific randomised treatment assignments \(\doop{X=x_i}\)
    and applies hierarchical three-axis attention across variables,
    samples within regimes, and experimental regimes to enable
    in-context causal prediction from multi-environment data.
    
    \item \textbf{Synthetic and real-world validation.}
    We demonstrate CIDER-FM's strong predictive performance across
    diverse synthetic graph and mechanism families compared with
    observation-only baselines under matched context budgets.
    On Causal Chambers, the same model supports simulator-guided
    experimental design and achieves strong causal prediction
    performance on the resulting real-world experiments.
\end{enumerate}
\end{contributionbox}

\section{Background and related work}

\paragraph{Structural causal models and interventions}
A structural causal model (SCM) \(\psi=(G,\mathcal M)\)
consists of a causal graph \(G\) and mechanisms \(\mathcal M\)
specifying how each variable is generated from its parents
and exogenous noise, together with the joint noise
distribution \citep{pearl2009causality,peters2017elements}.
A hard intervention \(\doop{T=t}\) replaces the structural
assignment for \(T\) by \(T:=t\), removing incoming edges to
\(T\) in \(G\) while leaving the other mechanisms unchanged.
Within a specified model class, the available population
distributions may be compatible with multiple SCMs.
The graph is identifiable if the set of compatible graphs
is a singleton. A causal query is identifiable if all these
compatible SCMs agree on the target quantity, such as
\(p_\psi(y\mid\doop{T=t},x)\).

\paragraph{Traditional causal methods.} Established causal methods span several families. Graphical methods use
adjustment criteria and do-calculus to derive causal estimands under an assumed
graph
\citep{pearl1995causal,pearl2009causality,shpitser2006identification};
outcome-regression and propensity-score methods estimate effects through
models of the outcome or treatment assignment
\citep{robins1986new,rosenbaum1983central}. Although these approaches
can provide guarantees under their respective conditions, their
application requires problem-specific assumptions and typically manual
choices of adjustment sets and estimation models.

\paragraph{Tabular Foundation Models.} An alternative to fitting a problem-specific estimator is to
amortise inference across a distribution of tasks. Tabular foundation models, such as TabPFN \citep{hollmann2023tabpfn, hollmann2025accurate}
, are amortised inference methods in the form of neural processes \citep{garnelo2018conditional, garnelo2018neural} that have revolutionised the domain of tabular machine learning, setting new benchmark records \citep{erickson2026tabarena}. They work by pretraining
on synthetic datasets and adapting to new tasks through in-context learning
(ICL), without task-specific parameter updates
\citep{muller2022transformers,hollmann2023tabpfn,hollmann2025accurate}. CIDER-FM, the model we propose, operates in the same synthetic-data training and in-context-learning paradigm as TFMs.

\paragraph{Causal Foundation Models.} Causal foundation models (CFMs) extend the ideas behind synthetically trained TFMs to causal queries such as conditional interventional distributions. Existing CFMs
broadly follow two strategies. Identifiable-prior CFMs restrict their
pretraining priors to structural assumptions in which the target query is identifiable from
observational data under specific causal assumptions. For instance, \citet{balazadeh2026causalpfn} propose CausalPFN, a CFM that achieves excellent predictive performance on causal effect estimation with binary treatments under unconfoundedness, while \citet{ma2025foundation} propose priors to train a CFM for varying causal scenarios, such as backdoor-adjustment or front-door adjustment. In contrast, posterior-averaging CFMs learn the posterior predictive distribution induced by a broader prior
over SCMs, retaining uncertainty when observationally compatible SCMs imply
different interventional distributions
\citep{robertson2026pfn,dhir2026estimating}. Both strategies primarily use
observational datasets as context, leaving open how to incorporate data from diverse experimental regimes. 

Recently, conditioning CFMs on partial graph information has been proposed as a way to bridge the gap between identifiable-prior and posterior-averaging CFMs \citep{reuter2026use}. Reliable causal graphs, however, are often unavailable in complex scientific setups. Accurately estimating them from data is challenging and treating them as trusted inputs can propagate graph-discovery errors into the final prediction. In this paper, we instead leverage data from multiple environments, not requiring any graphical assumptions.

\section{Benefits of surrogate experiments}
\label{sec:theoretical-benefits}

Causal inference from observational data is arguably a well-known and well-studied topic \citep{pearl2009causality, rubin1974estimating}, and having access to experimental data, where the experiment has been performed on precisely the variable $T$ whose effect is to be examined, remains mostly an estimation challenge, not a causal identification problem. In this section, we discuss the case of data from multiple surrogate experiments on different variables, which CIDER-FM takes as input. 

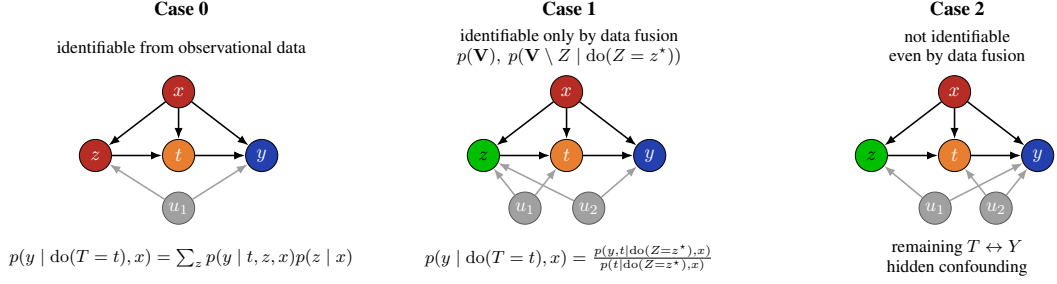
\begin{figure*}[t]
    \centering
    \resizebox{1.0\textwidth}{!}{%
        \begin{tikzpicture}[
    scale=1, transform shape,
    cov/.style={circle,draw=black,fill=covred,minimum size=5.8mm,inner sep=0pt,
        text=white,font=\bfseries\itshape\normalsize},
    trt/.style={circle,draw=black,fill=treatorange,minimum size=5.8mm,inner sep=0pt,
        text=white,font=\bfseries\itshape\normalsize},
    outcome/.style={circle,draw=black,fill=outblue,minimum size=5.8mm,inner sep=0pt,
        text=white,font=\bfseries\itshape\normalsize},
    unobs/.style={circle,draw=gray!70!black,fill=unobsgray,minimum size=5.8mm,inner sep=0pt,
        text=white,font=\bfseries\itshape\normalsize},
    sur/.style={circle,draw=black,fill=surrogateyellow,minimum size=5.8mm,inner sep=0pt,
        text=black,font=\bfseries\itshape\normalsize},
    arr/.style={-{Latex[length=2mm]},thick},
    harr/.style={-{Latex[length=2mm]},thick,gray!75},
    title/.style={font=\bfseries\normalsize,align=center},
    desc/.style={font=\small,align=center},
    formula/.style={font=\small,align=center}
]

\begin{scope}[xshift=0cm]
    \node[title] at (1.55,2.65) {Case 0};
    \node[desc] at (1.55,1.95) {identifiable from observational data};

    \node[cov] (z1) at (0,0) {$z$};
    \node[trt] (t1) at (1.5,0) {$t$};
    \node[outcome] (y1) at (3,0) {$y$};
    \node[cov] (x1) at (1.5,1.15) {$x$};
    \node[unobs] (u11) at (1.5,-0.95) {$u_1$};

    \draw[arr] (z1) -- (t1);
    \draw[arr] (t1) -- (y1);

    \draw[arr] (x1) -- (z1);
    \draw[arr] (x1) -- (t1);
    \draw[arr] (x1) -- (y1);

    \draw[harr] (u11) -- (z1);
    \draw[harr] (u11) -- (y1);

    \node[formula] at (1.55,-1.85)
    {$p(y\mid \mathrm{do}(T=t),x)=\sum_z p(y\mid t,z,x)p(z\mid x)$};
\end{scope}

\begin{scope}[xshift=7.0cm]
    \node[title] at (1.55,2.65) {Case 1};
    \node[desc] at (1.55,1.95) {identifiable only by data fusion\\$p(\mathbf{V}),\ p(\mathbf{V}\setminus Z\mid \mathrm{do}(Z = z^\star))$};

    \node[sur] (z2) at (0,0) {$z$};
    \node[trt] (t2) at (1.5,0) {$t$};
    \node[outcome] (y2) at (3,0) {$y$};
    \node[cov] (x2) at (1.5,1.15) {$x$};
    \node[unobs] (u12) at (0.75,-0.95) {$u_1$};
    \node[unobs] (u22) at (1.90,-0.95) {$u_2$};

    \draw[arr] (z2) -- (t2);
    \draw[arr] (t2) -- (y2);

    \draw[arr] (x2) -- (z2);
    \draw[arr] (x2) -- (t2);
    \draw[arr] (x2) -- (y2);

    \draw[harr] (u12) -- (z2);
    \draw[harr] (u12) -- (t2);
    \draw[harr] (u22) -- (z2);
    \draw[harr] (u22) -- (y2);

    \node[formula] at (1.55,-1.85)
    {$p(y\mid \mathrm{do}(T=t),x)=\frac{p(y,t\mid \mathrm{do}(Z=z^\star),x)}{p(t\mid \mathrm{do}(Z=z^\star),x)}$};
\end{scope}

\begin{scope}[xshift=14.0cm]
    \node[title] at (1.55,2.65) {Case 2};
    \node[desc] at (1.55,1.95) {not identifiable\\even by data fusion};

    \node[sur] (z3) at (0,0) {$z$};
    \node[trt] (t3) at (1.5,0) {$t$};
    \node[outcome] (y3) at (3,0) {$y$};
    \node[cov] (x3) at (1.5,1.15) {$x$};
    \node[unobs] (u13) at (1.15,-0.95) {$u_1$};
    \node[unobs] (u23) at (2.25,-0.95) {$u_2$};

    \draw[arr] (z3) -- (t3);
    \draw[arr] (t3) -- (y3);

    \draw[arr] (x3) -- (z3);
    \draw[arr] (x3) -- (t3);
    \draw[arr] (x3) -- (y3);

    \draw[harr] (u13) -- (z3);
    \draw[harr] (u13) -- (y3);
    \draw[harr] (u23) -- (t3);
    \draw[harr] (u23) -- (y3);

    \node[formula] at (1.55,-1.85)
    {remaining $T\leftrightarrow Y$ \\ hidden confounding};
\end{scope}

\end{tikzpicture}%
    }
    \caption{\textbf{Case studies:} Three roles of surrogate experiments for \(p(y\mid\doop{T=t},x)\). In Case~0, the CID is identifiable
    from observational data alone. In Case~1, it becomes identifiable after
    incorporating experiments on \(Z\). In Case~2, it remains
    non-identifiable but with a potentially smaller identified set.}
    \label{fig:fusion_graphical_id}
\end{figure*}

\paragraph{Point identification.}
We first consider the infinite-data limit, in which all available distributions
are known exactly. A target CID is point-identifiable if it is the same across
all SCMs compatible with the available population distributions. Conditional
generalised identification (\(c\)-\(g\)ID) characterises whether such a
conditional causal query can be uniquely recovered from an arbitrary
collection of observational and interventional distributions under an assumed
causal graph
\citep{bareinboim2012causal,lee2020general,kivva2023identifiability}.
\cref{fig:fusion_graphical_id} illustrates three possible cases.

\paragraph{Partial identification.}
Failure of point identification does not imply that a surrogate
experiment is uninformative.
Different SCMs can imply the same observational distribution
but different interventional distributions.
Requiring agreement with the experimental distribution can
therefore exclude some of these observationally compatible SCMs.
A graphical illustration arises in DAG models without latent
confounding: known perfect interventions can distinguish graphs
within an observational Markov equivalence class (MEC), refining
it to a potentially smaller interventional MEC.
This is because removing incoming edges to the intervention
targets can expose differences between otherwise observationally
equivalent causal structures~\citep{hauser2012characterization}.

To formalise this principle, consider a specified class of SCMs,
where \(\psi\) encodes both the graph and its mechanisms.
Let \([\psi^\star]_O\) denote the SCMs compatible with the
population observational distribution, and let
\([\psi^\star]_{O+I}\) additionally require compatibility with
the available surrogate interventional distribution.
For \(q_y(\psi)=p_\psi(y\mid\doop{T=t},x)\), define the
corresponding identified sets as
\(\mathcal{Q}_S=\{q_y(\psi):\psi\in[\psi^\star]_S\}\).
The additional compatibility requirement implies
\begin{equation}
    [\psi^\star]_{O+I}\subseteq[\psi^\star]_O
    \quad\Longrightarrow\quad
    \mathcal{Q}_{O+I}\subseteq\mathcal{Q}_O.
    \label{eq:identified-set-monotonicity}
\end{equation}
Thus, at the population level, surrogate experiments can shrink
the target's identified set or leave it unchanged.
The inclusion may be strict even when
\(\mathcal{Q}_{O+I}\) is not a singleton.


\paragraph{Finite data.}
For fixed query values \(t,x,y\), let
\(q_y(\psi)=p_\psi(y\mid\mathrm{do}(T=t),x)\), where
\(\psi\) encodes both the graph and its mechanisms.
The posterior predictive density (PPD) targeted by a CFM is
\[
    \widehat q_y(\mathcal D)
    =
    \int
    p_\psi(y\mid\mathrm{do}(T=t),x)\,
    p(\psi\mid\mathcal D)\,d\psi.
\]
Suppose that \(\psi\sim p(\psi)\) and both datasets are generated
from this same SCM under their respective regimes.
For \(S\in\{O,O+I\}\), define the prior-averaged mean squared error
\[
    R_S
    :=
    \mathbb E_{\psi,\mathcal D_S}
    \left[
        \bigl(q_y(\psi)-\widehat q_y(\mathcal D_S)\bigr)^2
    \right],
\]
where \(\mathcal D_O=\obsdata\) and
\(\mathcal D_{O+I}=(\obsdata,\intdata)\).
If \(\mathbb E_\psi[q_y(\psi)^2]<\infty\), then
\begin{equation}
    R_O-R_{O+I}
    =
    \mathbb E_{\obsdata,\intdata}
    \left[
        \bigl(
            \widehat q_y(\obsdata,\intdata)
            -\widehat q_y(\obsdata)
        \bigr)^2
    \right]
    \geq 0.
    \label{eq:finite-main-result}
\end{equation}
Thus, incorporating surrogate-interventional data cannot increase
the prior-averaged MSE between the exact PPD and the underlying
SCM's target density. The reduction is strict whenever the additional
data change the PPD at \(y\) with positive probability.
See Appendix~\ref{app:finite_mse} for the proof. Please note the CFM is trained to approximate $\hat{q}$, which might introduces additional approximation error.

\section{Input representation and model architecture}
\label{sec:architecture}

CIDER-FM receives context datasets generated by the same unknown causal system under different experimental regimes. For each query row, it predicts a conditional interventional distribution for the masked outcome. Its architecture represents both how each row was generated and how evidence is exchanged within and across regimes. Appendix~\ref{app:method} provides details of the input
representation, attention layers, and predictive distribution.

\paragraph{Intervention-aware input representation.}
A task contains \(R\) context datasets \(\regimedatasets\), where regime \(r\) provides data
\(\mathcal{D}_r=\{\mathbf{v}_{r,i}\}_{i=1}^{n_r}\)
over the same \(p\) variables. Each regime has an intervention-target mask
\(\mathbf{a}_r\in\{0,1\}^{p}\), where \(a_{r,j}=1\) indicates that variable \(j\) is intervened upon. Each row also has a vector of assigned intervention values
\(\mathbf{c}_{r,i}\in\mathbb{R}^{p}\), whose entries are relevant only for intervened variables. Together, \((\mathbf{a}_r,\mathbf{c}_{r,i})\) specifies which structural assignments were externally replaced and the values assigned in row \(i\). Defining
\(\mathcal{I}_r=\{j:a_{r,j}=1\}\), the row is generated according to
\begin{equation}
\mathbf{v}_{r,i}
\sim
p_{\psi}\!\left(
\mathbf{V}
\mid
\doop{
\mathbf{V}_{\mathcal{I}_r}
=
\mathbf{c}_{r,i,\mathcal{I}_r}
}
\right),
\label{eq:regime-input}
\end{equation}
where \(\psi\) denotes the underlying causal system. The observational regime is represented by \(\mathbf{a}_r=\mathbf{0}\), allowing observational and interventional data to share the same input format. CIDER-FM accommodates varying numbers of regimes and samples per regime.

CIDER-FM represents every table cell as a \(d\)-dimensional token. Each token combines information about the entire row, the individual cell value, the variable's identity and task role, the availability of its value, and its intervention metadata. Writing tildes for preprocessed data values and intervention assignments, the initial token for variable \(j\) in context row \(i\) of regime \(r\) is
\begin{equation}
\mathbf{h}^{C,0}_{r,i,j}
=
\alpha\,\phi_{\mathrm{row}}(\widetilde{\mathbf{v}}_{r,i})
+
\beta\,\phi_{\mathrm{val}}(\widetilde v_{r,i,j})
+
\mathbf{e}^{\mathrm{id}}_j
+
\mathbf{e}^{\mathrm{role}}_{\rho_j}
+
\mathbf{e}^{\mathrm{status}}_{\mathrm{obs}}
+
\mathbf{e}^{\mathrm{int}}
(a_{r,j},\widetilde c_{r,i,j}).
\label{eq:context-token}
\end{equation}
The function
\(\phi_{\mathrm{row}}:\mathbb{R}^{p}\to\mathbb{R}^{d}\)
encodes the entire row, and its output is shared across all cell tokens in that row. The function
\(\phi_{\mathrm{val}}:\mathbb{R}\to\mathbb{R}^{d}\)
encodes the individual cell value. The scalar coefficients \(\alpha\) and \(\beta\) weight these two contributions.

The indexed \(\mathbf{e}\) terms are learned embedding vectors selected by discrete labels. The vector \(\mathbf{e}^{\mathrm{id}}_j\) identifies the variable slot, while
\(\mathbf{e}^{\mathrm{role}}_{\rho_j}\) identifies its task role, with \(\rho_j\) denoting treatment, outcome, or conditioning variable. The vector
\(\mathbf{e}^{\mathrm{status}}_{\mathrm{obs}}\) indicates that the cell value is available to the model. 

The intervention encoding \(\mathbf{e}^{\mathrm{int}}(a,c)\) is a vector-valued function that combines an embedding of the intervention status \(a\) with an encoding of the assigned value \(c\): $\mathbf{e}^{\mathrm{int}}(a,c)
=
\mathbf{e}^{\mathrm{int\mbox{-}status}}_a
+
a\,\phi_{\mathrm{int}}(c).
\label{eq:intervention-token}$
Here, \(\mathbf{e}^{\mathrm{int\mbox{-}status}}_0\) and
\(\mathbf{e}^{\mathrm{int\mbox{-}status}}_1\) are learned embedding vectors indicating whether the variable is intervened upon, and
\(\phi_{\mathrm{int}}\)
encodes the assigned intervention value. For \(a=0\), the encoding contains only the non-intervened status vector. For \(a=1\), it also includes the encoded assignment \(\phi_{\mathrm{int}}(c)\). This distinguishes an intervention assigning the value zero from the absence of an intervention.

Query tokens use the same construction, with the unknown outcome replaced by a masked-target embedding and the requested intervention supplied as metadata. To keep intervention information available throughout the network, the intervention encoding is also injected into every attention layer through a learned gated residual connection.

\paragraph{Hierarchical three-axis attention.}
Existing table-aware architectures typically exchange information across the variable and sample axes. CIDER-FM introduces the experimental regime as a third axis. Each of its \(L\) layers first applies variable-axis attention to the \(p\) cell tokens within every context and query row, allowing each token to incorporate information from the other variables in that row. It then applies sample-axis attention independently for each regime and variable, allowing samples within a regime to exchange information while preserving the separation between regimes.

After sample-axis attention, each \((\text{regime},\text{variable})\) pair contains a set of sample tokens. Learned attention pooling compresses this set into \(K\) memory tokens, producing a fixed-size summary for each variable in each regime. For each variable, regime-axis self-attention then fuses the corresponding memories from all \(R\) regimes. These variable-aligned memories aggregate evidence about the same variable across observational and interventional settings. Finally, separate cross-attention blocks use the fused memories to update the context and query tokens before the next layer.

After the final layer, a linear readout maps the token corresponding to the masked query outcome to the parameters of a full-support bar distribution \citep{robertson2026pfn, hollmann2025accurate}. The model is trained end-to-end by minimising the negative log-likelihood of sampled query outcomes.

\section{Experiments}
\label{sec:experiments}

We evaluate CIDER-FM's ability to predict CIDs from data collected across multiple experimental regimes.
We begin with general random graphs, considering both
linear--Gaussian mechanisms and heterogeneous nonlinear mechanisms
with heavy-tailed noise, then use restricted graph families to examine
performance stratified by three identification settings in
\cref{fig:fusion_graphical_id}.
We then evaluate on real-world data from Causal Chambers
\citep{gamella2025causal}.
Finally, we conduct architectural ablations to assess the
computational efficiency of three-axis attention.

\paragraph{Baselines}
We compare CIDER-FM with predictive and causal inference
methods (\cref{tab:restricted_lingauss_baselines}), and CIDER-FM-Obs, a counterpart trained for the same task using only observational context.
The general-random-graph and real-world benchmarks include
Bayesian linear regression (Bayesian LR), TabPFN v2
\citep{hollmann2025accurate}, and Graph4CFM
\citep{reuter2026use}, with varying levels of ancestral
information.
The restricted-family benchmark additionally includes linear
regression (LR), random forests (RF), partially linear double
machine learning (DML) \citep{chernozhukov2018double},
Do-PFN \citep{robertson2026pfn}, and ArCO-GP \citep{pmlr-v258-toth25a}.
For the restricted families, we additionally report
\(c\)-\(g\)ID-based estimators supplied with the ground-truth
ADMG as graph-informed references, excluding them from the
main rankings. Given the misspecification affecting external CFMs in
our setting (Appendix~\ref{app:baselines-metrics}), we use
CIDER-FM-Obs as our primary matched control.

\paragraph{Metrics}
For each query, the model predicts
\(p(y\mid\doop{T=t},\mathbf{x},\mathcal{D})\).
We distinguish two evaluation targets: the sampled query
outcome \(Y_q\) and the generating SCM's conditional
interventional mean,
\(Y_{\mathrm{mean}}=\mathbb{E}_{\psi}[Y\mid\doop{T=t},\mathbf{x}]\).
Model predictions are evaluated using negative log-likelihood (NLL), and
its predictive mean using MSE and \(R^2\). Where available, comparison with
\(Y_{\mathrm{mean}}\) allows us to ignore the effect of noise added to the mean. We distinguish three context configurations:
\vspace{-0.3em}
\begin{itemize}[leftmargin=1.2em, nosep]
    \item \texttt{fusion}: observational and experimental samples from all
    available regimes.
    \item \texttt{obs\_total}: observational samples matching the total number of samples in the fusion context, giving a fair comparison of the contribution of experimental context.
    \item \texttt{obs\_fixed}: observational samples matching only the
    observational portion of the fusion context.
\end{itemize}

\subsection{Synthetic Data}

\subsubsection{General random graphs}
\label{sec:general-random-graphs}

\providecommand{\tblse}[1]{{\normalfont\fontsize{6}{7}\selectfont\color{black!55}\(\,\pm\,#1\)}}

\begin{table}[!t]
\centering
\captionsetup{font=small,justification=raggedright,singlelinecheck=false,skip=6pt}
\caption{Performance on general random graphs: 560 linear--Gaussian
and 512 ComplexMech test SCMs.
Entries report the indicated statistic with its SCM-level
standard error conditional on fixed checkpoints.
Best point estimates are in bold.
Infinite NLL arises when the predicted density is numerically
zero for a query outcome.
ComplexMech \(Y_{\mathrm{mean}}\) is omitted because reliable
ground-truth conditional means are unavailable.}
\label{tab:general-performance}

\begingroup
\fontsize{8.5}{10}\selectfont
\setlength{\tabcolsep}{1pt}
\renewcommand{\arraystretch}{1.20}
\renewcommand{\tabularxcolumn}[1]{m{#1}}
\providecommand{\tblse}[1]{}
\renewcommand{\tblse}[1]{%
  \scalebox{0.9}{{\normalfont\fontsize{7}{7}\selectfont
  \color{black!55}\(\,\pm\,#1\)}}}

\begin{tabularx}{\textwidth}{@{}c@{\hspace{10pt}}l*{5}{>{\centering\arraybackslash}X}@{}}
\toprule
\multicolumn{2}{@{}l}{\textbf{Method}}
& \textbf{CIDER-FM}
& CIDER-FM-Obs
& Bayesian LR
& TabPFN v2
& Graph4CFM \\
\multicolumn{2}{@{}l}{%
  \multirow[c]{2}{*}{\textbf{Data \& Info}}}
& \texttt{fusion}
& \texttt{obs\_total}
& \texttt{obs\_total}
& \texttt{obs\_total}
& \texttt{obs\_total} \\
\multicolumn{2}{@{}l}{}
& & & & & all ancestry \\
\midrule
\multirow{4}{*}{\shortstack[c]{\textbf{Linear-Gaussian}\\\(Y_{\mathrm{mean}}\)}} & MSE \(\downarrow\)
& \textbf{95.03}\tblse{35.10} & 471.72\tblse{167.56} & 795.25\tblse{247.80} & 386.37\tblse{213.81} & 160.10\tblse{60.11} \\
& \(R^2_{\mathrm{mean}}\uparrow\)
& \textbf{0.675}\tblse{0.045} & -0.016\tblse{0.030} & -21.719\tblse{7.873} & -3.608\tblse{1.736} & 0.474\tblse{0.082} \\
& \(R^2_{\mathrm{med}}\uparrow\)
& \textbf{0.947}\tblse{0.006} & -0.016\tblse{0.006} & 0.843\tblse{0.032} & 0.897\tblse{0.016} & 0.811\tblse{0.019} \\
& \(R^2_{\mathrm{pool}}\uparrow\)
& \textbf{0.895}\tblse{0.067} & 0.479\tblse{0.144} & 0.122\tblse{0.633} & 0.573\tblse{0.347} & 0.823\tblse{0.101} \\
\midrule
\multirow{5}{*}{\shortstack[c]{\textbf{Linear-Gaussian}\\\(Y_q\)}} & NLL \(\downarrow\)
& \textbf{2.059}\tblse{0.061} & 2.445\tblse{0.057} & 13.990\tblse{1.961} & \(+\infty\) & \(3.25\mathrm{e}{7}\)\tblse{2.40\mathrm{e}{7}} \\
& MSE \(\downarrow\)
& \textbf{138.97}\tblse{37.70} & 510.72\tblse{168.72} & 835.55\tblse{247.73} & 425.30\tblse{213.14} & 206.73\tblse{63.39} \\
& \(R^2_{\mathrm{mean}}\uparrow\)
& \textbf{0.587}\tblse{0.019} & 0.065\tblse{0.015} & -7.963\tblse{2.763} & -0.167\tblse{0.114} & 0.498\tblse{0.020} \\
& \(R^2_{\mathrm{med}}\uparrow\)
& \textbf{0.741}\tblse{0.024} & -0.009\tblse{0.003} & 0.559\tblse{0.060} & 0.681\tblse{0.047} & 0.599\tblse{0.037} \\
& \(R^2_{\mathrm{pool}}\uparrow\)
& \textbf{0.853}\tblse{0.080} & 0.461\tblse{0.141} & 0.118\tblse{0.583} & 0.551\tblse{0.327} & 0.782\tblse{0.110} \\
\midrule
\multirow{5}{*}{\shortstack[c]{\textbf{ComplexMech}\\\(Y_q\)}} & NLL \(\downarrow\)
& -0.395\tblse{0.078} & \textbf{-0.416}\tblse{0.079} & 2.081\tblse{0.437} & \(+\infty\) & \(2.12\mathrm{e}{8}\)\tblse{1.28\mathrm{e}{8}} \\
& MSE \(\downarrow\)
& \textbf{289.26}\tblse{94.06} & 299.26\tblse{95.40} & 2588.10\tblse{1314.80} & 771.98\tblse{405.36} & 404.89\tblse{124.91} \\
& \(R^2_{\mathrm{mean}}\uparrow\)
& \textbf{0.245}\tblse{0.022} & 0.224\tblse{0.027} & \(-4.14\mathrm{e}{4}\)\tblse{4.09\mathrm{e}{4}} & -0.054\tblse{0.119} & 0.133\tblse{0.032} \\
& \(R^2_{\mathrm{med}}\uparrow\)
& 0.110\tblse{0.026} & \textbf{0.112}\tblse{0.022} & -0.061\tblse{0.011} & 0.057\tblse{0.020} & 0.043\tblse{0.019} \\
& \(R^2_{\mathrm{pool}}\uparrow\)
& \textbf{0.935}\tblse{0.052} & 0.933\tblse{0.057} & 0.420\tblse{0.562} & 0.827\tblse{0.153} & 0.909\tblse{0.073} \\
\bottomrule
\end{tabularx}
\endgroup
\end{table}

We first sample random acyclic directed mixed graphs with 4--10 observed
variables, allowing hidden confounding through bidirected edges.
We consider linear--Gaussian mechanisms and ComplexMech, which combines
randomly sampled MLP mechanisms, additive or non-additive noise,
different noise scales, as well as heavy-tailed noise.
Appendix~\ref{app:general} provides the data construction.
Some tasks have very little within-task target variation and therefore
exhibit extreme negative \(R^2\). We therefore report the mean and median task-level
\(R^2\), together with \(R^2\) computed by pooling query rows across SCMs.

\begin{samepage}
Table~\ref{tab:general-performance} shows that CIDER-FM achieves
the best point estimates across all reported metrics for both
\(Y_{\mathrm{mean}}\) and \(Y_q\) under linear--Gaussian mechanisms.
Its advantage over CIDER-FM-Obs and Graph4CFM highlights the
benefit of experimental context, while its comparison with
TabPFN illustrates the difference between
\(p(y\mid\doop{T=t},\mathbf{x})\) and \(p(y\mid T=t,\mathbf{x})\).

\end{samepage}

Under ComplexMech, CIDER-FM continues to outperform both
predictive baselines across the reported metrics.
Compared with TabPFN, query MSE decreases from \(771.98\)
to \(289.26\), while mean task-level
\(R^2\) increases from \(-0.054\) to \(0.245\).
The gains over CIDER-FM-Obs are smaller than under
linear--Gaussian mechanisms, with particularly similar
NLL, median \(R^2\), and pooled \(R^2\).

An advantage over the observation-only counterpart is not
guaranteed when experimental regimes are selected randomly
under a fixed total context budget.
For example, intervening on an isolated surrogate \(Z\)
leaves the joint distribution of \((T,Y)\) unchanged and
does not resolve any existing confounding between treatment
and outcome.
To better understand when experimental context is beneficial,
we next stratify tasks by the identifiability of the target
causal query.

\subsubsection{Stratification by identifiability}

\label{sec:restricted-graphs}

\begin{figure*}[!t]
  \centering
  \captionsetup{font=small,skip=3pt}
  \captionsetup[subfigure]{skip=2pt}

  \begin{subfigure}{\textwidth}
    \centering
    \includegraphics[width=\linewidth]
      {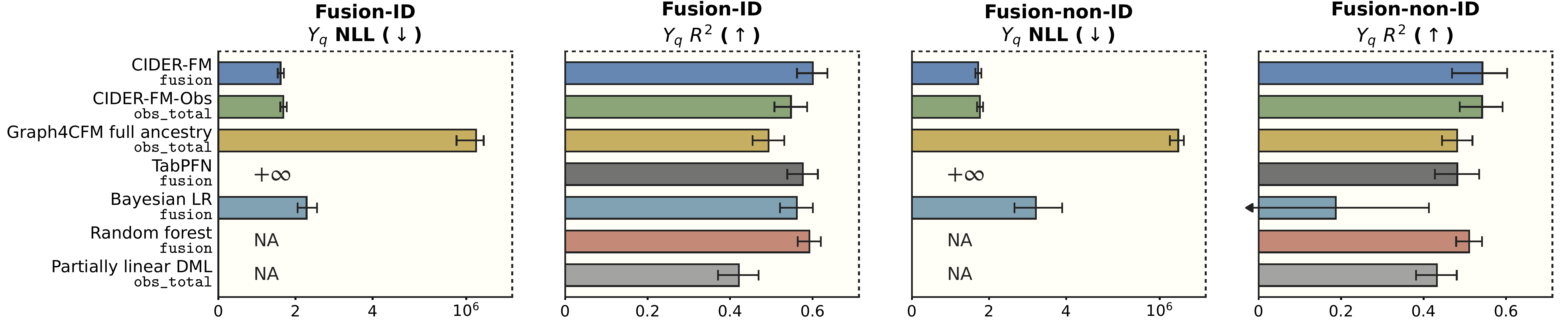}
    \caption{Linear--Gaussian}
    \label{fig:restricted_linear}
  \end{subfigure}
  \begin{subfigure}{\textwidth}
    \centering
    \includegraphics[width=\linewidth]
      {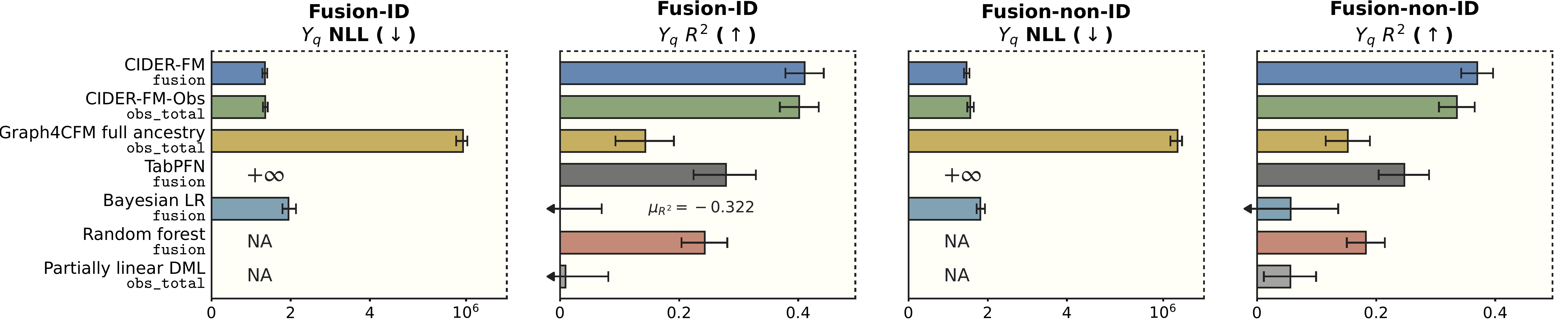}
    \caption{Nonlinear--Gaussian}
    \label{fig:restricted_nonlinear}
  \end{subfigure}

  \caption{Selected baseline comparisons on restricted graph
  families in \texttt{fusion\_id} and \texttt{fusion\_non\_id}.
  Error bars indicate 95\% confidence intervals.}
  \label{fig:restricted_comparison}
\end{figure*}

To examine how predictive performance varies with identifiability,
we next consider restricted graph families in which \(Y\) is
a descendant of both the treatment \(T\) and surrogate \(Z\),
with hidden confounding represented by bidirected edges.
The query conditions on \(\mathbf{X}\), comprising all observed
variables other than \(T\) and \(Y\), including \(Z\).
Contexts contain observational data and interventional data
on \(Z\), optionally supplemented by an intervention on another
observed covariate \(X_j\in\mathbf{X}\setminus\{Z\}\).
Using \(c\)-\(g\)ID under the generating graph, we distinguish
\texttt{obs\_id}, \texttt{fusion\_id}, and \texttt{fusion\_non\_id} settings
as defined in \cref{fig:fusion_graphical_id}. Appendix~\ref{app:restricted} describes the detailed construction.

The rank comparison contains 19 method configurations for point-prediction
metrics and 11 with reported density scores. It includes both observational
and pooled-data baselines, together with Graph4CFM variants with and without
ancestral information. Graph-informed \(c\)-\(g\)ID estimators are shown as
additional references in the full tables (\cref{tab:restricted-lg-all-methods,tab:restricted-nonlinear-all-methods}).

\noindent\begin{minipage}{\textwidth}
\setlength{\parskip}{4pt}
\setlength{\parindent}{0pt}
\begin{wraptable}{r}{0.48\textwidth}
\vspace{-\intextsep}
\captionsetup{font=small,justification=raggedright,singlelinecheck=false,skip=4pt,hypcap=false}
\centering
\caption{Rank-1 and top-3 frequencies
of CIDER-FM over five metrics in each of three settings, giving 15
setting--metric combinations per mechanism family.}
\label{tab:restricted-validation-rank}
\begingroup
\fontsize{8.5}{10}\selectfont
\setlength{\tabcolsep}{3pt}
\renewcommand{\arraystretch}{1.10}
\begin{tabularx}{\linewidth}{@{}>{\raggedright\arraybackslash}Xcc@{}}
\toprule
\textbf{Mechanism} & \textbf{Rank 1} & \textbf{Top 3} \\
\midrule
Linear--Gaussian & 9/15 (60\%) & 15/15 (100\%) \\
Nonlinear--Gaussian & 7/15 (46.7\%) & 13/15 (86.7\%) \\
\bottomrule
\end{tabularx}
\endgroup
\end{wraptable}

The three identification settings and five evaluation metrics (NLL, MSE, and \(R^2\) for \(Y_q\);
MSE and \(R^2\) for \(Y_{\mathrm{mean}}\))
yield 15 setting--metric combinations per mechanism family.
CIDER-FM ranks among the top three configurations in all
15 linear--Gaussian combinations and 13 of 15 nonlinear--Gaussian
combinations (Table~\ref{tab:restricted-validation-rank}).
Detailed rankings appear in
\cref{tab:restricted-lg-validation-ranking,tab:restricted-complex-validation-ranking}.

\par\WFclear
\end{minipage}

\cref{fig:restricted_linear} shows that, under linear--Gaussian
mechanisms, CIDER-FM achieves the lowest NLL and highest
\(Y_q\) \(R^2\) point estimates in both fusion settings.
In \texttt{fusion\_id}, where the available experiments enable
identification under the generating graph, CIDER-FM improves
over CIDER-FM-Obs on both metrics, highlighting the benefit
of experimental context.
In \texttt{fusion\_non\_id}, CIDER-FM continues to perform best,
despite the query remaining non-identifiable.
However, its advantage over CIDER-FM-Obs is smaller in this
setting, with nearly identical \(Y_q\) \(R^2\).

\subsection{Real-world evaluation}
\label{sec:real-world}

We then evaluate CIDER-FM on the light-tunnel task from Causal
Chambers~\citep{gamella2025causal}.
The task concerns how intervening on the red-channel light-source
setting \(T=\mathrm{red}\) affects the downstream light-intensity
measurement \(Y=\mathrm{vis}_3\), conditional on the green- and
blue-channel settings
\(\mathbf{X}=(\mathrm{green},\mathrm{blue})\).
We withhold both polarizer variables, leaving the models with
incomplete information about the optical system.
The goal is to predict the conditional outcome distribution
under interventions on the red channel using observational
measurements and experiments on other observed variables controlling the light-tunnel experiment.

We select the experimental data sources through
\emph{simulation-based experimental design}.
Using data generated by the official simulator corresponding to the light-tunnel experiment \citep{gamella2025sanity}, we set up different combinations of observational and interventional
regimes under a fixed total context budget. 
We evaluate these candidate designs with the frozen ComplexMech CIDER-FM in \cref{sec:general-random-graphs}
and select based on NLL on the simulated data.
This procedure selects observational data combined with
interventions on \(\mathrm{green}\).
We then freeze the design and evaluate it using real measurements
(without reselecting regimes based on real-data performance).
No model retraining is performed, and simulated samples are
excluded from the real-data context.
(See Appendix~\ref{app:chambers}).

Table~\ref{tab:chambers-performance} shows that the selected design
reduces MSE by approximately \(24.7\%\) relative to the
equal-budget TabPFN baseline.
Relative to CIDER-FM-Obs, point-prediction scores are slightly
better, while NLL is comparable but slightly worse.
These results suggest that CIDER-FM could support future experiments
by selecting query-specific experimental designs using only
simulator-generated data.
The selected interventions could then be carried out in the real
world, enabling targeted data collection aimed at reducing
uncertainty about a causal query, alongside causal prediction
from the resulting real world context again with CIDER-FM.

\begin{table}[!t]
\centering
\captionsetup{font=small,justification=raggedright,singlelinecheck=false,skip=6pt}
\caption{Real-world performance on Causal Chambers.
Values are estimates \(\pm\) standard errors.
Bold indicates the best point estimate in each column.
For Graph4CFM, arrows denote supplied ancestral relations;
R, G, and B denote the red, green, and blue channels, respectively.}
\label{tab:chambers-performance}

\begingroup
\fontsize{8.5}{10}\selectfont
\setlength{\tabcolsep}{2pt}
\renewcommand{\arraystretch}{1.16}
\renewcommand{\tabularxcolumn}[1]{m{#1}}
\providecommand{\tblse}[1]{}
\renewcommand{\tblse}[1]{%
  \scalebox{0.9}{{\normalfont\fontsize{7}{7}\selectfont
  \color{black!55}\(\,\pm\,#1\)}}}

\begin{tabularx}{\textwidth}{@{}ll*{5}{>{\raggedleft\arraybackslash}X}@{}}
\toprule
\textbf{Method}
& \textbf{Data \& Info}
& \(\boldsymbol{Y_q}\) \textbf{NLL} \(\downarrow\)
& \(\boldsymbol{Y_q}\) \textbf{MSE} \(\downarrow\)
& \(\boldsymbol{Y_q\,R^2_{\mathrm{mean}}}\uparrow\)
& \(\boldsymbol{Y_q\,R^2_{\mathrm{med}}}\uparrow\)
& \(\boldsymbol{Y_q\,R^2_{\mathrm{pool}}}\uparrow\) \\
\midrule
\textbf{CIDER-FM} & \texttt{fusion}
& 5.772\tblse{0.031} & \textbf{6810}\tblse{399}
& \textbf{0.367}\tblse{0.026} & \textbf{0.380}\tblse{0.038} & \textbf{0.374}\tblse{0.026} \\
CIDER-FM-Obs & \texttt{obs\_total}
& \textbf{5.770}\tblse{0.036} & 6861\tblse{340}
& 0.360\tblse{0.026} & 0.351\tblse{0.029} & 0.370\tblse{0.026} \\
CIDER-FM-Obs & \texttt{obs\_fixed}
& 5.816\tblse{0.042} & 7057\tblse{392}
& 0.343\tblse{0.028} & 0.318\tblse{0.038} & 0.352\tblse{0.028} \\
CIDER-FM & \texttt{obs\_total}
& 5.792\tblse{0.029} & 6956\tblse{375}
& 0.352\tblse{0.025} & 0.350\tblse{0.026} & 0.361\tblse{0.026} \\
Bayesian LR & \texttt{obs\_total}
& 5.850\tblse{0.027} & 7099\tblse{400}
& 0.338\tblse{0.029} & 0.352\tblse{0.051} & 0.348\tblse{0.029} \\
TabPFN v2 & \texttt{obs\_total}
& 5.930\tblse{0.062} & 9042\tblse{670}
& 0.156\tblse{0.062} & 0.172\tblse{0.078} & 0.169\tblse{0.052} \\
Graph4CFM & \texttt{obs\_total}
& 5.870\tblse{0.044} & 7156\tblse{366}
& 0.334\tblse{0.025} & 0.316\tblse{0.034} & 0.342\tblse{0.025} \\
Graph4CFM & \texttt{obs\_total}; \(\mathrm{R}\to Y\)
& 5.970\tblse{0.043} & 7946\tblse{445}
& 0.259\tblse{0.035} & 0.256\tblse{0.044} & 0.270\tblse{0.035} \\
Graph4CFM & \texttt{obs\_total}; \(\mathrm{R},\mathrm{G},\mathrm{B}\to Y\)
& 6.039\tblse{0.045} & 8582\tblse{481}
& 0.199\tblse{0.041} & 0.219\tblse{0.054} & 0.211\tblse{0.038} \\
\bottomrule
\end{tabularx}
\endgroup
\end{table}

\subsection{Ablation studies}
\label{sec:ablation-studies}

\noindent\begin{minipage}{\textwidth}
\setlength{\parskip}{4pt}
\setlength{\parindent}{0pt}

\begin{wraptable}{r}{0.60\textwidth}
\vspace{-\intextsep}
\centering
\captionsetup{
    font=small,
    justification=raggedright,
    singlelinecheck=false,
    skip=4pt,
    hypcap=false
}
\caption{Throughput ratios (three-axis / two-axis) from
short-run profiling on an NVIDIA A100, with 10 variables,
\(R=3\), and eight memory tokens per regime and variable.
Same width: \(d_{\mathrm{3D}}=d_{\mathrm{2D}}=128\).
Parameter matched: \(d_{\mathrm{3D}}=128\),
\(d_{\mathrm{2D}}=160\).}
\label{tab:axis-context-throughput}

\begingroup
\fontsize{8.5}{10}\selectfont
\setlength{\tabcolsep}{2.5pt}
\renewcommand{\arraystretch}{1.10}

\begin{tabularx}{\linewidth}{
    @{}*{5}{>{\centering\arraybackslash}X}@{}
}
\toprule
\multirow{2}{*}{
    \shortstack{\textbf{Rows per}\\\textbf{regime}}
}
& \multicolumn{2}{c}{\textbf{Same width}}
& \multicolumn{2}{c}{\textbf{Parameter matched}}
\\
\cmidrule(lr){2-3}
\cmidrule(lr){4-5}
& \textbf{Inference}
& \textbf{Training}
& \textbf{Inference}
& \textbf{Training}
\\
\midrule
256
& 0.814 & 0.865 & 0.940 & \textbf{1.002}
\\
512
& 0.941 & \textbf{1.001} & \textbf{1.112} & \textbf{1.167}
\\
1024
& \textbf{1.108} & \textbf{1.199}
& \textbf{1.297} & \textbf{1.383}
\\
2048
& \textbf{1.357} & \textbf{1.484}
& \textbf{1.561} & \textbf{1.684}
\\
\bottomrule
\end{tabularx}
\endgroup
\end{wraptable}

Although two-axis architectures can learn from multiple
experimental regimes when scaled up and supplied with
intervention-aware inputs, our three-axis design aims to
improve computational efficiency as multi-regime contexts grow.

Table~\ref{tab:axis-context-throughput} shows that three-axis
attention can be slower for short contexts, but its relative
throughput improves as context length increases.
At 1024 and 2048 rows per regime, it achieves higher inference
and training throughput under both same-width and
parameter-matched comparisons.
These results support its computational advantage when
scaling to more samples per regime in the tested configuration.

\par
\ifnum\value{WF@wrappedlines}>1\relax
  \vspace{\dimexpr
    \value{WF@wrappedlines}\baselineskip
  \relax}
\fi
\WFclear
\end{minipage}

\section{Discussion, Limitations and Outlook}

CIDER-FM demonstrates that CFMs are a promising method for utilising multiple experimental regimes for causal inference. Due to limited available computational resources, our
current evaluation is limited to graphs with at most ten variables and does
not yet establish how the approach scales to larger systems. More compute would also enable longer pretraining and exploration
of more diverse and complex SCM priors.
Although the proposed input representation supports simultaneous
interventions on multiple variables, our experiments use only one intervention
target per regime and assume that all regimes share the same population and
non-intervened mechanisms. While in causality experiments on synthetic data are typically considered best practice to establish the fundamental efficacy of any causal inference approach \citep{poinsotposition, reuter2026use}, and we validate CIDER-FM on real-world data using Causal Chambers, the next step is to apply CIDER-FM to challenging real-world scientific applications such as single-cell perturbation studies \citep{gao2026perturbpfn}. We especially believe that the general setup of using simulators together with CIDER-FM for deciding which experiments to carry out is a very promising avenue.

\newpage
\bibliographystyle{iclr2027_conference}
\bibliography{references}

\newpage
\appendix

\section{Finite-data MSE reduction of the PPD}
\label{app:finite_mse}

Fix the query values \(t,x,y\).
All expectations below are taken under the joint distribution of SCM and data sampled from 
\[
    p(\psi,\obsdata,\intdata)
    =
    p(\psi)\,p(\obsdata,\intdata\mid\psi),
\]

Recall that the interventional posterior predictive density (PPD) is
\[
    \widehat q_y(\mathcal D)
    =
    \int
    p_\psi(y\mid\mathrm{do}(T=t),x)\,
    p(\psi\mid\mathcal D)\,d\psi.
\]
Integration over \(\psi\) includes summation over graphs and
integration over their mechanism parameters.
We need to assume
\[
    \mathbb E_\psi
    \left[
        p_\psi(y\mid\mathrm{do}(T=t),x)^2
    \right]
    <\infty.
\]
Note that by Jensen's inequality, this implies that the PPDs also have finite
second moments, so all expectations below are finite and well-defined. 

\paragraph{Proof of \cref{eq:finite-main-result}.}
By the definition of the PPD, for almost every pair of datasets, we have:
\begin{equation}
    \int
    \Bigl[
        p_\psi(y\mid\mathrm{do}(T=t),x)
        -\widehat q_y(\obsdata,\intdata)
    \Bigr]
    p(\psi\mid\obsdata,\intdata)\,d\psi
    =0.
    \label{eq:finite-zero-posterior-error}
\end{equation}

Here, the first term integrates to
\(\widehat q_y(\obsdata,\intdata)\) by definition, while the second
is constant with respect to \(\psi\) and the posterior
\(p(\psi\mid\obsdata,\intdata)\) integrates to one.

Now let's decompose the difference using observational data alone:
\[
\begin{aligned}
    &p_\psi(y\mid\mathrm{do}(T=t),x)
        -\widehat q_y(\obsdata)
    \\
    &=
    \Bigl[
        p_\psi(y\mid\mathrm{do}(T=t),x)
        -\widehat q_y(\obsdata,\intdata)
    \Bigr]
    +
    \Bigl[
        \widehat q_y(\obsdata,\intdata)
        -\widehat q_y(\obsdata)
    \Bigr].
\end{aligned}
\]
Squaring and taking expectations gives
\[
\begin{aligned}
    R_O
    ={}& R_{O+I}
    +
    \mathbb E_{\obsdata,\intdata}
    \left[
        \bigl(
            \widehat q_y(\obsdata,\intdata)
            -\widehat q_y(\obsdata)
        \bigr)^2
    \right]
    \\
    &+
    2\mathbb E_{\obsdata,\intdata}
    \Biggl[
        \bigl(
            \widehat q_y(\obsdata,\intdata)
            -\widehat q_y(\obsdata)
        \bigr)
        \\
    &\hspace{15mm}\cdot
        \int
        \Bigl[
            p_\psi(y\mid\mathrm{do}(T=t),x)
            -\widehat q_y(\obsdata,\intdata)
        \Bigr]
        p(\psi\mid\obsdata,\intdata)\,d\psi
    \Biggr].
\end{aligned}
\]
Here, the cross term is written by first averaging over the SCM
conditional on both datasets, and then averaging over the datasets.
The change between the two PPDs depends only on the datasets,
so it can be taken outside the inner integral.

The inner integral is zero by
\cref{eq:finite-zero-posterior-error}. Consequently,
\[
    R_O-R_{O+I}
    =
    \mathbb E_{\obsdata,\intdata}
    \left[
        \bigl(
            \widehat q_y(\obsdata,\intdata)
            -\widehat q_y(\obsdata)
        \bigr)^2
    \right]
    \geq 0.
\]
Since the integrand is a square, the inequality is strict exactly
when the two PPDs differ at \(y\) with positive probability under
the marginal distribution of \((\obsdata,\intdata)\).
\hfill\(\square\).

\section{Finite-sample asymptotic estimation error reduction}
\label{app:finite_asym}

We also provide theoretical results under the assumptions of a fully-parametric, continuously parametrised SCM. While continuous relaxions exist \citep{zheng2018dags}, SCMs, as normally defined, are semi-discrete objects with their DAGs being inherently discrete objects. In this case, the theory in this section only applies \emph{conditional} on a known causal DAG. 

Taking those caveats into account, we ask how additional interventional samples affect the estimation error of a
posterior predictive CID,
\begin{equation}
    \widehat q_y(\dataset)
    =
    p(y\mid \mathrm{do}(T=t),x,\dataset)
    =
    \int p_\psi(y\mid \mathrm{do}(T=t),x)p(\psi\mid \dataset)\,d\psi.
\end{equation}

\begin{assumption}[Parametric SCM and sampling scheme]
\label{ass:parametric-scm-sampling}
The data-generating process belongs to a correctly specified parametric SCM
family
\begin{equation}
    \{M_\psi:\psi\in\Psi\subseteq\mathbb{R}^d\},
\end{equation}
with true parameter \(\psi^\star\in\Psi\). Observational and
interventional samples are conditionally independent given \(\psi^\star\), and
the source label of each sample is known.
\end{assumption}

\begin{assumption}[Bernstein--von Mises regularity]
\label{ass:bvm-regularity}
The joint observational--interventional likelihood satisfies the standard
regularity conditions for a finite-dimensional parametric
Bernstein--von Mises approximation \citep{van2000asymptotic}.
\end{assumption}

\begin{assumption}[Smooth target CID]
For each fixed outcome value \(y\), the target CID functional
\begin{equation}
    q_y(\psi)=p_\psi(y\mid \mathrm{do}(T=t),x)
\end{equation}
is twice continuously differentiable in a neighbourhood of \(\psi^\star\).
\end{assumption}

\begin{assumption}[Local point-identifiability of the target CID]
The target CID is locally point-identifiable from the combined observational
and interventional distributions.
\end{assumption}

Let \(\obsdata=\{V_i^{\mathrm{obs}}\}_{i=1}^{n_O}\) be observational
samples from
\begin{equation}
    p_{\psi^\star}^{O}(\mathbf{v})
    =
    p_{\psi^\star}(\mathbf{V}=\mathbf{v}),
\end{equation}
and let \(\intdata=\{V_j^{\mathrm{int}}\}_{j=1}^{n_I}\) be additional
samples from an interventional source, for example
\begin{equation}
    p_{\psi^\star}^{I}(\mathbf{v})
    =
    p_{\psi^\star}(\mathbf{V}\setminus Z=\mathbf{v}\mid
    \mathrm{do}(Z=z^\star)).
\end{equation}
The joint log-likelihood is then
\begin{equation}
    \ell_{O+I}(\psi)
    =
    \sum_{i=1}^{n_O}
    \log p_\psi^{O}(V_i^{\mathrm{obs}})
    +
    \sum_{j=1}^{n_I}
    \log p_\psi^{I}(V_j^{\mathrm{int}}).
\end{equation}
Let
\begin{equation}
    I_O(\psi^\star)
    =
    \mathbb{E}_{\psi^\star}
    \left[
        \nabla_\psi \log p_\psi^{O}(V)
        \nabla_\psi \log p_\psi^{O}(V)^\top
    \right]_{\psi=\psi^\star}
\end{equation}
and
\begin{equation}
    I_I(\psi^\star)
    =
    \mathbb{E}_{\psi^\star}
    \left[
        \nabla_\psi \log p_\psi^{I}(V)
        \nabla_\psi \log p_\psi^{I}(V)^\top
    \right]_{\psi=\psi^\star}
\end{equation}
denote the per-sample Fisher information matrices from the observational and
interventional sources, respectively. 
Under the source independence in \cref{ass:parametric-scm-sampling},
the total Fisher information based on \(\obsdata\) and
\(\intdata\) is
\begin{equation}
    I_{O+I}^{(n)}
    =
    n_O I_O(\psi^\star)
    +
    n_I I_I(\psi^\star)
    \succeq
    n_O I_O(\psi^\star)s
\end{equation}
where \(\succeq\) denotes the Loewner order. The inequality follows from
\(I_I(\psi^\star)\succeq 0\).
For compactness, define
\begin{equation}
    \Sigma_{O+I}
    =
    \left(
        n_O I_O(\psi^\star)
        +
        n_I I_I(\psi^\star)
    \right)^{-1}.
\end{equation}
By the Bernstein--von Mises theorem \citep{van2000asymptotic}, the posterior
over the SCM parameter is asymptotically locally Gaussian around an efficient centre
\(\widehat{\psi}_{O+I}\), which may be taken to be the maximum likelihood
estimator or posterior mode:
\begin{equation}
    \psi\mid \obsdata,\intdata
    \;\dot{\sim}\;
    \mathcal{N}
    \left(
        \widehat{\psi}_{O+I},
        \Sigma_{O+I}
    \right).
\end{equation}
Since the target CID is a smooth functional of \(\psi\), the posterior mean of
\(q_y(\psi)\) is first-order equivalent to the plug-in value at \(\widehat{\psi}_{O+I}\):
\begin{equation}
    \widehat q_y(\obsdata,\intdata)
    =
    \int q_y(\psi)
    p(\psi\mid \obsdata,\intdata)\,d\psi
    =
    q_y(\widehat{\psi}_{O+I})
    +
    o_p(n^{-1/2}),
\label{eq:posterior-mean-plugin}
\end{equation}
where \(n=n_O+n_I\). This follows from the local Gaussian posterior approximation and a Taylor
expansion of \(q_y(\psi)\) around \(\widehat{\psi}_{O+I}\). The efficient centre itself satisfies the usual regular parametric
asymptotic normality \citep{van2000asymptotic}:
\begin{equation}
    \widehat{\psi}_{O+I}-\psi^\star
    \;\dot{\sim}\;
    \mathcal{N}
    \left(
        0,
        \Sigma_{O+I}
    \right).
\end{equation}
Thus a first-order Taylor expansion of the CID functional around \(\psi^\star\)
gives
\begin{equation}
    q_y(\widehat{\psi}_{O+I})
    -
    q_y(\psi^\star)
    =
    \nabla_\psi q_y(\psi^\star)^\top
    (\widehat{\psi}_{O+I}-\psi^\star)
    +
    o_p(\|\widehat{\psi}_{O+I}-\psi^\star\|).
\label{eq:taylor-plugin-cid}
\end{equation}
Combining \cref{eq:posterior-mean-plugin} with the Taylor expansion in \cref{eq:taylor-plugin-cid} yields
\begin{equation}
    \widehat q_y(\obsdata,\intdata)
    -
    q_y(\psi^\star)
    =
    \nabla_\psi q_y(\psi^\star)^\top
    (\widehat{\psi}_{O+I}-\psi^\star)
    +
    o_p(n^{-1/2}).
\end{equation}
Using the asymptotic normality of \(\widehat{\psi}_{O+I}\), we obtain the
local Gaussian approximation
\begin{equation}
    \widehat q_y(\obsdata,\intdata)
    -
    q_y(\psi^\star)
    \;\dot{\sim}\;
    \mathcal{N}
    \left(
        0,\;
        \nabla_\psi q_y(\psi^\star)^\top
        \Sigma_{O+I}
        \nabla_\psi q_y(\psi^\star)
    \right).
\end{equation}
The leading asymptotic variance of the posterior predictive CID is therefore
\begin{equation}
\boxed{
    \nabla_\psi q_y(\psi^\star)^\top
    \left(
        n_O I_O(\psi^\star)
        +
        n_I I_I(\psi^\star)
    \right)^{-1}
    \nabla_\psi q_y(\psi^\star).
}
\label{eq:var_cid}
\end{equation}

If the target CID is already locally identifiable from observational data
alone, then the corresponding observational-only approximation is
\begin{equation}
    \widehat q_y(\obsdata)
    -
    q_y(\psi^\star)
    =
    O_p
    \left(
        \sqrt{
            g_y^\top
            \left(
                n_O I_O(\psi^\star)
            \right)^{-1}
            g_y
        }
    \right).
\end{equation}
Because, whenever the inverses are well defined,
\begin{equation}
    g_y^\top
    \left(
        n_O I_O(\psi^\star)
        +
        n_I I_I(\psi^\star)
    \right)^{-1}
    g_y
    \leq
    g_y^\top
    \left(
        n_O I_O(\psi^\star)
    \right)^{-1}
    g_y.
\end{equation}
Thus, in a regular parametric setting, adding interventional samples cannot
increase the leading asymptotic variance of the posterior predictive CID when
the target is point-identifiable from the observational source. It may strictly
reduce the asymptotic error constant whenever the interventional source
provides information in a direction relevant to the CID functional. One
sufficient condition is
\begin{equation}
    I_I(\psi^\star)^{1/2}
    \left(
        n_O I_O(\psi^\star)
    \right)^{-1}
    g_y
    \neq 0,
\end{equation}
which states that the interventional likelihood contains information along a
direction that affects \(q_y(\psi)\).

\paragraph{Geometry of query-relevant information.}
\cref{eq:var_cid} shows that finite-sample CID error is controlled not by
the full Fisher information matrix alone, but by its inverse projected onto the
gradient direction of the target functional,
\(\nabla_\psi q_y(\psi^\star)\). This projection gives a local geometric view
of why interventional data can help.

Consider a small perturbation of the true SCM parameter in a tangent direction
\(v\),
\begin{equation}
    \psi_\epsilon=\psi^\star+\epsilon v .
\end{equation}
By the likelihood differentiability conditions underlying
\cref{ass:bvm-regularity}, the observational KL divergence between
the distributions induced by \(\psi^\star\) and \(\psi_\epsilon\) admits a
second-order expansion. To see this, write
\(\ell_O(\psi;V)=\log p_\psi^O(V)\). Then
\begin{equation}
    \ell_O(\psi^\star+\epsilon v;V)
    =
    \ell_O(\psi^\star;V)
    +
    \epsilon v^\top \nabla_\psi \ell_O(\psi^\star;V)
    +
    \frac{1}{2}\epsilon^2
    v^\top \nabla_\psi^2 \ell_O(\psi^\star;V)v
    +
    o(\epsilon^2).
\end{equation}
Taking expectation under \(p_{\psi^\star}^O\), the first-order term vanishes
because the score has mean zero, and the negative expected Hessian equals the
Fisher information. Hence
\begin{equation}
    \mathrm{KL}
    \bigl(
        p_{\psi^\star}^{O}
        \,\|\, 
        p_{\psi^\star+\epsilon v}^{O}
    \bigr)
    =
    \frac{1}{2}\epsilon^2
    v^\top I_O(\psi^\star)v
    +
    o(\epsilon^2).
\end{equation}
Thus, \(v^\top I_O(\psi^\star)v\) measures how distinguishable the local
perturbation \(\psi^\star+\epsilon v\) is from \(\psi^\star\) using
observational data. If
\begin{equation}
    v^\top I_O(\psi^\star)v=0,
\end{equation}
then observational data contain no local second-order information in this
direction: perturbations of the SCM along \(v\) are locally invisible to the
observational likelihood.

The same perturbation may nevertheless change the target CID. By \cref{eq:taylor-plugin-cid},
\begin{equation}
    q_y(\psi^\star+\epsilon v)
    -
    q_y(\psi^\star)
    =
    \epsilon
    \nabla_\psi q_y(\psi^\star)^\top v
    +
    o(\epsilon).
\end{equation}
Therefore, if
\begin{equation}
    \nabla_\psi q_y(\psi^\star)^\top v\neq 0,
\end{equation}
then moving along \(v\) changes the target CID even though observational data
cannot locally distinguish that movement. If the interventional source supplies information along the
same direction, for example in case
\begin{equation}
    v^\top I_I(\psi^\star)v>0,
\end{equation}
the interventional likelihood can locally distinguish perturbations along \(v\).
Then the combined information matrix
\begin{equation}
    n_O I_O(\psi^\star)+n_I I_I(\psi^\star)
\end{equation}
can become nonsingular on the query-relevant subspace. In this sense,
interventional data can change the problem from one in which the CID cannot be
regularly estimated from observational data alone to one in which the CID
admits a regular asymptotic approximation.

In conclusion, additional interventional samples can improve finite-sample CID
estimation by increasing the Fisher information available in query-relevant
directions.

\section{Details on representation and architecture}
\label{app:method}

\subsection{Data fusion: setting and scope}
\label{sec:architecture-setting}

We consider a task generated by a single causal system over a vector of
\(p\) observed continuous variables \(\mathbf{V}\in\mathbb{R}^{p}\). These include
a treatment \(T\), an outcome \(Y\), and conditioning or surrogate intervention
variables \(\mathbf{X}\) , with \(Z\in\mathbf{X}\) used where necessary to denote a
surrogate intervention target. The model receives a non-empty collection of finite datasets,
denoted \(\regimedatasets\), collected from this system under different
experimental regimes.

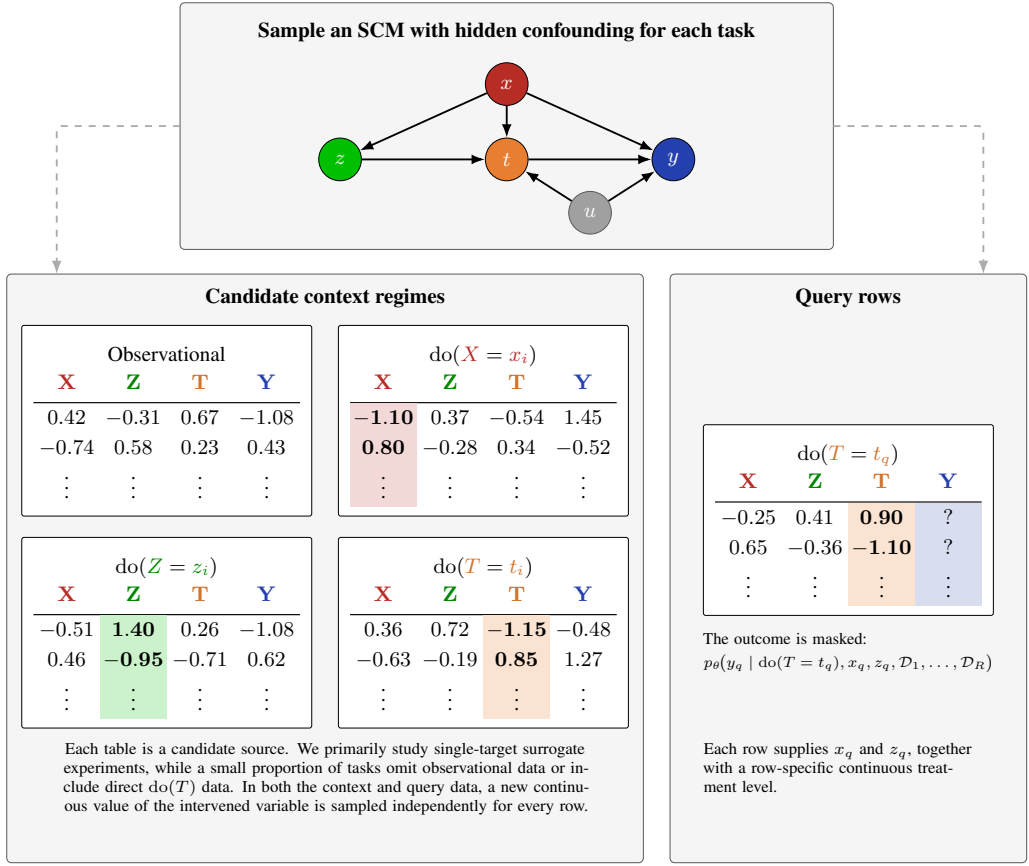
\begin{figure}[!t]
    \centering
    \resizebox{\textwidth}{!}{%
        \begingroup
\colorlet{revisionred}{black}
\setlength{\belowrulesep}{0pt}
\def\regcell#1{%
    \makebox[1.00cm][c]{\rule[-0.42em]{0pt}{1.42em}#1}%
}
\def\regmark#1#2{%
    \begingroup
    \setlength{\fboxsep}{0pt}%
    \colorbox{#1}{\makebox[1.00cm][c]{\rule[-0.42em]{0pt}{1.42em}#2}}%
    \endgroup
}
\def\regtitle#1{%
    \rule[-0.30em]{0pt}{1.55em}#1%
}
\begin{tikzpicture}[
    scale=1,
    transform shape,
    cov/.style={circle,draw=black,fill=covred,minimum size=6.4mm,inner sep=0pt,
        text=white,font=\bfseries\itshape\small},
    sur/.style={circle,draw=black,fill=surrogateyellow,minimum size=6.4mm,inner sep=0pt,
        text=white,font=\bfseries\itshape\small},
    trt/.style={circle,draw=black,fill=treatorange,minimum size=6.4mm,inner sep=0pt,
        text=white,font=\bfseries\itshape\small},
    outcome/.style={circle,draw=black,fill=outblue,minimum size=6.4mm,inner sep=0pt,
        text=white,font=\bfseries\itshape\small},
    unobs/.style={circle,draw=gray!70!black,fill=unobsgray,minimum size=6.4mm,inner sep=0pt,
        text=white,font=\bfseries\itshape\small},
    arr/.style={-{Latex[length=2mm]},thick},
    hiddenarr/.style={-{Latex[length=2mm]},thick},
    panel/.style={draw=black!65,rounded corners=2pt,fill=black!4},
    regtable/.style={draw=black,rounded corners=1pt,fill=white,
        minimum width=4.35cm,minimum height=2.86cm,inner sep=0pt},
    regcontent/.style={inner sep=0pt,align=center,font=\footnotesize},
    paneltitle/.style={font=\bfseries\small,align=center},
    reglabel/.style={font=\bfseries\scriptsize,align=center},
    note/.style={font=\scriptsize,align=center,inner sep=0pt}
]

\draw[panel] (2.6,1.15) rectangle (12.4,4.85);
\node[paneltitle] at (7.5,4.42)
    {Sample an SCM with hidden confounding for each task};

\begin{scope}[yshift=1.5mm]
\node[cov]     (x) at (7.5,3.48) {$x$};
\node[sur]     (z) at (5.0,2.35) {$z$};
\node[trt]     (t) at (7.5,2.35) {$t$};
\node[outcome] (y) at (10.0,2.35) {$y$};
\node[unobs]   (u) at (8.75,1.55) {$u$};

\draw[arr] (z) -- (t);
\draw[arr] (t) -- (y);
\draw[arr] (x) -- (z);
\draw[arr] (x) -- (t);
\draw[arr] (x) -- (y);
\draw[hiddenarr] (u) -- (t);
\draw[hiddenarr] (u) -- (y);
\end{scope}

\draw[panel] (0,-8.05) rectangle (9.55,0.78);
\node[paneltitle] at (4.775,0.43) {Candidate context regimes};

\node[regtable] at (2.40,-1.42) {};
\node[regcontent] at (2.40,-1.42) {%
    \begingroup
    \centering
    \setlength{\tabcolsep}{0pt}%
    \renewcommand{\arraystretch}{1}%
    \begin{tabular}{@{}cccc@{}}
        \multicolumn{4}{c}{\regtitle{Observational}} \\
        \regcell{\textcolor{covred}{$\mathbf{X}$}} &
        \regcell{\textcolor{surrogateyellow!70!black}{$\mathbf{Z}$}} &
        \regcell{\textcolor{treatorange!90!black}{$\mathbf{T}$}} &
        \regcell{\textcolor{outblue}{$\mathbf{Y}$}} \\
        \midrule
        \regcell{$0.42$} & \regcell{$-0.31$} & \regcell{$0.67$} & \regcell{$-1.08$} \\
        \regcell{$-0.74$} & \regcell{$0.58$} & \regcell{$0.23$} & \regcell{$0.43$} \\
        \regcell{$\vdots$} & \regcell{$\vdots$} & \regcell{$\vdots$} & \regcell{$\vdots$}
    \end{tabular}%
    \endgroup
};

\node[regtable] at (7.15,-4.60) {};
\node[regcontent] at (7.15,-4.60) {%
    \begingroup
    \centering
    \setlength{\tabcolsep}{0pt}%
    \renewcommand{\arraystretch}{1}%
    \begin{tabular}{@{}cccc@{}}
        \multicolumn{4}{c}{\regtitle{\bfseries $\operatorname{do}(\textcolor{treatorange!90!black}{T}=\textcolor{treatorange!90!black}{t_i})$}} \\
        \regcell{\textcolor{covred}{$\mathbf{X}$}} &
        \regcell{\textcolor{surrogateyellow!70!black}{$\mathbf{Z}$}} &
        \regcell{\textcolor{treatorange!90!black}{$\mathbf{T}$}} &
        \regcell{\textcolor{outblue}{$\mathbf{Y}$}} \\
        \midrule
        \regcell{$0.36$} & \regcell{$0.72$} & \regmark{treatorange!22}{$\mathbf{-1.15}$} & \regcell{$-0.48$} \\
        \regcell{$-0.63$} & \regcell{$-0.19$} & \regmark{treatorange!22}{$\mathbf{0.85}$} & \regcell{$1.27$} \\
        \regcell{$\vdots$} & \regcell{$\vdots$} & \regmark{treatorange!22}{$\vdots$} & \regcell{$\vdots$}
    \end{tabular}%
    \endgroup
};
\node[regtable] at (2.40,-4.60) {};
\node[regcontent] at (2.40,-4.60) {%
    \begingroup
    \centering
    \setlength{\tabcolsep}{0pt}%
    \renewcommand{\arraystretch}{1}%
    \begin{tabular}{@{}cccc@{}}
        \multicolumn{4}{c}{\regtitle{\bfseries $\operatorname{do}(\textcolor{surrogateyellow!70!black}{Z}=\textcolor{surrogateyellow!70!black}{z_i})$}} \\
        \regcell{\textcolor{covred}{$\mathbf{X}$}} &
        \regcell{\textcolor{surrogateyellow!70!black}{$\mathbf{Z}$}} &
        \regcell{\textcolor{treatorange!90!black}{$\mathbf{T}$}} &
        \regcell{\textcolor{outblue}{$\mathbf{Y}$}} \\
        \midrule
        \regcell{$-0.51$} & \regmark{surrogateyellow!20}{$\mathbf{1.40}$} & \regcell{$0.26$} & \regcell{$-1.08$} \\
        \regcell{$0.46$} & \regmark{surrogateyellow!20}{$\mathbf{-0.95}$} & \regcell{$-0.71$} & \regcell{$0.62$} \\
        \regcell{$\vdots$} & \regmark{surrogateyellow!20}{$\vdots$} & \regcell{$\vdots$} & \regcell{$\vdots$}
    \end{tabular}%
    \endgroup
};
\node[regtable] at (7.15,-1.42) {};
\node[regcontent] at (7.15,-1.42) {%
    \begingroup
    \centering
    \setlength{\tabcolsep}{0pt}%
    \renewcommand{\arraystretch}{1}%
    \begin{tabular}{@{}cccc@{}}
        \multicolumn{4}{c}{\regtitle{\bfseries $\operatorname{do}(\textcolor{covred}{X}=\textcolor{covred}{x_i})$}} \\
        \regcell{\textcolor{covred}{$\mathbf{X}$}} &
        \regcell{\textcolor{surrogateyellow!70!black}{$\mathbf{Z}$}} &
        \regcell{\textcolor{treatorange!90!black}{$\mathbf{T}$}} &
        \regcell{\textcolor{outblue}{$\mathbf{Y}$}} \\
        \midrule
        \regmark{covred!18}{$\mathbf{-1.10}$} & \regcell{$0.37$} & \regcell{$-0.54$} & \regcell{$1.45$} \\
        \regmark{covred!18}{$\mathbf{0.80}$} & \regcell{$-0.28$} & \regcell{$0.34$} & \regcell{$-0.52$} \\
        \regmark{covred!18}{$\vdots$} & \regcell{$\vdots$} & \regcell{$\vdots$} & \regcell{$\vdots$}
    \end{tabular}%
    \endgroup
};
\node[note,text width=9.10cm,anchor=north] at (4.775,-6.25)
    {Each table is a candidate source. We primarily study single-target surrogate experiments,
     while a small proportion of tasks omit observational data or include direct
     $\operatorname{do}(T)$ data. In both the context and query data, a new continuous value
     of the intervened variable is sampled independently for every row.};

\draw[panel] (9.95,-8.05) rectangle (15.3,0.78);
\node[paneltitle] at (12.625,0.43) {Query rows};

\node[regtable] at (12.625,-2.91) {};
\node[regcontent] at (12.625,-2.91) {%
    \begingroup
    \centering
    \setlength{\tabcolsep}{0pt}%
    \renewcommand{\arraystretch}{1}%
    \begin{tabular}{@{}cccc@{}}
        \multicolumn{4}{c}{\regtitle{\bfseries $\operatorname{do}(\textcolor{treatorange!90!black}{T}=\textcolor{treatorange!90!black}{t_q})$}} \\
        \regcell{\textcolor{covred}{$\mathbf{X}$}} &
        \regcell{\textcolor{surrogateyellow!70!black}{$\mathbf{Z}$}} &
        \regcell{\textcolor{treatorange!90!black}{$\mathbf{T}$}} &
        \regcell{\textcolor{outblue}{$\mathbf{Y}$}} \\
        \midrule
        \regcell{$-0.25$} & \regcell{$0.41$} & \regmark{treatorange!22}{$\mathbf{0.90}$} & \regmark{outblue!18}{$\mathbf{?}$} \\
        \regcell{$0.65$} & \regcell{$-0.36$} & \regmark{treatorange!22}{$\mathbf{-1.10}$} & \regmark{outblue!18}{$\mathbf{?}$} \\
        \regcell{$\vdots$} & \regcell{$\vdots$} & \regmark{treatorange!22}{$\vdots$} & \regmark{outblue!18}{$\vdots$}
    \end{tabular}%
    \endgroup
};
\node[note,align=left,text width=4.35cm,anchor=north west]
      at (10.45,-4.59)
    {The outcome is masked:\\[2pt]
     \resizebox{4.35cm}{!}{$p_{\theta}\!\bigl(y_q\mid
      \operatorname{do}(T=t_q),x_q,z_q,\regimedatasets\bigr)$}};

\node[note,align=left,text width=4.35cm,anchor=north west] at (10.45,-6.25)
    {Each row supplies $x_q$ and $z_q$, together with a row-specific continuous treatment~level.};

\draw[-{Latex[length=2mm]},thick,dashed,gray!65,rounded corners=2pt]
    (2.6,3.00) -- (0.75,3.00) -- (0.75,0.78);
\draw[-{Latex[length=2mm]},thick,dashed,gray!65,rounded corners=2pt]
    (12.4,3.00) -- (14.65,3.00) -- (14.65,0.78);

\end{tikzpicture}
\endgroup%
    }
    \caption[An example of data fusion from diverse experimental regimes.]{\textbf{Data fusion example:} diverse experimental regimes with continuous, row-wise
interventions. The upper panel shows one possible latent-confounded
data-generating SCM. An actual task contains an observational table and/or
single-target interventional tables, while the right table contains CID
queries. During pre-training, many SCMs are sampled and used to generate finite
context and query data. At inference, only these data are provided; the
underlying SCM remains unknown, and CIDER-FM predicts the CID in a single
forward pass.}
    \label{fig:fusioncfm-task-setting}
\end{figure}

Given this context, the model predicts the conditional interventional distribution (CID)
\begin{equation}
    p_{\theta}\!\left(
        y_i
        \mid
        \doop{T=t_i},
        x_i,
        z_i,
        \regimedatasets
    \right)
    \label{eq:architecture-general-query} 
\end{equation}
for query covariate values \(x_i\) and \(z_i\), and a
continuous treatment assignment \(t_i\). The schematic task in
\cref{fig:fusioncfm-task-setting}
illustrates the available source types and the distinction between context data and
prediction queries.

\paragraph{Context regimes and query.}
Let \(\mathcal{R}\) denote the set of regimes available in a particular task.
Regime \(r\in\mathcal{R}\) contributes
\begin{equation}
    \regimedata{r}
    =
    \left\{
        \mathbf{v}_{r,i}
    \right\}_{i=1}^{n_r},
    \qquad
    \mathbf{v}_{r,i}\in\mathbb{R}^{p}.
    \label{eq:architecture-regime-datasets}
\end{equation}
It is accompanied by an intervention-target mask
\(\mathbf{a}_r\in\{0,1\}^{p}\) and row-specific intervention values
\(\mathbf{c}_{r,i}\in\mathbb{R}^{p}\). Defining the active target set as
\(
    \mathcal{I}_r=\{j:a_{r,j}=1\},
\)
row \(i\) is generated according to
\begin{equation}
    \mathbf{v}_{r,i}
    \sim
    p_{\psi}\!\left(
        \mathbf{V}
        \mid
        \doop{
            \mathbf{V}_{\mathcal{I}_r}
            =
            \mathbf{c}_{r,i,\mathcal{I}_r}
        }
    \right).
    \label{eq:architecture-rowwise-regime}
\end{equation}
The observational regime has \(\mathcal{I}_r=\varnothing\), equivalently
\(\mathbf{a}_r=\mathbf{0}\), so observational and interventional datasets share a
common representation.

Crucially, an interventional regime is defined by its target rather than by one
fixed intervention value. The assigned value is sampled separately for every row.
A direct-treatment query table therefore contains rows generated under
\(
    \doop{T=t_{r,1}},\ldots,\doop{T=t_{r,n_r}},
\)
with generally different continuous values. The same construction is used for surrogate experiments \(\doop{X=x_{r,i}}\). Each
such table consequently covers a range of intervention levels rather than repeated
observations at a single fixed dose.

For each query row \(i\), the model receives the observed
conditioning values \(\mathbf{x}_i\), the continuous assignment \(t_i\), and
query intervention metadata, while \(y_i\) is marked as the prediction target.

\paragraph{Availability of regimes.}
We focus primarily on settings in which causal information is provided by surrogate
experiments. We also sample less
common cases in which observational data are unavailable, as well as cases in
which data from a direct intervention on \(T\) are available. The model therefore
does not assume a fixed number of regimes or a fixed set of intervention targets.

\paragraph{Set-structured input.}
The context is treated as a set of regimes, each of which is itself a set of rows.
Neither the order of regimes nor the order of rows within a regime carries
information. Since the numbers of regimes and rows vary across tasks, inputs are
padded for batching and padding entries are excluded using validity masks.

\subsection{Token representation}
\label{sec:token-representation}

CIDER-FM represents each cell in the table as a token rather than assigning one token to an entire row. This retains the identity and role of each variable while allowing the
model to alternate between variable-wise and sample-wise attention.
A token combines information about the full row, the scalar cell
value, the variable identity, its semantic role (treatment, outcome, conditioning), its cell status (such as target-masked), and the
intervention associated with the regime.

\subsubsection{Task-wise feature normalisation}

Before tokenisation, every variable is standardised using all valid context entries
for the current task. For context row \(i\) in regime \(r\),
let \(v_{r,i,j}\) denote the value of variable \(j\), and let
\(m_{r,i,j}\in\{0,1\}\) indicate whether this entry is valid rather than padding.
The normalisation statistics are
\begin{equation}
\begin{aligned}
    \mu_j
    =
    \frac{
        \sum_{r=1}^{R}\sum_{i=1}^{n_r}
        m_{r,i,j}v_{r,i,j}
    }{
        \sum_{r=1}^{R}\sum_{i=1}^{n_r}m_{r,i,j}
    },\qquad
    \sigma_j^2
    &=
    \frac{
        \sum_{r=1}^{R}\sum_{i=1}^{n_r}
        m_{r,i,j}(v_{r,i,j}-\mu_j)^2
    }{
        \sum_{r=1}^{R}\sum_{i=1}^{n_r}m_{r,i,j}
    },\\
    \widetilde v_{r,i,j}
    &=
    \frac{v_{r,i,j}-\mu_j}{\max(\sigma_j,\epsilon)}.
    \label{eq:architecture-standardisation}
\end{aligned}
\end{equation}
The same \((\mu_j,\sigma_j)\) are applied to context intervention values and
query values. Thus all regimes use a common task-local coordinate system. The
outcome statistics are retained so that the predictive density can later be mapped
back to the original outcome scale.

\subsubsection{Context and query cell tokens}

Let \(d\) denote the hidden dimension. The tokenisation combines complementary
signals so that the model can distinguish cells by their values, variable
identities and roles, intervention status, and surrounding row context. Three independent SwiGLU multilayer
perceptrons \citep{shazeer2020glu} define the continuous-input encoders
\(
    \phi_{\mathrm{row}}:\mathbb{R}^{p_{\max}}\to\mathbb{R}^{d}
\),
\(
    \phi_{\mathrm{val}}:\mathbb{R}\to\mathbb{R}^{d}
\), and
\(
    \phi_{\mathrm{int}}:\mathbb{R}\to\mathbb{R}^{d}
\)
for a complete standardised row, an individual cell value, and an intervention
value, respectively. We use \(\mathbf{e}\) instead for learned lookup embeddings
of discrete metadata, such as a variable's semantic role, identity, and cell
status. For context row \(i\), regime \(r\), and variable \(j\), the
initial token is
\begin{equation}
    \mathbf{h}^{C,0}_{r,i,j}
    =
    \alpha\,\phi_{\mathrm{row}}(\widetilde{\mathbf{v}}_{r,i})
    +
    \beta\,\phi_{\mathrm{val}}(\widetilde v_{r,i,j})
    +
    \mathbf{e}^{\mathrm{role}}_{\rho_j}
    +
    \mathbf{e}^{\mathrm{id}}_j
    +
    \mathbf{e}^{\mathrm{status}}_{\mathrm{obs}}+
    \mathbf{e}^{\mathrm{int}}(a_{r,j},\widetilde c_{r,i,j}),
    \label{eq:architecture-context-token}
\end{equation}
Here, \(C\) denotes a context token and the superscript \(0\) denotes initial
representation; \(Q\) will analogously denote a
query token. The index \(\rho_j\) identifies a variable as treatment, outcome,
conditioning. The embedding
\(\mathbf{e}^{\mathrm{status}}_{\mathrm{obs}}\) marks the context cell value as
observed by the model rather than missing or masked; it does not denote an observational regime.
The learnable scalar coefficients \(\alpha\) and
\(\beta\) balance the row-level and cell-level encodings. The variable embedding \(\mathbf{e}^{\mathrm{id}}_j\)
distinguishes variable slots.

The intervention embedding combines a discrete intervention-target indicator with
its continuous assignment:
\begin{equation}
    \mathbf{e}^{\mathrm{int}}(a,c)
    =
    \mathbf{e}^{\mathrm{int\mbox{-}status}}_{a}
    +
    a\,\phi_{\mathrm{int}}(c).
    \label{eq:architecture-intervention-embedding}
\end{equation}
Here, \(\mathbf{e}^{\mathrm{int\mbox{-}status}}_{a}\) is the vector selected from
a learned two-entry lookup table for \(a\in\{0,1\}\). It distinguishes a
non-target cell from an intervention-target cell, while
\(a\,\phi_{\mathrm{int}}(c)\) records the intervention value only when the variable
is actively intervened upon. The resulting vector
\(\mathbf{e}^{\mathrm{int}}(a,c)\) therefore keeps the intervention target and its
row-specific assignment within a common \(d\)-dimensional representation.

Query tokens follow the same construction. For each query cell,
\(s_{i,j}\) indicates whether its value is observed or masked as the prediction
target. The masked outcome is replaced by zero before standardisation, yielding
the model-visible row \(\widetilde{\mathbf{v}}^{\mathrm{masked}}_i\); its status
embedding marks that entry as a prediction target. For query row \(i\), the
initial token is
\begin{equation}
    \mathbf{h}^{Q,0}_{i,j}
    =
    \alpha\phi_{\mathrm{row}}(\widetilde{\mathbf{v}}^{\mathrm{masked}}_i)
    +
    \beta\phi_{\mathrm{val}}(\widetilde v^{\mathrm{masked}}_{i,j})
    +
    \mathbf{e}^{\mathrm{role}}_{\rho_j}
    +
    \mathbf{e}^{\mathrm{id}}_j
    +
    \mathbf{e}^{\mathrm{status}}_{s_{i,j}}+
    \mathbf{e}^{\mathrm{int}}(a^{Q}_{j},
    \widetilde c^{Q}_{i,j}).
    \label{eq:architecture-query-token}
\end{equation}
Equations~\eqref{eq:architecture-context-token} and
\eqref{eq:architecture-query-token} produce context tokens
\(\mathbf{H}^{C,0}\in
\mathbb{R}^{R_{\max}\times n_{\max}\times p_{\max}\times d}\)
and query tokens
\(\mathbf{H}^{Q,0}\in
\mathbb{R}^{n_q\times p_{\max}\times d}\).

\subsubsection{Intervention metadata residual connection}

CIDER-FM gives every attention layer direct access to the intervention target and
value through a gated residual connection. Let
\(\mathbf{E}^{C,\mathrm{int}}\) and \(\mathbf{E}^{Q,\mathrm{int}}\) denote the
tensors obtained by stacking the per-cell vectors
\(\mathbf{e}^{\mathrm{int}}(a,c)\) across the context and query, respectively.
Let \(L\) denote the number of attention layers. At layer
\(\ell\in\{0,\ldots,L-1\}\), the context and query representations are
\begin{equation}
    \mathbf{H}^{C,\ell}_{\mathrm{meta}}
    =
    \mathbf{H}^{C,\ell}
    +
    \operatorname{sigmoid}(g_C^{\ell})
    W_C^{\ell}
    \operatorname{LN}
    \!\left(\mathbf{E}^{C,\mathrm{int}}\right).
    \label{eq:architecture-context-metadata-residual}
\end{equation}
\begin{equation}
    \mathbf{H}^{Q,\ell}_{\mathrm{meta}}
    =
    \mathbf{H}^{Q,\ell}
    +
    \operatorname{sigmoid}(g_Q^{\ell})
    W_Q^{\ell}
    \operatorname{LN}
    \!\left(\mathbf{E}^{Q,\mathrm{int}}\right).
    \label{eq:architecture-query-metadata-residual}
\end{equation}
Here, \(\operatorname{sigmoid}\) is the logistic sigmoid
and \(\operatorname{LN}\) denotes layer normalisation.
\(g_C^{\ell}\) and \(g_Q^{\ell}\) are learned scalar gate logits, and
\(W_C^{\ell}\) and \(W_Q^{\ell}\) are learned linear projections. Equations~\eqref{eq:architecture-context-metadata-residual}
and~\eqref{eq:architecture-query-metadata-residual} are written for valid cells;
padded positions are reset to zero after each residual update. Repeatedly injecting this metadata prevents intervention semantics
from being diluted across layers and allows otherwise similar values to be
interpreted differently across causal regimes.

\begin{algorithm}[H]
\small
\caption{Hierarchical Three-Axis Attention Stack}
\label{alg:three-axis-attention}
\begin{algorithmic}[1]
\Statex \makebox[5em][l]{\textbf{Require:}}Initial tokens
    \(\mathbf{H}^{C,0}\), \(\mathbf{H}^{Q,0}\)
\Statex \makebox[5em][l]{}intervention metadata
    \(\mathbf{E}^{C,\mathrm{int}}\), \(\mathbf{E}^{Q,\mathrm{int}}\)
\Statex \makebox[5em][l]{}Layer count \(L\)
\Statex \makebox[5em][l]{\textbf{Ensure:}}Updated tokens
    \(\mathbf{H}^{C,L}\), \(\mathbf{H}^{Q,L}\)
\For{\(\ell=0,\ldots,L-1\)}
    \State \(\mathbf{H}^{C,\ell}_{\mathrm{meta}}\gets
        \mathbf{H}^{C,\ell}
        +\operatorname{sigmoid}(g_C^{\ell})W_C^{\ell}
        \operatorname{LN}(\mathbf{E}^{C,\mathrm{int}})\)
        \Comment{interventional metadata residual connection}
    \State \(\mathbf{H}^{Q,\ell}_{\mathrm{meta}}\gets
        \mathbf{H}^{Q,\ell}
        +\operatorname{sigmoid}(g_Q^{\ell})W_Q^{\ell}
        \operatorname{LN}(\mathbf{E}^{Q,\mathrm{int}})\)
    \State \(\mathbf{H}^{C,\ell}_{\mathrm{var}}\gets
        \mathcal{A}^{\ell}_{V}(\mathbf{H}^{C,\ell}_{\mathrm{meta}})\)
        \Comment{variable axis; row-wise}
    \State \(\mathbf{H}^{Q,\ell}_{\mathrm{var}}\gets
        \mathcal{A}^{\ell}_{V}(\mathbf{H}^{Q,\ell}_{\mathrm{meta}})\)
        \Comment{shared block; row-wise}
    \For{each valid regime--variable pair \((r,j)\)}
        \State \(\left(\mathbf{H}^{C,\ell}_{\mathrm{samp}}\right)_{r,:,j,:}
            \gets \mathcal{A}^{\ell}_{S}
            \!\left(\left(\mathbf{H}^{C,\ell}_{\mathrm{var}}\right)_{r,:,j,:}\right)\)
            \Comment{sample axis}
        \State \(\left(\mathbf{M}^{\ell}_{\mathrm{local}}\right)_{r,:,j,:}
            \gets \operatorname{PMA}^{\ell}
            \!\left(\mathbf{Q}^{\ell}_{\mathrm{mem}},
            \left(\mathbf{H}^{C,\ell}_{\mathrm{samp}}\right)_{r,:,j,:}\right)\)
    \EndFor
    \For{each valid variable \(j\)}
        \State \(\left(\mathbf{M}^{\ell}_{\mathrm{fused}}\right)_{:,:,j,:}
            \gets \mathcal{A}^{\ell}_{R}
            \!\left(\left(\mathbf{M}^{\ell}_{\mathrm{local}}\right)_{:,:,j,:}\right)\)
            \Comment{regime-axis attention over memories}
        \State \(\left(\mathbf{H}^{C,\ell+1}\right)_{:,:,j,:}
            \gets \mathcal{A}^{\ell}_{C\leftarrow M}
            \!\left(\left(\mathbf{H}^{C,\ell}_{\mathrm{samp}}\right)_{:,:,j,:},
            \left(\mathbf{M}^{\ell}_{\mathrm{fused}}\right)_{:,:,j,:}\right)\)
            \Comment{context feedback}
        \State \(\left(\mathbf{H}^{Q,\ell+1}\right)_{:,j,:}
            \gets \mathcal{A}^{\ell}_{Q\leftarrow M}
            \!\left(\left(\mathbf{H}^{Q,\ell}_{\mathrm{var}}\right)_{:,j,:},
            \left(\mathbf{M}^{\ell}_{\mathrm{fused}}\right)_{:,:,j,:}\right)\)
            \Comment{query reads only}
    \EndFor
\EndFor
\State \Return \(\mathbf{H}^{C,L},\mathbf{H}^{Q,L}\)
\end{algorithmic}
\end{algorithm}

\subsection{Hierarchical three-axis attention}
\label{sec:hierarchical-attention}

A dense attention block over all cells forms a pairwise attention matrix whose
computational and memory costs grow quadratically with the number of tokens.
Unlike existing tabular foundation models, CIDER-FM explicitly introduces the experimental
regime as a third axis. These operations are repeated for \(L\) layers, with intervention
metadata supplied at every layer and fused memories returned to both context and
query tokens.
\cref{alg:three-axis-attention} summarises the resulting attention stack; the
following subsections describe its components.

We use \(\mathbf{H}\) for cell-token states,
\(\mathbf{M}_{\mathrm{local}}\) and \(\mathbf{M}_{\mathrm{fused}}\) for memory
tokens, and \(\mathbf{Q}^{\ell}_{\mathrm{mem}}\in\mathbb{R}^{K\times d}\) for
the \(K\) learned pooling queries. The blocks
\(\mathcal{A}^{\ell}_{V}\), \(\mathcal{A}^{\ell}_{S}\), and
\(\mathcal{A}^{\ell}_{R}\) operate along the three axes, while
\(\mathcal{A}^{\ell}_{C\leftarrow M}\) and
\(\mathcal{A}^{\ell}_{Q\leftarrow M}\) perform memory feedback. Batch dimensions,
validity masks, and reshaping are omitted for clarity.

\subsubsection{Variable-axis attention}

The first operation exchanges information across variables independently within
each context or query row:
\begin{equation}
\begin{aligned}
    \left(\mathbf{H}^{C,\ell}_{\mathrm{var}}\right)_{r,i,:,:}
    &=
    \mathcal{A}^{\ell}_{V}
    \!\left(
        \left(\mathbf{H}^{C,\ell}_{\mathrm{meta}}\right)_{r,i,:,:}
    \right),
    \\
    \left(\mathbf{H}^{Q,\ell}_{\mathrm{var}}\right)_{i,:,:}
    &=
    \mathcal{A}^{\ell}_{V}
    \!\left(
        \left(\mathbf{H}^{Q,\ell}_{\mathrm{meta}}\right)_{i,:,:}
    \right).
    \label{eq:architecture-variable-axis}
\end{aligned}
\end{equation}
This operation allows every variable token to incorporate information from the other
variables in the same row. Context and query use the same variable-axis block, but
each row forms a separate attention sequence.

\subsubsection{Within-regime sample attention}

The second operation exchanges information across context samples separately for
each regime and variable:
\begin{equation}
    \left(\mathbf{H}^{C,\ell}_{\mathrm{samp}}\right)_{r,:,j,:}
    =
    \mathcal{A}^{\ell}_{S}
    \!\left(
        \left(\mathbf{H}^{C,\ell}_{\mathrm{var}}\right)_{r,:,j,:}
    \right),
    \qquad
    r=1,\ldots,R,
    \quad
    j=1,\ldots,p.
    \label{eq:architecture-sample-axis}
\end{equation}
As no sample positional embeddings are used and the same attention projections
are applied to every row, sample-axis attention is permutation equivariant within
each regime. Regimes are processed independently during this step. Query rows
bypass this operation so each
prediction depends only on its own query inputs and the shared context.

\subsubsection{Per-variable memories and regime-axis attention}

CIDER-FM compresses each \((\text{regime},\text{variable})\) sample set into
\(K\) memory tokens before exchanging information across regimes. Let
\[
    \mathbf{Q}^{\ell}_{\mathrm{mem}}
    =
    \begin{bmatrix}
        \left(\mathbf{q}^{\ell}_{\mathrm{mem},1}\right)^{\!\top}\\
        \vdots\\
        \left(\mathbf{q}^{\ell}_{\mathrm{mem},K}\right)^{\!\top}
    \end{bmatrix}
    \in\mathbb{R}^{K\times d},
    \qquad
    \mathbf{q}^{\ell}_{\mathrm{mem},k}\in\mathbb{R}^{d}
\]
denote the layer-specific learned pooling queries. They are shared across all
regimes and variables within layer \(\ell\). 

Pooling by Multihead Attention (PMA)
uses these vectors as queries in a residual cross-attention block, followed by a
residual feed-forward update. It gives
\begin{equation}
    \left(\mathbf{M}^{\ell}_{\mathrm{local}}\right)_{r,:,j,:}
    =
    \operatorname{PMA}^{\ell}
    \!\left(
        \mathbf{Q}^{\ell}_{\mathrm{mem}},
        \left(\mathbf{H}^{C,\ell}_{\mathrm{samp}}\right)_{r,:,j,:}
    \right)
    \in
    \mathbb{R}^{K\times d}.
    \label{eq:architecture-memory-pooling}
\end{equation}
Here, the sample tokens \(\mathbf{H}^{C,\ell}_{\mathrm{samp}}\) provide the keys
and values. Collecting the outputs over all regimes and variables yields
\(\mathbf{M}^{\ell}_{\mathrm{local}}
\in\mathbb{R}^{R\times K\times p\times d}\), remaining invariance under permutations of samples within a regime.

For a fixed variable \(j\), there are \(K\) local memories from each of the
\(R\) regimes, giving \(R\times K\) memory tokens in total. Regime-axis
self-attention processes these tokens jointly:
\begin{equation}
    \left(\mathbf{M}^{\ell}_{\mathrm{fused}}\right)_{:,:,j,:}
    =
    \mathcal{A}^{\ell}_{R}
    \!\left(
        \left(\mathbf{M}^{\ell}_{\mathrm{local}}\right)_{:,:,j,:}
    \right)
    \in\mathbb{R}^{R\times K\times d}.
    \label{eq:architecture-regime-axis}
\end{equation}
Across all variables,
\(\mathbf{M}^{\ell}_{\mathrm{fused}}
\in\mathbb{R}^{R\times K\times p\times d}\) and is equivariant to
permutations of the regimes. Regime-axis attention is applied separately for each variable, while
cross-variable interactions are handled by variable-axis attention.

\subsubsection{Memory feedback to context and query}

For each variable \(j\), the fused memories update the context and query tokens
through separate cross-attention blocks:
\begin{equation}
\begin{aligned}
    \left(\mathbf{H}^{C,\ell+1}\right)_{:,:,j,:}
    &=
    \mathcal{A}^{\ell}_{C\leftarrow M}
    \!\left(
        \left(\mathbf{H}^{C,\ell}_{\mathrm{samp}}\right)_{:,:,j,:},
        \left(\mathbf{M}^{\ell}_{\mathrm{fused}}\right)_{:,:,j,:}
    \right),
    \\
    \left(\mathbf{H}^{Q,\ell+1}\right)_{:,j,:}
    &=
    \mathcal{A}^{\ell}_{Q\leftarrow M}
    \!\left(
        \left(\mathbf{H}^{Q,\ell}_{\mathrm{var}}\right)_{:,j,:},
        \left(\mathbf{M}^{\ell}_{\mathrm{fused}}\right)_{:,:,j,:}
    \right).
    \label{eq:architecture-memory-feedback}
\end{aligned}
\end{equation}
In both blocks, the cell tokens provide the queries and the fused memories provide
the keys and values. The updated context tokens enter the next layer, allowing
cross-regime information to be refined iteratively. Query tokens read from the
memories but do not contribute to their construction.

\subsubsection{Computational complexity}
Let \(N=\sum_{r=1}^{R}n_r\) be the total number of context rows. Ignoring constant
factors and the hidden dimension, the principal attention cost per layer is
\begin{equation}
    \mathcal{O}\!\Biggl(
        \underbrace{\vphantom{\displaystyle p\sum_{r=1}^{R}n_r^2}(N+n_q)p^2}_{\text{variable axis}}
        +
        \underbrace{p\sum_{r=1}^{R}n_r^2}_{\text{sample axis}}
        +
        \underbrace{\vphantom{\displaystyle p\sum_{r=1}^{R}n_r^2}pKN}_{\text{memory pooling}}
        +
        \underbrace{\vphantom{\displaystyle p\sum_{r=1}^{R}n_r^2}p(RK)^2}_{\text{regime axis}}
        +
        \underbrace{\vphantom{\displaystyle p\sum_{r=1}^{R}n_r^2}p(N+n_q)RK}_{\text{memory feedback}}
    \Biggr).
    \label{eq:architecture-complexity}
\end{equation}
This factorisation preserves the permutation symmetries induced by the
set-structured geometry of the input, while avoiding dense attention over all
cells. The computational profiling in
\cref{sec:ablation-studies} shows that the relative efficiency of
three-axis attention improves as multi-regime contexts grow.
In the tested configuration, it achieves higher inference and training
throughput than conventional alternating attention for 2-D tables at
1024 and 2048 rows per regime, under both same-width and
parameter-matched comparisons. These results motivate three-axis
attention as a structured and computationally attractive approach
for scaling to larger multi-regime contexts.

\subsection{Predictive distribution and learning objective}
\label{sec:architecture-output}

After the final layer, the model selects the query token corresponding to the designated
outcome variable. A linear readout maps this token to the parameters of a full-support
bar distribution. With \(J\) interior bars, the head emits \(J+2\) mixture logits---one
for each bar and one for each tail---together with two positive tail-scale parameters.
The density is piecewise uniform over fixed intervals in standardised outcome space and
uses half-Gaussian tails beyond the outer edges. The
experiments in this work use \(J=32\) bars.

Let \(z=(y-\mu_Y)/\sigma_Y\) denote the standardised outcome and let
\(p^{\mathrm{bar}}_{\theta}(z\mid\cdot)\) be the density represented by the output
head. The corresponding density on the original outcome scale is
\begin{equation}
    p_{\theta}(y\mid\cdot)
    =
    \frac{1}{\sigma_Y}
    p^{\mathrm{bar}}_{\theta}\!\left(
        \frac{y-\mu_Y}{\sigma_Y}
        \biggm|\cdot
    \right).
    \label{eq:architecture-original-scale-density}
\end{equation}
For a minibatch containing \(B\) tasks and \(n_q\) queries per task, training minimises
the original-scale negative log-likelihood
\begin{equation}
    \mathcal{L}(\theta)
    =
    -\frac{1}{Bn_q}
    \sum_{b=1}^{B}
    \sum_{i=1}^{n_q}
    \log
    p_{\theta}\!\left(
        y_{b,i}
        \mid
        \mathcal{C}_b,
        \mathcal{Q}_{b,i}
    \right),
    \label{eq:architecture-training-objective}
\end{equation}
where \(\mathcal{C}_b\) denotes the multi-regime context and
\(\mathcal{Q}_{b,i}\) denotes the intervention query. The objective uses only the
sampled query outcome: oracle conditional means or graph-derived labels do not enter
the loss. Because the network predicts a full density, conditional means, uncertainty
intervals, quantiles, and samples can all be obtained from the same output distribution.

\section{Baselines and metrics}
\label{app:baselines-metrics}

\begin{table}[htbp]
\centering
\caption{Methods and context configurations used across the
synthetic and real-world experiments.}
\label{tab:restricted_lingauss_baselines}

\begingroup
\scriptsize
\setlength{\tabcolsep}{3pt}
\renewcommand{\arraystretch}{0.98}

\begin{tabularx}{\textwidth}{
    @{}
    >{\raggedright\arraybackslash}p{0.27\textwidth}
    >{\raggedright\arraybackslash}p{0.16\textwidth}
    >{\raggedright\arraybackslash}p{0.31\textwidth}
    >{\raggedright\arraybackslash}X
    @{}
}
\toprule
Full method name
& Context policy
& Data supplied
& Causal information supplied \\
\midrule

\multicolumn{4}{@{}l}{\bfseries
CIDER-FM and observational controls} \\[2pt]
CIDER-FM
& \texttt{fusion}
& \(N_r\) obs rows and \((R-1)N_r\) int rows
& Intervention targets and values \\
CIDER-FM
& \texttt{obs\_total}
& \(RN_r\) obs rows
& None \\
CIDER-FM-Obs
& \texttt{obs\_total}
& \(RN_r\) obs rows
& None \\
CIDER-FM-Obs
& \texttt{obs\_fixed}
& \(N_r\) obs rows
& None \\

\midrule
\multicolumn{4}{@{}l}{\bfseries Predictive methods} \\[2pt]
Linear regression
& \texttt{obs\_total}
& \(RN_r\) obs rows
& None \\
Linear regression
& \texttt{fusion}
& \(N_r\) obs rows and \((R-1)N_r\) int rows, pooled
& None \\
Random forest
& \texttt{obs\_total}
& \(RN_r\) obs rows
& None \\
Random forest
& \texttt{fusion}
& \(N_r\) obs rows and \((R-1)N_r\) int rows, pooled
& None \\
TabPFN v2
& \texttt{obs\_total}
& \(RN_r\) obs rows
& None \\
TabPFN v2
& \texttt{fusion}
& \(N_r\) obs rows and \((R-1)N_r\) int rows, pooled
& None \\
TabPFN v2
& \texttt{obs\_fixed}
& \(N_r\) obs rows
& None \\
Bayesian linear regression
& \texttt{obs\_total}
& \(RN_r\) obs rows
& None \\
Bayesian linear regression
& \texttt{fusion}
& \(N_r\) obs rows and \((R-1)N_r\) int rows, pooled
& None \\

\midrule
\multicolumn{4}{@{}l}{\bfseries
Additional causal methods} \\[2pt]
Partially linear DML
& \texttt{obs\_total}
& \(RN_r\) obs rows
& None \\
Partially linear DML
& \texttt{obs\_fixed}
& \(N_r\) obs rows
& None \\
Do-PFN
& \texttt{obs\_total}
& \(RN_r\) obs rows
& None \\
Do-PFN
& \texttt{obs\_fixed}
& \(N_r\) obs rows
& None \\
ArCO-GP
& \texttt{fusion}
& \(N_r\) obs rows and \((R-1)N_r\) int rows
& Intervention targets and values \\

\midrule
\multicolumn{4}{@{}l}{\bfseries Graph4CFM variants} \\[2pt]
Graph4CFM
& \texttt{obs\_total}
& \(RN_r\) obs rows
& None \\
Graph4CFM
& \texttt{obs\_total}
& \(RN_r\) obs rows
& Oracle ancestry into \(Y\) \\
Graph4CFM
& \texttt{obs\_total}
& \(RN_r\) obs rows
& Oracle all-pairs directed ancestry \\
Graph4CFM
& \texttt{obs\_total}
& \(RN_r\) obs rows
& \(\mathrm{red}\to Y\) \\
Graph4CFM
& \texttt{obs\_total}
& \(RN_r\) obs rows
& \(\mathrm{red},\mathrm{green},\mathrm{blue}\to Y\) \\

\midrule
\multicolumn{4}{@{}l}{\bfseries
Graph-informed references (excluded from rankings)} \\[2pt]
\(c\text{-}g\text{ID}\) + linear regression
& \texttt{obs\_total}
& \(RN_r\) obs rows
& Ground-truth ADMG \\
\(c\text{-}g\text{ID}\) + linear regression
& \texttt{fusion}
& \(N_r\) obs rows and \((R-1)N_r\) int rows
& Ground-truth ADMG \\
\(c\text{-}g\text{ID}\) + Bayesian linear regression
& \texttt{obs\_total}
& \(RN_r\) obs rows
& Ground-truth ADMG \\
\(c\text{-}g\text{ID}\) + Bayesian linear regression
& \texttt{fusion}
& \(N_r\) obs rows and \((R-1)N_r\) int rows
& Ground-truth ADMG \\

\bottomrule
\end{tabularx}

\par\smallskip
\begin{minipage}{\textwidth}
\scriptsize
\textit{Protocol.}
\(R\) denotes the total number of regimes, and \(N_r\) denotes the number of rows
per regime under equal allocation.
Obs and int denote observational and interventional data.
Predictive baselines pool observational and interventional
rows without labels indicating their experimental regimes
or intervention targets.
The CIDER-FM \texttt{obs\_total} control uses the
fusion-trained checkpoint.
Graph4CFM receives ancestral relations rather than a full
adjacency graph; arrows denote ancestry.
The \(c\text{-}g\text{ID}\) references are evaluated only
when the query is identifiable from the supplied sources
under the ground-truth ADMG.
\end{minipage}

\endgroup
\end{table}

\paragraph{Comparison with causal foundation models.}
We include Do-PFN \citep{robertson2026pfn} as an external
causal foundation model using observational context.
Its released checkpoint is trained with binary treatments,
whereas our benchmarks use continuous treatment values.
We pass these values directly to its prediction interface
without discretisation or retraining, and report its
predictive mean. Consequently, this comparison evaluates
Do-PFN under treatment-domain misspecification.
We also evaluate Graph4CFM \citep{reuter2026use} with
varying amounts of ancestral information.
Its synthetic training prior assumes causal sufficiency,
excluding hidden confounders.
Graph4CFM is therefore evaluated under prior misspecification
on tasks containing hidden confounding, even when supplied
with correct directed ancestry.
We do not include CausalPFN \citep{balazadeh2026causalpfn}
or CausalFM \citep{ma2025foundation}, whose released
implementations address different estimation targets.
CausalPFN estimates conditional expected potential outcomes
for binary treatments under strong ignorability.
CausalFM's released models target binary-treatment CATE
estimation under specified back-door, front-door, or
instrumental-variable assumptions.
These implementations do not directly provide the conditional
outcome distribution
\(p(y\mid\doop{T=t},\mathbf{x},\mathcal{D})\)
for continuous \(t\) across our graph families with hidden
confounding. A comparison would therefore require adapting
their estimation targets and training setups, beyond applying
the published checkpoints.

Given these prior and task mismatches, we train CIDER-FM-Obs
using the same SCM prior and prediction task as CIDER-FM
but only observational context, providing our primary matched
baseline for assessing the benefit of interventional data.

\paragraph{Metrics.}
In the main text, we evaluate predictions against sampled
query outcomes \(Y_q\), reporting negative log-likelihood
(NLL) for the predictive density and MSE and \(R^2\) for
its predictive mean.
In the appendix, we additionally report MSE and \(R^2\)
against the generating SCM's conditional interventional mean,
\(Y_{\mathrm{mean}}
=\mathbb{E}_{\psi}[Y\mid\doop{T=t},\mathbf{x}]\),
where this quantity can be computed reliably.
These additional metrics assess conditional-mean prediction
without the outcome sampling noise present in \(Y_q\).

\section{Stratified results on restricted graphs}
\label{app:restricted}

\subsection{Experimental protocol}

\begin{algorithm}[H]
\small
\caption{Restricted Graph-Family Sampling}
\label{alg:restricted-family}
\begin{algorithmic}[1]
\Statex \makebox[5em][l]{\textbf{Require:}}
    Node-count range \([N_{\min},N_{\max}]\), optional target class \(c^\star\)
\Statex \makebox[5em][l]{\textbf{Ensure:}}
    ADMG \(G\), available data regimes \(\mathcal{S}\), class \(c\)

\Repeat
    \State Sample
        \(N\sim\operatorname{Uniform}\{N_{\min},\ldots,N_{\max}\}\)
        and set \(V=\{Z,T,Y,X_0,\ldots,X_{N-4}\}\).
    \State Uniformly sample one of the five directed cores in which
        \(Y\) is a descendant of both \(Z\) and \(T\).
    \For{\((A,B)\in\{(Z,T),(Z,Y),(T,Y)\}\)}
        \State Add \(A\leftrightarrow B\) if an independent
        \(\operatorname{Bernoulli}(0.5)\) draw succeeds.
        \Comment{Hidden confounding}
    \EndFor
    \State Add \(X_0\to Z\), \(X_0\to T\), and \(X_0\to Y\).
    \For{\(i=1,\ldots,N-4\)}
        \For{\(W\in\{Z,T,Y\}\)}
            \State Add \(X_i\to W\) if an independent
            \(\operatorname{Bernoulli}(0.5)\) draw succeeds.
        \EndFor
    \EndFor
    \State Initialise the available regimes as
        \[
            \mathcal{S}
            \gets
            \left\{
                P(V),
                P\!\left(V\setminus\{Z\}\mid do(Z)\right)
            \right\}.
        \]
    \If{an independent \(\operatorname{Bernoulli}(0.5)\) draw succeeds}
        \State Uniformly select \(X_j\in\{X_0,\ldots,X_{N-4}\}\) and set
        \[
            \mathcal{S}
            \gets
            \mathcal{S}
            \cup
            \left\{
                P\!\left(V\setminus\{X_j\}\mid do(X_j)\right)
            \right\}.
        \]
    \EndIf
    \State Classify
        \(P(Y\mid do(T),V\setminus\{T,Y\})\) using \(c\)-\(g\)ID.
\Until{\(c^\star\) is unspecified or the sampled task belongs to \(c^\star\)}
\State \Return \(G,\mathcal{S},c\)
\end{algorithmic}
\end{algorithm}

We sample graphs with \(4\)--\(10\) nodes from five directed cores in which
\(Y\) is a descendant of both \(Z\) and \(T\) as shown in \cref{alg:restricted-family}. Bidirected edges between
\(Z\), \(T\), and \(Y\) introduce hidden confounding, while additional
variables act as observed covariates. Each task contains one observational
regime and \(K\in\{0,1,2\}\) interventional regimes. For \(K=0\), the
context is observational only; for \(K=1\), it also includes \(do(Z)\);
for \(K=2\), it additionally includes an experiment on a randomly chosen
covariate \(X_j\). Both context and query interventions use continuous assignments, with
\(z_i\) in \(\doop{Z=z_i}\) and \(t_i\) in \(\doop{T=t_i}\) varying across rows. We instantiate the graphs using either linear--Gaussian or
nonlinear--Gaussian mechanisms and assign each task to one of the three cases
in Section~\ref{sec:theoretical-benefits} using \(c\)-\(g\)ID.

Each regime contains \(N_r=512\) rows. We evaluate the predictive distribution
of \(Y_q\) using NLL and its posterior mean using MSE and \(R^2\); MSE and
\(R^2\) are also reported for the ground-truth conditional causal mean
\(Y_{\mathrm{mean}}\). Table~\ref{tab:restricted_lingauss_baselines} summarises
the comparison methods and the data and causal information supplied to each.
The \texttt{obs\_total} variants match the total row budget of the fusion
methods, whereas \texttt{obs\_fixed} variants use only the observational
portion. Graph-informed methods receive the ground-truth ADMG, and the SCM
oracle is included only as a reference. All model training and evaluation experiments were conducted on NVIDIA A100
GPUs.

\subsection{Stratified results}

Tables~\ref{tab:restricted-lg-validation-ranking}
and~\ref{tab:restricted-complex-validation-ranking} report the full results,
including means, standard errors, and CIDER-FM's rank for each setting and
metric. CIDER-FM is competitive across all three settings and is particularly
strong on NLL and \(R^2\). Under nonlinear--Gaussian mechanisms, it achieves
the best NLL in every setting and the best \(R^2\) for both targets in the two
data-fusion settings. Raw MSE is less uniform, with graph-informed estimators
or pooled TabPFN performing best in some cases.

\begin{table}[htbp]
\centering
\caption{Restricted-family linear--Gaussian detailed ranking.
Values are mean \(\pm\) standard error. Rank gives the position of
CIDER-FM among 19 methods, excluding all \(c\text{-}g\text{ID}\) methods.}
\label{tab:restricted-lg-validation-ranking}

\begingroup
\scriptsize
\setlength{\tabcolsep}{1.8pt}
\renewcommand{\arraystretch}{1.12}

\begin{tabular}{
    @{}
    >{\raggedright\arraybackslash}m{0.130\textwidth}
    >{\centering\arraybackslash}m{0.075\textwidth}
    >{\centering\arraybackslash}m{0.065\textwidth}
    >{\centering\arraybackslash}m{0.175\textwidth}
    >{\raggedleft\arraybackslash}m{0.025\textwidth}
    @{\,/\,}
    >{\raggedright\arraybackslash}m{0.032\textwidth}
    >{\raggedright\arraybackslash}m{0.235\textwidth}
    >{\centering\arraybackslash}m{0.175\textwidth}
    @{}
}
\toprule
\textbf{Setting}
& \textbf{Target}
& \textbf{Metric}
& \textbf{CIDER-FM mean \(\boldsymbol{\pm}\) SE}
& \multicolumn{2}{c}{\textbf{Rank}}
& \textbf{Best method}
& \textbf{Best mean \(\boldsymbol{\pm}\) SE}
\\
\midrule

\multirow{5}{=}{\texttt{obs\_id}}
& \multirow{3}{=}{\(Y_q\)}
& NLL
& 0.9456 \(\pm\) 0.0440
& 2 & 11
& CIDER-FM-Obs (\texttt{obs\_total})
& 0.9338 \(\pm\) 0.0394
\\
& & MSE
& 0.9097 \(\pm\) 0.0922
& 3 & 19
& CIDER-FM-Obs (\texttt{obs\_total})
& 0.8679 \(\pm\) 0.0766
\\
& & \(R^2\)
& \textbf{0.7330 \(\pm\) 0.0174}
& \textbf{1} & \textbf{19}
& \textbf{CIDER-FM}
& \textbf{0.7330 \(\pm\) 0.0174}
\\
\cmidrule(lr){2-8}
& \multirow{2}{=}{\(Y_{\mathrm{mean}}\)}
& MSE
& 0.4131 \(\pm\) 0.0769
& 3 & 19
& CIDER-FM-Obs (\texttt{obs\_total})
& 0.3711 \(\pm\) 0.0592
\\
& & \(R^2\)
& \textbf{0.8498 \(\pm\) 0.0241}
& \textbf{1} & \textbf{19}
& \textbf{CIDER-FM}
& \textbf{0.8498 \(\pm\) 0.0241}
\\

\midrule
\multirow{5}{=}{\texttt{fusion\_id}}
& \multirow{3}{=}{\(Y_q\)}
& NLL
& \textbf{1.6150 \(\pm\) 0.0403}
& \textbf{1} & \textbf{11}
& \textbf{CIDER-FM}
& \textbf{1.6150 \(\pm\) 0.0403}
\\
& & MSE
& \textbf{1.8332 \(\pm\) 0.1013}
& \textbf{1} & \textbf{19}
& \textbf{CIDER-FM}
& \textbf{1.8332 \(\pm\) 0.1013}
\\
& & \(R^2\)
& \textbf{0.6006 \(\pm\) 0.0188}
& \textbf{1} & \textbf{19}
& \textbf{CIDER-FM}
& \textbf{0.6006 \(\pm\) 0.0188}
\\
\cmidrule(lr){2-8}
& \multirow{2}{=}{\(Y_{\mathrm{mean}}\)}
& MSE
& \textbf{0.6746 \(\pm\) 0.0585}
& \textbf{1} & \textbf{19}
& \textbf{CIDER-FM}
& \textbf{0.6746 \(\pm\) 0.0585}
\\
& & \(R^2\)
& 0.7089 \(\pm\) 0.0398
& 3 & 19
& RF (\texttt{fusion})
& 0.7212 \(\pm\) 0.0186
\\

\midrule
\multirow{5}{=}{\texttt{fusion\_non\_id}}
& \multirow{3}{=}{\(Y_q\)}
& NLL
& \textbf{1.7229 \(\pm\) 0.0391}
& \textbf{1} & \textbf{11}
& \textbf{CIDER-FM}
& \textbf{1.7229 \(\pm\) 0.0391}
\\
& & MSE
& 2.8267 \(\pm\) 0.3385
& 2 & 19
& CIDER-FM-Obs (\texttt{obs\_total})
& 2.7895 \(\pm\) 0.2367
\\
& & \(R^2\)
& \textbf{0.5428 \(\pm\) 0.0345}
& \textbf{1} & \textbf{19}
& \textbf{CIDER-FM}
& \textbf{0.5428 \(\pm\) 0.0345}
\\
\cmidrule(lr){2-8}
& \multirow{2}{=}{\(Y_{\mathrm{mean}}\)}
& MSE
& 1.5011 \(\pm\) 0.3232
& 2 & 19
& CIDER-FM-Obs (\texttt{obs\_total})
& 1.4586 \(\pm\) 0.2111
\\
& & \(R^2\)
& \textbf{0.6243 \(\pm\) 0.0551}
& \textbf{1} & \textbf{19}
& \textbf{CIDER-FM}
& \textbf{0.6243 \(\pm\) 0.0551}
\\

\bottomrule
\end{tabular}
\endgroup
\end{table}

\begin{table}[htbp]
\centering
\caption{Restricted-family nonlinear--Gaussian detailed ranking.
Values are mean \(\pm\) standard error. Rank gives the position of
CIDER-FM among 19 methods, excluding all \(c\text{-}g\text{ID}\) methods.}
\label{tab:restricted-complex-validation-ranking}

\begingroup
\scriptsize
\setlength{\tabcolsep}{1.8pt}
\renewcommand{\arraystretch}{1.12}

\begin{tabular}{
    @{}
    >{\raggedright\arraybackslash}m{0.130\textwidth}
    >{\centering\arraybackslash}m{0.075\textwidth}
    >{\centering\arraybackslash}m{0.065\textwidth}
    >{\centering\arraybackslash}m{0.175\textwidth}
    >{\raggedleft\arraybackslash}m{0.025\textwidth}
    @{\,/\,}
    >{\raggedright\arraybackslash}m{0.032\textwidth}
    >{\raggedright\arraybackslash}m{0.235\textwidth}
    >{\centering\arraybackslash}m{0.175\textwidth}
    @{}
}
\toprule
\textbf{Setting}
& \textbf{Target}
& \textbf{Metric}
& \textbf{CIDER-FM mean \(\boldsymbol{\pm}\) SE}
& \multicolumn{2}{c}{\textbf{Rank}}
& \textbf{Best method}
& \textbf{Best mean \(\boldsymbol{\pm}\) SE}
\\
\midrule

\multirow{5}{=}{\texttt{obs\_id}}
& \multirow{3}{=}{\(Y_q\)}
& NLL
& \textbf{0.9537 \(\pm\) 0.0323}
& \textbf{1} & \textbf{11}
& \textbf{CIDER-FM}
& \textbf{0.9537 \(\pm\) 0.0323}
\\
& & MSE
& 1.5096 \(\pm\) 0.2347
& 3 & 19
& TabPFN v2 (\texttt{obs\_total})
& 1.3793 \(\pm\) 0.2732
\\
& & \(R^2\)
& 0.3737 \(\pm\) 0.0154
& 2 & 19
& TabPFN v2 (\texttt{obs\_total})
& 0.3936 \(\pm\) 0.0156
\\
\cmidrule(lr){2-8}
& \multirow{2}{=}{\(Y_{\mathrm{mean}}\)}
& MSE
& 1.1354 \(\pm\) 0.2335
& 3 & 19
& TabPFN v2 (\texttt{obs\_total})
& 1.0021 \(\pm\) 0.2722
\\
& & \(R^2\)
& 0.5057 \(\pm\) 0.0176
& 2 & 19
& TabPFN v2 (\texttt{obs\_total})
& 0.5299 \(\pm\) 0.0182
\\

\midrule
\multirow{5}{=}{\texttt{fusion\_id}}
& \multirow{3}{=}{\(Y_q\)}
& NLL
& \textbf{1.3479 \(\pm\) 0.0306}
& \textbf{1} & \textbf{11}
& \textbf{CIDER-FM}
& \textbf{1.3479 \(\pm\) 0.0306}
\\
& & MSE
& 3.5265 \(\pm\) 0.7781
& 3 & 19
& TabPFN v2 (\texttt{fusion})
& 3.0792 \(\pm\) 0.4604
\\
& & \(R^2\)
& \textbf{0.4109 \(\pm\) 0.0169}
& \textbf{1} & \textbf{19}
& \textbf{CIDER-FM}
& \textbf{0.4109 \(\pm\) 0.0169}
\\
\cmidrule(lr){2-8}
& \multirow{2}{=}{\(Y_{\mathrm{mean}}\)}
& MSE
& 2.7177 \(\pm\) 0.7732
& 3 & 19
& TabPFN v2 (\texttt{fusion})
& 2.2722 \(\pm\) 0.4526
\\
& & \(R^2\)
& \textbf{0.5280 \(\pm\) 0.0199}
& \textbf{1} & \textbf{19}
& \textbf{CIDER-FM}
& \textbf{0.5280 \(\pm\) 0.0199}
\\

\midrule
\multirow{5}{=}{\texttt{fusion\_non\_id}}
& \multirow{3}{=}{\(Y_q\)}
& NLL
& \textbf{1.4653 \(\pm\) 0.0341}
& \textbf{1} & \textbf{11}
& \textbf{CIDER-FM}
& \textbf{1.4653 \(\pm\) 0.0341}
\\
& & MSE
& 3.1356 \(\pm\) 0.4350
& 4 & 19
& TabPFN v2 (\texttt{fusion})
& 2.6392 \(\pm\) 0.2905
\\
& & \(R^2\)
& \textbf{0.3699 \(\pm\) 0.0135}
& \textbf{1} & \textbf{19}
& \textbf{CIDER-FM}
& \textbf{0.3699 \(\pm\) 0.0135}
\\
\cmidrule(lr){2-8}
& \multirow{2}{=}{\(Y_{\mathrm{mean}}\)}
& MSE
& 2.1801 \(\pm\) 0.4270
& 4 & 19
& TabPFN v2 (\texttt{fusion})
& 1.6919 \(\pm\) 0.2768
\\
& & \(R^2\)
& \textbf{0.5787 \(\pm\) 0.0166}
& \textbf{1} & \textbf{19}
& \textbf{CIDER-FM}
& \textbf{0.5787 \(\pm\) 0.0166}
\\

\bottomrule
\end{tabular}
\endgroup
\end{table}

\subsection{Effect of interventional data ratio}

To isolate the effect of context allocation, we fix a four-node
\texttt{fusion\_id} graph and vary the fraction of interventional rows while
holding the total context budget at \(192\). All allocations are evaluated on
the same held-out SCMs and query rows.

\begin{figure*}[t]
  \centering
  \includegraphics[width=0.95\textwidth]
  {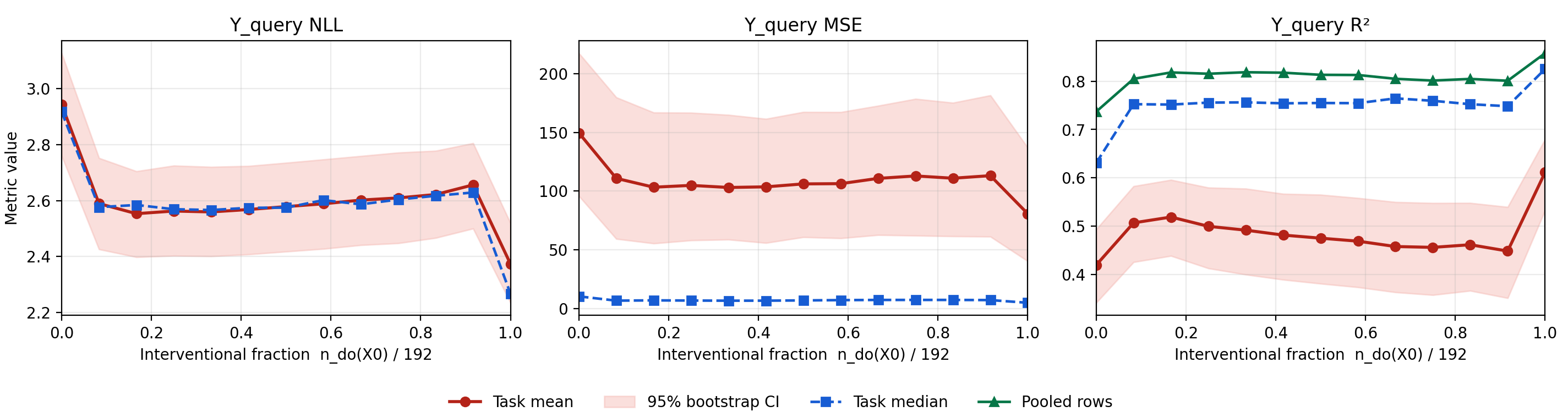}
  \caption{\textbf{Case studies:} Effect of observational--interventional context allocation on
  linear--Gaussian CIDER-FM for a fixed four-node \texttt{fusion\_id}
  graph. The total context budget is fixed at 192 rows, while the horizontal
  axis varies the fraction of randomised interventional samples,
  \(n_{\operatorname{do}(X_0)}/192\), from entirely observational data
  (\(0\)) to entirely interventional data (\(1\)).}
  \label{fig:linear_gaussian_context_ratio}
\end{figure*}

As shown in Figure~\ref{fig:linear_gaussian_context_ratio}, introducing even a
small fraction of interventional rows substantially improves performance over
the observation-only allocation, followed by relatively stable performance
across mixed allocations.

\begin{table}[p]
\centering
\caption{Restricted-family linear--Gaussian validation: all comparison methods.
Mean \(\pm\) SE across 520 SCMs per setting (130 each at 4, 6, 8 and 10 nodes).
Bold denotes the best applicable mean in each panel, excluding gray \(c\text{-}g\text{ID}\) rows. \textit{N/A}: no density or identification route.}
\label{tab:restricted-lg-all-methods}
\begingroup
\scriptsize
\setlength{\tabcolsep}{1.5pt}
\renewcommand{\arraystretch}{0.98}
\begin{tabular}{@{}>{\raggedright\arraybackslash}p{0.255\textwidth}>{\centering\arraybackslash}p{0.161\textwidth}*{4}{>{\centering\arraybackslash}p{0.131\textwidth}}@{}}
\toprule
Method & \multicolumn{3}{c}{\(Y_q\)} & \multicolumn{2}{c}{\(Y_{\mathrm{mean}}\)} \\
\cmidrule(lr){2-4}\cmidrule(lr){5-6}
& NLL \(\downarrow\) & MSE \(\downarrow\) & \(R^2\) \(\uparrow\) & MSE \(\downarrow\) & \(R^2\) \(\uparrow\) \\
\midrule
\multicolumn{6}{@{}l}{\textbf{\texttt{obs\_id}}} \\
CIDER-FM & 0.9456 \(\pm\) 0.0440 & 0.9097 \(\pm\) 0.0922 & \textbf{0.7330 \(\pm\) 0.0174} & 0.4131 \(\pm\) 0.0769 & \textbf{0.8498 \(\pm\) 0.0241} \\
CIDER-FM-Obs (\texttt{obs\_total}) & \textbf{0.9338 \(\pm\) 0.0394} & \textbf{0.8679 \(\pm\) 0.0766} & 0.7299 \(\pm\) 0.0190 & \textbf{0.3711 \(\pm\) 0.0592} & 0.8345 \(\pm\) 0.0305 \\
LR (\texttt{obs\_total}) & \textit{N/A} & 1.1995 \(\pm\) 0.2220 & 0.6022 \(\pm\) 0.0768 & 0.7032 \(\pm\) 0.2122 & 0.5799 \(\pm\) 0.1247 \\
LR (\texttt{fusion}) & \textit{N/A} & 1.2662 \(\pm\) 0.2240 & 0.5798 \(\pm\) 0.0787 & 0.7703 \(\pm\) 0.2143 & 0.5556 \(\pm\) 0.1255 \\
RF (\texttt{obs\_total}) & \textit{N/A} & 1.5607 \(\pm\) 0.1293 & 0.6296 \(\pm\) 0.0123 & 1.0665 \(\pm\) 0.1165 & 0.7299 \(\pm\) 0.0151 \\
RF (\texttt{fusion}) & \textit{N/A} & 1.5814 \(\pm\) 0.1392 & 0.6362 \(\pm\) 0.0117 & 1.0902 \(\pm\) 0.1273 & 0.7381 \(\pm\) 0.0139 \\
TabPFN v2 (\texttt{obs\_total}) & \(+\infty\) & 0.9016 \(\pm\) 0.1170 & 0.7328 \(\pm\) 0.0229 & 0.4054 \(\pm\) 0.0995 & 0.7977 \(\pm\) 0.0496 \\
TabPFN v2 (\texttt{fusion}) & \(+\infty\) & 0.9386 \(\pm\) 0.1222 & 0.7211 \(\pm\) 0.0240 & 0.4422 \(\pm\) 0.1056 & 0.7816 \(\pm\) 0.0511 \\
BLR (\texttt{obs\_total}) & 1.4017 \(\pm\) 0.3783 & 1.1984 \(\pm\) 0.2216 & 0.6027 \(\pm\) 0.0765 & 0.7021 \(\pm\) 0.2118 & 0.5805 \(\pm\) 0.1244 \\
BLR (\texttt{fusion}) & 1.4831 \(\pm\) 0.3913 & 1.2653 \(\pm\) 0.2237 & 0.5802 \(\pm\) 0.0785 & 0.7694 \(\pm\) 0.2140 & 0.5561 \(\pm\) 0.1252 \\
DML (\texttt{obs\_total}) & \textit{N/A} & 1.6023 \(\pm\) 0.2267 & 0.6214 \(\pm\) 0.0207 & 1.1095 \(\pm\) 0.2185 & 0.6621 \(\pm\) 0.0378 \\
Do-PFN (\texttt{obs\_total}) & \textit{N/A} & 1.7719 \(\pm\) 0.1718 & 0.6358 \(\pm\) 0.0152 & 1.2787 \(\pm\) 0.1635 & 0.7282 \(\pm\) 0.0239 \\
TabPFN v2 (\texttt{obs\_fixed}) & \(+\infty\) & 0.9652 \(\pm\) 0.1202 & 0.7167 \(\pm\) 0.0223 & 0.4700 \(\pm\) 0.1027 & 0.7786 \(\pm\) 0.0464 \\
DML (\texttt{obs\_fixed}) & \textit{N/A} & 1.9409 \(\pm\) 0.2622 & 0.5783 \(\pm\) 0.0182 & 1.4473 \(\pm\) 0.2555 & 0.6135 \(\pm\) 0.0337 \\
Do-PFN (\texttt{obs\_fixed}) & \textit{N/A} & 1.9165 \(\pm\) 0.1909 & 0.6089 \(\pm\) 0.0169 & 1.4234 \(\pm\) 0.1832 & 0.6834 \(\pm\) 0.0280 \\
ArCO-GP & 1.5814 \(\pm\) 0.0574 & 3.0684 \(\pm\) 0.9451 & 0.5162 \(\pm\) 0.0193 & 2.5714 \(\pm\) 0.9443 & 0.5553 \(\pm\) 0.0285 \\
Graph4CFM (no graph) & \((1.23\pm0.44)\!\times\!10^{7}\) & 3.3129 \(\pm\) 0.4744 & 0.4499 \(\pm\) 0.0181 & 2.8223 \(\pm\) 0.4707 & 0.5159 \(\pm\) 0.0219 \\
Graph4CFM (ancestors of \(Y\)) & \((9.26\pm4.03)\!\times\!10^{6}\) & 1.6375 \(\pm\) 0.1809 & 0.6337 \(\pm\) 0.0140 & 1.1479 \(\pm\) 0.1727 & 0.7051 \(\pm\) 0.0202 \\
Graph4CFM (full ancestry) & \((1.66\pm0.67)\!\times\!10^{7}\) & 1.3903 \(\pm\) 0.1015 & 0.6447 \(\pm\) 0.0141 & 0.8997 \(\pm\) 0.0862 & 0.7137 \(\pm\) 0.0214 \\
\textcolor{gray!70}{\(c\text{-}g\text{ID}\) + LR (\texttt{obs\_total})} & \textcolor{gray!70}{\textit{N/A}} & \textcolor{gray!70}{1.2001 \(\pm\) 0.2225} & \textcolor{gray!70}{0.6026 \(\pm\) 0.0768} & \textcolor{gray!70}{0.7037 \(\pm\) 0.2128} & \textcolor{gray!70}{0.5810 \(\pm\) 0.1248} \\
\textcolor{gray!70}{\(c\text{-}g\text{ID}\) + LR (\texttt{fusion})} & \textcolor{gray!70}{\textit{N/A}} & \textcolor{gray!70}{1.4543 \(\pm\) 0.2541} & \textcolor{gray!70}{0.5216 \(\pm\) 0.0844} & \textcolor{gray!70}{0.9594 \(\pm\) 0.2444} & \textcolor{gray!70}{0.4640 \(\pm\) 0.1329} \\
\textcolor{gray!70}{\(c\text{-}g\text{ID}\) + BLR (\texttt{obs\_total})} & \textcolor{gray!70}{0.6731 \(\pm\) 0.0420} & \textcolor{gray!70}{0.7925 \(\pm\) 0.1470} & \textcolor{gray!70}{0.7012 \(\pm\) 0.0721} & \textcolor{gray!70}{0.2997 \(\pm\) 0.1436} & \textcolor{gray!70}{0.8091 \(\pm\) 0.1050} \\
\textcolor{gray!70}{\(c\text{-}g\text{ID}\) + BLR (\texttt{fusion})} & \textcolor{gray!70}{0.7522 \(\pm\) 0.0434} & \textcolor{gray!70}{1.0066 \(\pm\) 0.1682} & \textcolor{gray!70}{0.6267 \(\pm\) 0.0789} & \textcolor{gray!70}{0.5142 \(\pm\) 0.1642} & \textcolor{gray!70}{0.6986 \(\pm\) 0.1140} \\
\midrule
\multicolumn{6}{@{}l}{\textbf{\texttt{fusion\_id}}} \\
CIDER-FM & \textbf{1.6150 \(\pm\) 0.0403} & \textbf{1.8332 \(\pm\) 0.1013} & \textbf{0.6006 \(\pm\) 0.0188} & \textbf{0.6746 \(\pm\) 0.0585} & 0.7089 \(\pm\) 0.0398 \\
CIDER-FM-Obs (\texttt{obs\_total}) & 1.6873 \(\pm\) 0.0424 & 2.1612 \(\pm\) 0.1396 & 0.5477 \(\pm\) 0.0206 & 0.9983 \(\pm\) 0.1029 & 0.6093 \(\pm\) 0.0399 \\
LR (\texttt{obs\_total}) & \textit{N/A} & 4.2547 \(\pm\) 0.9145 & 0.1409 \(\pm\) 0.1760 & 3.0911 \(\pm\) 0.9017 & -0.3499 \(\pm\) 0.3784 \\
LR (\texttt{fusion}) & \textit{N/A} & 2.0682 \(\pm\) 0.1213 & 0.5616 \(\pm\) 0.0204 & 0.9135 \(\pm\) 0.0791 & 0.6332 \(\pm\) 0.0331 \\
RF (\texttt{obs\_total}) & \textit{N/A} & 2.2537 \(\pm\) 0.1189 & 0.5497 \(\pm\) 0.0172 & 1.0964 \(\pm\) 0.0804 & 0.6278 \(\pm\) 0.0279 \\
RF (\texttt{fusion}) & \textit{N/A} & 1.9928 \(\pm\) 0.0941 & 0.5924 \(\pm\) 0.0145 & 0.8392 \(\pm\) 0.0508 & \textbf{0.7212 \(\pm\) 0.0186} \\
TabPFN v2 (\texttt{obs\_total}) & \(+\infty\) & 2.8170 \(\pm\) 0.2231 & 0.4275 \(\pm\) 0.0327 & 1.6560 \(\pm\) 0.1896 & 0.3209 \(\pm\) 0.0779 \\
TabPFN v2 (\texttt{fusion}) & \(+\infty\) & 1.9752 \(\pm\) 0.1103 & 0.5762 \(\pm\) 0.0191 & 0.8193 \(\pm\) 0.0679 & 0.6664 \(\pm\) 0.0297 \\
BLR (\texttt{obs\_total}) & 4.4714 \(\pm\) 0.5673 & 4.2543 \(\pm\) 0.9143 & 0.1410 \(\pm\) 0.1760 & 3.0908 \(\pm\) 0.9015 & -0.3498 \(\pm\) 0.3783 \\
BLR (\texttt{fusion}) & 2.2876 \(\pm\) 0.1280 & 2.0682 \(\pm\) 0.1213 & 0.5616 \(\pm\) 0.0204 & 0.9135 \(\pm\) 0.0791 & 0.6332 \(\pm\) 0.0331 \\
DML (\texttt{obs\_total}) & \textit{N/A} & 2.8567 \(\pm\) 0.1671 & 0.4214 \(\pm\) 0.0255 & 1.6960 \(\pm\) 0.1274 & 0.3471 \(\pm\) 0.0497 \\
Do-PFN (\texttt{obs\_total}) & \textit{N/A} & 2.1919 \(\pm\) 0.1199 & 0.4261 \(\pm\) 0.1261 & 1.0304 \(\pm\) 0.0812 & 0.0798 \(\pm\) 0.5483 \\
TabPFN v2 (\texttt{obs\_fixed}) & \(+\infty\) & 2.6481 \(\pm\) 0.2073 & 0.4610 \(\pm\) 0.0284 & 1.4834 \(\pm\) 0.1759 & 0.4162 \(\pm\) 0.0582 \\
DML (\texttt{obs\_fixed}) & \textit{N/A} & 3.2070 \(\pm\) 0.1722 & 0.4034 \(\pm\) 0.0211 & 2.0389 \(\pm\) 0.1358 & 0.3664 \(\pm\) 0.0377 \\
Do-PFN (\texttt{obs\_fixed}) & \textit{N/A} & 2.3191 \(\pm\) 0.1256 & 0.2242 \(\pm\) 0.2896 & 1.1569 \(\pm\) 0.0889 & -0.7139 \(\pm\) 1.2678 \\
ArCO-GP & 1.7217 \(\pm\) 0.0425 & 2.1228 \(\pm\) 0.1168 & 0.5351 \(\pm\) 0.0203 & 0.9687 \(\pm\) 0.0730 & 0.5748 \(\pm\) 0.0392 \\
Graph4CFM (no graph) & \((5.60\pm3.89)\!\times\!10^{6}\) & 2.0031 \(\pm\) 0.1041 & 0.5899 \(\pm\) 0.0148 & 0.8510 \(\pm\) 0.0624 & 0.7108 \(\pm\) 0.0204 \\
Graph4CFM (ancestors of \(Y\)) & \((3.62\pm2.97)\!\times\!10^{6}\) & 2.4361 \(\pm\) 0.1311 & 0.4987 \(\pm\) 0.0197 & 1.2800 \(\pm\) 0.0893 & 0.5025 \(\pm\) 0.0377 \\
Graph4CFM (full ancestry) & \((3.68\pm3.06)\!\times\!10^{6}\) & 2.4239 \(\pm\) 0.1272 & 0.4933 \(\pm\) 0.0199 & 1.2717 \(\pm\) 0.0851 & 0.4945 \(\pm\) 0.0387 \\
\textcolor{gray!70}{\(c\text{-}g\text{ID}\) + LR (\texttt{fusion})} & \textcolor{gray!70}{\textit{N/A}} & \textcolor{gray!70}{1.2398 \(\pm\) 0.0610} & \textcolor{gray!70}{0.7165 \(\pm\) 0.0109} & \textcolor{gray!70}{0.0807 \(\pm\) 0.0054} & \textcolor{gray!70}{0.9611 \(\pm\) 0.0031} \\
\textcolor{gray!70}{\(c\text{-}g\text{ID}\) + BLR (\texttt{fusion})} & \textcolor{gray!70}{1.2273 \(\pm\) 0.0279} & \textcolor{gray!70}{1.2409 \(\pm\) 0.0611} & \textcolor{gray!70}{0.7163 \(\pm\) 0.0109} & \textcolor{gray!70}{0.0819 \(\pm\) 0.0055} & \textcolor{gray!70}{0.9607 \(\pm\) 0.0032} \\
\midrule
\multicolumn{6}{@{}l}{\textbf{\texttt{fusion\_non\_id}}} \\
CIDER-FM & \textbf{1.7229 \(\pm\) 0.0391} & 2.8267 \(\pm\) 0.3385 & \textbf{0.5428 \(\pm\) 0.0345} & 1.5011 \(\pm\) 0.3232 & \textbf{0.6243 \(\pm\) 0.0551} \\
CIDER-FM-Obs (\texttt{obs\_total}) & 1.7671 \(\pm\) 0.0395 & \textbf{2.7895 \(\pm\) 0.2367} & 0.5420 \(\pm\) 0.0266 & \textbf{1.4586 \(\pm\) 0.2111} & 0.6103 \(\pm\) 0.0477 \\
LR (\texttt{obs\_total}) & \textit{N/A} & 6.2806 \(\pm\) 1.6874 & 0.1120 \(\pm\) 0.1614 & 4.9374 \(\pm\) 1.6745 & -0.0567 \(\pm\) 0.2038 \\
LR (\texttt{fusion}) & \textit{N/A} & 5.8364 \(\pm\) 1.6790 & 0.1872 \(\pm\) 0.1544 & 4.4960 \(\pm\) 1.6662 & 0.0878 \(\pm\) 0.1904 \\
RF (\texttt{obs\_total}) & \textit{N/A} & 3.0442 \(\pm\) 0.1482 & 0.5068 \(\pm\) 0.0164 & 1.7309 \(\pm\) 0.1156 & 0.5857 \(\pm\) 0.0251 \\
RF (\texttt{fusion}) & \textit{N/A} & 2.9690 \(\pm\) 0.1366 & 0.5104 \(\pm\) 0.0159 & 1.6524 \(\pm\) 0.0995 & 0.5952 \(\pm\) 0.0226 \\
TabPFN v2 (\texttt{obs\_total}) & \(+\infty\) & 3.4143 \(\pm\) 0.2938 & 0.4447 \(\pm\) 0.0319 & 2.0785 \(\pm\) 0.2677 & 0.4183 \(\pm\) 0.0580 \\
TabPFN v2 (\texttt{fusion}) & \(+\infty\) & 3.1366 \(\pm\) 0.2622 & 0.4817 \(\pm\) 0.0274 & 1.8006 \(\pm\) 0.2343 & 0.4915 \(\pm\) 0.0490 \\
BLR (\texttt{obs\_total}) & 4.2846 \(\pm\) 0.4914 & 6.2784 \(\pm\) 1.6858 & 0.1122 \(\pm\) 0.1613 & 4.9353 \(\pm\) 1.6729 & -0.0564 \(\pm\) 0.2036 \\
BLR (\texttt{fusion}) & 3.2158 \(\pm\) 0.3140 & 5.8342 \(\pm\) 1.6773 & 0.1874 \(\pm\) 0.1543 & 4.4938 \(\pm\) 1.6646 & 0.0880 \(\pm\) 0.1902 \\
DML (\texttt{obs\_total}) & \textit{N/A} & 3.4549 \(\pm\) 0.2172 & 0.4324 \(\pm\) 0.0255 & 2.1251 \(\pm\) 0.1843 & 0.3882 \(\pm\) 0.0495 \\
Do-PFN (\texttt{obs\_total}) & \textit{N/A} & 3.3871 \(\pm\) 0.2355 & 0.5069 \(\pm\) 0.0200 & 2.0648 \(\pm\) 0.2167 & 0.5800 \(\pm\) 0.0325 \\
TabPFN v2 (\texttt{obs\_fixed}) & \(+\infty\) & 3.2801 \(\pm\) 0.2779 & 0.4662 \(\pm\) 0.0298 & 1.9469 \(\pm\) 0.2512 & 0.4637 \(\pm\) 0.0524 \\
DML (\texttt{obs\_fixed}) & \textit{N/A} & 3.5700 \(\pm\) 0.2166 & 0.4243 \(\pm\) 0.0232 & 2.2489 \(\pm\) 0.1839 & 0.3932 \(\pm\) 0.0428 \\
Do-PFN (\texttt{obs\_fixed}) & \textit{N/A} & 3.3797 \(\pm\) 0.2198 & 0.4956 \(\pm\) 0.0207 & 2.0565 \(\pm\) 0.1995 & 0.5630 \(\pm\) 0.0372 \\
ArCO-GP & 2.4001 \(\pm\) 0.0816 & 4.4181 \(\pm\) 0.2995 & 0.3444 \(\pm\) 0.0249 & 3.0921 \(\pm\) 0.2790 & 0.3353 \(\pm\) 0.0398 \\
Graph4CFM (no graph) & \((1.76\pm0.48)\!\times\!10^{7}\) & 5.1302 \(\pm\) 0.4418 & 0.3810 \(\pm\) 0.0169 & 3.8112 \(\pm\) 0.4340 & 0.4570 \(\pm\) 0.0209 \\
Graph4CFM (ancestors of \(Y\)) & \((8.18\pm3.01)\!\times\!10^{6}\) & 3.3115 \(\pm\) 0.1773 & 0.4789 \(\pm\) 0.0186 & 1.9872 \(\pm\) 0.1416 & 0.4982 \(\pm\) 0.0304 \\
Graph4CFM (full ancestry) & \((1.17\pm0.57)\!\times\!10^{7}\) & 3.1786 \(\pm\) 0.1636 & 0.4817 \(\pm\) 0.0191 & 1.8515 \(\pm\) 0.1224 & 0.4969 \(\pm\) 0.0322 \\
\bottomrule
\end{tabular}
\endgroup
\end{table}

\begin{table}[p]
\centering
\caption{Restricted-family nonlinear--Gaussian validation: all comparison methods.
Mean \(\pm\) SE across 520 SCMs per setting (130 each at 4, 6, 8 and 10 nodes).
Bold denotes the best applicable mean in each panel, excluding gray \(c\text{-}g\text{ID}\) rows. \textit{N/A}: no density or identification route.}
\label{tab:restricted-nonlinear-all-methods}
\begingroup
\scriptsize
\setlength{\tabcolsep}{1.5pt}
\renewcommand{\arraystretch}{0.98}
\begin{tabular}{@{}>{\raggedright\arraybackslash}p{0.255\textwidth}>{\centering\arraybackslash}p{0.161\textwidth}*{4}{>{\centering\arraybackslash}p{0.131\textwidth}}@{}}
\toprule
Method & \multicolumn{3}{c}{\(Y_q\)} & \multicolumn{2}{c}{\(Y_{\mathrm{mean}}\)} \\
\cmidrule(lr){2-4}\cmidrule(lr){5-6}
& NLL \(\downarrow\) & MSE \(\downarrow\) & \(R^2\) \(\uparrow\) & MSE \(\downarrow\) & \(R^2\) \(\uparrow\) \\
\midrule
\multicolumn{6}{@{}l}{\textbf{\texttt{obs\_id}}} \\
CIDER-FM & \textbf{0.9537 \(\pm\) 0.0323} & 1.5096 \(\pm\) 0.2347 & 0.3737 \(\pm\) 0.0154 & 1.1354 \(\pm\) 0.2335 & 0.5057 \(\pm\) 0.0176 \\
CIDER-FM-Obs (\texttt{obs\_total}) & 1.0179 \(\pm\) 0.0337 & 1.7253 \(\pm\) 0.3161 & 0.3430 \(\pm\) 0.0167 & 1.3486 \(\pm\) 0.3153 & 0.4258 \(\pm\) 0.0229 \\
LR (\texttt{obs\_total}) & \textit{N/A} & 4.4135 \(\pm\) 0.9504 & 0.2465 \(\pm\) 0.0185 & 4.0367 \(\pm\) 0.9513 & 0.3103 \(\pm\) 0.0285 \\
LR (\texttt{fusion}) & \textit{N/A} & 4.4630 \(\pm\) 0.9372 & 0.2212 \(\pm\) 0.0171 & 4.0890 \(\pm\) 0.9378 & 0.2764 \(\pm\) 0.0240 \\
RF (\texttt{obs\_total}) & \textit{N/A} & 4.3342 \(\pm\) 0.8977 & 0.3009 \(\pm\) 0.0147 & 3.9576 \(\pm\) 0.8965 & 0.3807 \(\pm\) 0.0194 \\
RF (\texttt{fusion}) & \textit{N/A} & 4.1705 \(\pm\) 0.8432 & 0.2727 \(\pm\) 0.0138 & 3.7958 \(\pm\) 0.8417 & 0.3359 \(\pm\) 0.0180 \\
TabPFN v2 (\texttt{obs\_total}) & \(+\infty\) & \textbf{1.3793 \(\pm\) 0.2732} & \textbf{0.3936 \(\pm\) 0.0156} & \textbf{1.0021 \(\pm\) 0.2722} & \textbf{0.5299 \(\pm\) 0.0182} \\
TabPFN v2 (\texttt{fusion}) & \(+\infty\) & 1.4054 \(\pm\) 0.2700 & 0.3591 \(\pm\) 0.0151 & 1.0302 \(\pm\) 0.2687 & 0.4754 \(\pm\) 0.0176 \\
BLR (\texttt{obs\_total}) & 1.0837 \(\pm\) 0.0390 & 4.4135 \(\pm\) 0.9504 & 0.2465 \(\pm\) 0.0185 & 4.0367 \(\pm\) 0.9513 & 0.3103 \(\pm\) 0.0285 \\
BLR (\texttt{fusion}) & 1.1323 \(\pm\) 0.0379 & 4.4630 \(\pm\) 0.9372 & 0.2212 \(\pm\) 0.0171 & 4.0890 \(\pm\) 0.9378 & 0.2764 \(\pm\) 0.0240 \\
DML (\texttt{obs\_total}) & \textit{N/A} & 4.3299 \(\pm\) 0.8829 & 0.1817 \(\pm\) 0.0190 & 3.9548 \(\pm\) 0.8832 & 0.1295 \(\pm\) 0.0299 \\
Do-PFN (\texttt{obs\_total}) & \textit{N/A} & 3.0252 \(\pm\) 0.5519 & 0.1326 \(\pm\) 0.0717 & 2.6511 \(\pm\) 0.5511 & 0.1007 \(\pm\) 0.1159 \\
TabPFN v2 (\texttt{obs\_fixed}) & \(+\infty\) & 1.7490 \(\pm\) 0.3133 & 0.3518 \(\pm\) 0.0159 & 1.3727 \(\pm\) 0.3121 & 0.4565 \(\pm\) 0.0196 \\
DML (\texttt{obs\_fixed}) & \textit{N/A} & 5.2764 \(\pm\) 1.0323 & 0.1212 \(\pm\) 0.0198 & 4.9028 \(\pm\) 1.0320 & 0.0448 \(\pm\) 0.0295 \\
Do-PFN (\texttt{obs\_fixed}) & \textit{N/A} & 3.4477 \(\pm\) 0.6594 & 0.0918 \(\pm\) 0.0531 & 3.0729 \(\pm\) 0.6582 & -0.0570 \(\pm\) 0.1174 \\
ArCO-GP & 1.1430 \(\pm\) 0.0378 & 3.7139 \(\pm\) 0.8339 & 0.2526 \(\pm\) 0.0163 & 3.3400 \(\pm\) 0.8300 & 0.2804 \(\pm\) 0.0197 \\
Graph4CFM (no graph) & \((5.78\pm1.08)\!\times\!10^{6}\) & 4.3147 \(\pm\) 1.1176 & 0.2483 \(\pm\) 0.0162 & 3.9361 \(\pm\) 1.1178 & 0.2755 \(\pm\) 0.0225 \\
Graph4CFM (ancestors of \(Y\)) & \((3.04\pm0.58)\!\times\!10^{6}\) & 2.7311 \(\pm\) 0.5113 & 0.2433 \(\pm\) 0.0176 & 2.3546 \(\pm\) 0.5101 & 0.2179 \(\pm\) 0.0260 \\
Graph4CFM (full ancestry) & \((3.65\pm0.68)\!\times\!10^{6}\) & 2.9989 \(\pm\) 0.5628 & 0.2469 \(\pm\) 0.0173 & 2.6199 \(\pm\) 0.5608 & 0.2316 \(\pm\) 0.0255 \\
\textcolor{gray!70}{\(c\text{-}g\text{ID}\) + LR (\texttt{obs\_total})} & \textcolor{gray!70}{\textit{N/A}} & \textcolor{gray!70}{4.3975 \(\pm\) 0.9482} & \textcolor{gray!70}{0.2489 \(\pm\) 0.0184} & \textcolor{gray!70}{4.0206 \(\pm\) 0.9491} & \textcolor{gray!70}{0.3161 \(\pm\) 0.0283} \\
\textcolor{gray!70}{\(c\text{-}g\text{ID}\) + LR (\texttt{fusion})} & \textcolor{gray!70}{\textit{N/A}} & \textcolor{gray!70}{4.5985 \(\pm\) 0.9735} & \textcolor{gray!70}{0.1966 \(\pm\) 0.0236} & \textcolor{gray!70}{4.2216 \(\pm\) 0.9742} & \textcolor{gray!70}{0.2088 \(\pm\) 0.0445} \\
\textcolor{gray!70}{\(c\text{-}g\text{ID}\) + BLR (\texttt{obs\_total})} & \textcolor{gray!70}{1.0793 \(\pm\) 0.0389} & \textcolor{gray!70}{4.3914 \(\pm\) 0.9500} & \textcolor{gray!70}{0.2531 \(\pm\) 0.0184} & \textcolor{gray!70}{4.0147 \(\pm\) 0.9510} & \textcolor{gray!70}{0.3259 \(\pm\) 0.0287} \\
\textcolor{gray!70}{\(c\text{-}g\text{ID}\) + BLR (\texttt{fusion})} & \textcolor{gray!70}{1.1201 \(\pm\) 0.0396} & \textcolor{gray!70}{4.5879 \(\pm\) 0.9745} & \textcolor{gray!70}{0.1992 \(\pm\) 0.0245} & \textcolor{gray!70}{4.2114 \(\pm\) 0.9752} & \textcolor{gray!70}{0.2154 \(\pm\) 0.0473} \\
\midrule
\multicolumn{6}{@{}l}{\textbf{\texttt{fusion\_id}}} \\
CIDER-FM & \textbf{1.3479 \(\pm\) 0.0306} & 3.5265 \(\pm\) 0.7781 & \textbf{0.4109 \(\pm\) 0.0169} & 2.7177 \(\pm\) 0.7732 & \textbf{0.5280 \(\pm\) 0.0199} \\
CIDER-FM-Obs (\texttt{obs\_total}) & 1.3596 \(\pm\) 0.0306 & 3.5618 \(\pm\) 0.8391 & 0.4018 \(\pm\) 0.0170 & 2.7515 \(\pm\) 0.8332 & 0.5083 \(\pm\) 0.0192 \\
LR (\texttt{obs\_total}) & \textit{N/A} & 14.6202 \(\pm\) 2.2011 & -0.6410 \(\pm\) 0.4039 & 13.7846 \(\pm\) 2.1971 & -1.8777 \(\pm\) 0.8024 \\
LR (\texttt{fusion}) & \textit{N/A} & 13.7198 \(\pm\) 1.9163 & -0.3225 \(\pm\) 0.2689 & 12.8941 \(\pm\) 1.9118 & -1.1239 \(\pm\) 0.5856 \\
RF (\texttt{obs\_total}) & \textit{N/A} & 7.9823 \(\pm\) 1.0253 & 0.2100 \(\pm\) 0.0209 & 7.1586 \(\pm\) 1.0167 & 0.0497 \(\pm\) 0.0453 \\
RF (\texttt{fusion}) & \textit{N/A} & 7.7285 \(\pm\) 1.0228 & 0.2431 \(\pm\) 0.0198 & 6.9169 \(\pm\) 1.0148 & 0.1285 \(\pm\) 0.0399 \\
TabPFN v2 (\texttt{obs\_total}) & \(+\infty\) & 3.3423 \(\pm\) 0.4945 & 0.1852 \(\pm\) 0.0308 & 2.5294 \(\pm\) 0.4851 & -0.1412 \(\pm\) 0.0836 \\
TabPFN v2 (\texttt{fusion}) & \(+\infty\) & \textbf{3.0792 \(\pm\) 0.4604} & 0.2789 \(\pm\) 0.0270 & \textbf{2.2722 \(\pm\) 0.4526} & 0.1165 \(\pm\) 0.0710 \\
BLR (\texttt{obs\_total}) & 2.1456 \(\pm\) 0.1137 & 14.6190 \(\pm\) 2.2004 & -0.6406 \(\pm\) 0.4036 & 13.7834 \(\pm\) 2.1964 & -1.8770 \(\pm\) 0.8020 \\
BLR (\texttt{fusion}) & 1.9460 \(\pm\) 0.0893 & 13.7190 \(\pm\) 1.9160 & -0.3222 \(\pm\) 0.2687 & 12.8933 \(\pm\) 1.9115 & -1.1234 \(\pm\) 0.5853 \\
DML (\texttt{obs\_total}) & \textit{N/A} & 7.6543 \(\pm\) 0.9398 & 0.0093 \(\pm\) 0.0391 & 6.8368 \(\pm\) 0.9310 & -0.5047 \(\pm\) 0.1010 \\
Do-PFN (\texttt{obs\_total}) & \textit{N/A} & 8.0483 \(\pm\) 1.7881 & 0.1679 \(\pm\) 0.0440 & 7.2132 \(\pm\) 1.7739 & 0.0605 \(\pm\) 0.0699 \\
TabPFN v2 (\texttt{obs\_fixed}) & \(+\infty\) & 4.8245 \(\pm\) 0.6867 & 0.1790 \(\pm\) 0.0328 & 4.0117 \(\pm\) 0.6780 & -0.0517 \(\pm\) 0.0640 \\
DML (\texttt{obs\_fixed}) & \textit{N/A} & 12.8796 \(\pm\) 1.6096 & -0.0641 \(\pm\) 0.0434 & 12.0626 \(\pm\) 1.6049 & -0.6073 \(\pm\) 0.1409 \\
Do-PFN (\texttt{obs\_fixed}) & \textit{N/A} & 8.7356 \(\pm\) 1.3301 & -0.3076 \(\pm\) 0.3601 & 7.9071 \(\pm\) 1.3155 & -0.6048 \(\pm\) 0.4368 \\
ArCO-GP & 1.8391 \(\pm\) 0.0749 & 6.5506 \(\pm\) 0.9071 & 0.2190 \(\pm\) 0.0227 & 5.7445 \(\pm\) 0.8990 & 0.0820 \(\pm\) 0.0477 \\
Graph4CFM (no graph) & \((1.39\pm0.46)\!\times\!10^{6}\) & 4.9189 \(\pm\) 0.7050 & 0.3303 \(\pm\) 0.0174 & 4.1078 \(\pm\) 0.6983 & 0.3522 \(\pm\) 0.0221 \\
Graph4CFM (ancestors of \(Y\)) & \((1.12\pm0.58)\!\times\!10^{6}\) & 6.1153 \(\pm\) 0.8604 & 0.1558 \(\pm\) 0.0234 & 5.3115 \(\pm\) 0.8547 & -0.1210 \(\pm\) 0.0485 \\
Graph4CFM (full ancestry) & \((6.78\pm2.39)\!\times\!10^{5}\) & 6.1298 \(\pm\) 0.7988 & 0.1433 \(\pm\) 0.0255 & 5.3170 \(\pm\) 0.7914 & -0.1372 \(\pm\) 0.0505 \\
\textcolor{gray!70}{\(c\text{-}g\text{ID}\) + LR (\texttt{fusion})} & \textcolor{gray!70}{\textit{N/A}} & \textcolor{gray!70}{12.8379 \(\pm\) 1.7700} & \textcolor{gray!70}{0.2174 \(\pm\) 0.0149} & \textcolor{gray!70}{12.0144 \(\pm\) 1.7642} & \textcolor{gray!70}{0.2480 \(\pm\) 0.0193} \\
\textcolor{gray!70}{\(c\text{-}g\text{ID}\) + BLR (\texttt{fusion})} & \textcolor{gray!70}{1.6651 \(\pm\) 0.0457} & \textcolor{gray!70}{12.8258 \(\pm\) 1.7674} & \textcolor{gray!70}{0.2173 \(\pm\) 0.0149} & \textcolor{gray!70}{12.0021 \(\pm\) 1.7616} & \textcolor{gray!70}{0.2476 \(\pm\) 0.0193} \\
\midrule
\multicolumn{6}{@{}l}{\textbf{\texttt{fusion\_non\_id}}} \\
CIDER-FM & \textbf{1.4653 \(\pm\) 0.0341} & 3.1356 \(\pm\) 0.4350 & \textbf{0.3699 \(\pm\) 0.0135} & 2.1801 \(\pm\) 0.4270 & \textbf{0.5787 \(\pm\) 0.0166} \\
CIDER-FM-Obs (\texttt{obs\_total}) & 1.5593 \(\pm\) 0.0417 & 3.1330 \(\pm\) 0.4179 & 0.3355 \(\pm\) 0.0153 & 2.1793 \(\pm\) 0.4076 & 0.4809 \(\pm\) 0.0241 \\
LR (\texttt{obs\_total}) & \textit{N/A} & 5.6222 \(\pm\) 0.7918 & 0.0712 \(\pm\) 0.0488 & 4.6761 \(\pm\) 0.7868 & -0.0468 \(\pm\) 0.0721 \\
LR (\texttt{fusion}) & \textit{N/A} & 5.6966 \(\pm\) 0.7927 & 0.0565 \(\pm\) 0.0482 & 4.7565 \(\pm\) 0.7883 & -0.0489 \(\pm\) 0.0682 \\
RF (\texttt{obs\_total}) & \textit{N/A} & 5.5545 \(\pm\) 0.8677 & 0.2272 \(\pm\) 0.0166 & 4.6064 \(\pm\) 0.8600 & 0.2147 \(\pm\) 0.0339 \\
RF (\texttt{fusion}) & \textit{N/A} & 5.8814 \(\pm\) 0.9057 & 0.1829 \(\pm\) 0.0164 & 4.9353 \(\pm\) 0.8978 & 0.1273 \(\pm\) 0.0353 \\
TabPFN v2 (\texttt{obs\_total}) & \(+\infty\) & 2.6881 \(\pm\) 0.2951 & 0.2627 \(\pm\) 0.0228 & 1.7397 \(\pm\) 0.2812 & 0.2314 \(\pm\) 0.0541 \\
TabPFN v2 (\texttt{fusion}) & \(+\infty\) & \textbf{2.6392 \(\pm\) 0.2905} & 0.2473 \(\pm\) 0.0214 & \textbf{1.6919 \(\pm\) 0.2768} & 0.2204 \(\pm\) 0.0470 \\
BLR (\texttt{obs\_total}) & 1.9414 \(\pm\) 0.0727 & 5.6221 \(\pm\) 0.7918 & 0.0713 \(\pm\) 0.0488 & 4.6760 \(\pm\) 0.7868 & -0.0466 \(\pm\) 0.0720 \\
BLR (\texttt{fusion}) & 1.8149 \(\pm\) 0.0557 & 5.6965 \(\pm\) 0.7927 & 0.0566 \(\pm\) 0.0481 & 4.7564 \(\pm\) 0.7883 & -0.0487 \(\pm\) 0.0682 \\
DML (\texttt{obs\_total}) & \textit{N/A} & 6.6966 \(\pm\) 1.0043 & 0.0561 \(\pm\) 0.0223 & 5.7535 \(\pm\) 1.0000 & -0.1741 \(\pm\) 0.0538 \\
Do-PFN (\texttt{obs\_total}) & \textit{N/A} & 6.3113 \(\pm\) 1.1085 & 0.0890 \(\pm\) 0.0613 & 5.3506 \(\pm\) 1.1013 & 0.0199 \(\pm\) 0.1124 \\
TabPFN v2 (\texttt{obs\_fixed}) & \(+\infty\) & 3.3522 \(\pm\) 0.4076 & 0.2415 \(\pm\) 0.0213 & 2.4030 \(\pm\) 0.3977 & 0.2131 \(\pm\) 0.0447 \\
DML (\texttt{obs\_fixed}) & \textit{N/A} & 7.3199 \(\pm\) 1.0678 & 0.0349 \(\pm\) 0.0207 & 6.3790 \(\pm\) 1.0627 & -0.1935 \(\pm\) 0.0445 \\
Do-PFN (\texttt{obs\_fixed}) & \textit{N/A} & 6.4828 \(\pm\) 1.0465 & 0.0206 \(\pm\) 0.1036 & 5.5272 \(\pm\) 1.0423 & -0.1177 \(\pm\) 0.1821 \\
ArCO-GP & 1.8616 \(\pm\) 0.0494 & 5.7433 \(\pm\) 0.9454 & 0.1645 \(\pm\) 0.0334 & 4.7852 \(\pm\) 0.9368 & 0.0425 \(\pm\) 0.1217 \\
Graph4CFM (no graph) & \((7.87\pm2.32)\!\times\!10^{6}\) & 7.1976 \(\pm\) 1.3303 & 0.2155 \(\pm\) 0.0152 & 6.2658 \(\pm\) 1.3313 & 0.2595 \(\pm\) 0.0232 \\
Graph4CFM (ancestors of \(Y\)) & \((4.36\pm1.66)\!\times\!10^{6}\) & 4.1800 \(\pm\) 0.4962 & 0.1516 \(\pm\) 0.0192 & 3.2313 \(\pm\) 0.4857 & -0.0012 \(\pm\) 0.0389 \\
Graph4CFM (full ancestry) & \((6.50\pm2.35)\!\times\!10^{6}\) & 4.2344 \(\pm\) 0.5250 & 0.1525 \(\pm\) 0.0191 & 3.2814 \(\pm\) 0.5137 & -0.0021 \(\pm\) 0.0397 \\
\bottomrule
\end{tabular}
\endgroup
\end{table}

\section{Construction of general random graphs and complexMech}
\label{app:general}

\begin{algorithm}[H]
\small
\caption{General Random-Graph Sampling}
\label{alg:general-random-graph}
\begin{algorithmic}[1]
\Statex \makebox[5em][l]{\textbf{Require:}}
    Node-count range \([4,10]\)
\Statex \makebox[5em][l]{\textbf{Ensure:}}
    ADMG \(G=(V,E^{\to},E^{\leftrightarrow})\)

\State Initialise \(E^{\to}\gets\varnothing\) and
    \(E^{\leftrightarrow}\gets\varnothing\).
\State Sample \(N\sim\operatorname{Uniform}\{4,\ldots,10\}\) and set
    \(V=\{T,Y,X_0,\ldots,X_{N-3}\}\).
\State Sample a graph-level directed-edge probability
    \(p_{\to}\sim\operatorname{Beta}(2,3)\).
\State Uniformly sample a permutation
    \(\pi=(\pi_1,\ldots,\pi_N)\) of \(V\).
\For{\(1\leq a<b\leq N\)}
    \State Add \(\pi_a\to\pi_b\) if an independent
        \(\operatorname{Bernoulli}(p_{\to})\) draw succeeds.
\EndFor
\State Independently sample a graph-level bidirected-edge probability
    \(p_{\leftrightarrow}\sim\operatorname{Beta}(2,3)\).
\For{each unordered pair \(\{A,B\}\subset V\)}
    \State Add \(A\leftrightarrow B\) if an independent
        \(\operatorname{Bernoulli}(p_{\leftrightarrow})\) draw succeeds.
\EndFor
\State \Return \(G=(V,E^{\to},E^{\leftrightarrow})\)
\end{algorithmic}
\end{algorithm}

\paragraph{General random graphs.}
For each task, we first sample the number of observed variables as
\(N\sim\operatorname{Uniform}\{4,\ldots,10\}\), with node set
\(\{T,Y,X_0,\ldots,X_{N-3}\}\) as shown in \cref{alg:general-random-graph}. We independently draw graph-level edge
probabilities
\(p_{\to},p_{\leftrightarrow}\sim\operatorname{Beta}(2,3)\).
After sampling a uniformly random ordering
\(\pi=(\pi_1,\ldots,\pi_N)\), each directed edge
\(\pi_a\to\pi_b\), \(a<b\), is included independently with probability
\(p_{\to}\). This construction guarantees acyclicity while allowing the
directed-edge density to vary across SCMs. Each unordered node pair is
independently connected by a bidirected edge with probability
\(p_{\leftrightarrow}\), representing an independent latent common cause.
Bows and graphs without bidirected edges are both allowed. The number \(K\) of experimental regimes is sampled from a truncated geometric
prior, \(p(K=k\mid N)\propto 2^{-(k-1)}\) for
\(k=1,\ldots,N-2\); the resulting tasks contain an empirical mean of \(1.76\)
experimental regimes.

\paragraph{ComplexMech.}
ComplexMech replaces linear structural equations with randomly sampled
continuous MLP mechanisms. Each observed node and each latent root associated
with a bidirected edge receives its own MLP, whose width, depth, activation,
and noise-injection mode are sampled from the mechanism prior and then held
fixed across all regimes of the SCM. Exogenous noise is sampled independently
across rows as
\begin{equation}
\begin{aligned}
    \epsilon_{j,i} &= s_{j,i}\xi_{j,i},\\
    \xi_{j,i}
    &\sim
    0.33\,\mathcal{N}(0,1)
    +0.33\,\operatorname{Laplace}\!\left(0,\frac{1}{\sqrt{2}}\right)
    +0.34\,\frac{t_3}{\sqrt{3}},
    \qquad
    s_{j,i}\sim\operatorname{Gamma}(\alpha,\theta).
\end{aligned}
\label{eq:complexmech-noise}
\end{equation}
Separate Gamma scale distributions are sampled for root and non-root
mechanisms within each SCM. The resulting family combines nonlinear
functional relationships, heterogeneous noise scales, heavy-tailed
disturbances, and nonlinear hidden confounding.

\section{Details on causal chamber evaluation}
\label{app:chambers}

\paragraph{Data and task.}
We use the light-tunnel dataset
\texttt{lt\_crl\_benchmark\_v1}
\citep{gamella2025causal,gamella2025sanity}, following the official
collection protocol.
It contains 10,000 measurements per environment, with light-source and
polarizer commands generated by a software SCM and randomized assignments
to individual commands in the intervention environments.
The task is \(\mathrm{red}\to\mathrm{vis}_3\mid\mathrm{green},\mathrm{blue}\),
with both polarizers hidden. Query rows come from the red-intervention
environment; context sources are observations and interventions on green
or blue. The reported case is drawn from an exploratory light-tunnel study.

\paragraph{Simulator-guided design.}
We use the released deterministic counterparts of these environments,
generated by the official \texttt{lt.Deterministic} sensor simulator
\citep{gamella2025sanity}.
We enumerate the seven subsets of
\(\{\mathrm{Obs},\doop{\mathrm{green}},\doop{\mathrm{blue}}\}\). Each candidate receives
192 context rows, divided equally among its active regimes
(\cref{tab:chambers-context-designs}).
Using the frozen ComplexMech CIDER-FM, we evaluate each candidate on ten
disjoint simulator-data blocks, each with 128 query rows and three context
draws. Query rows are shared across candidates. We average NLL over repeats
and blocks and select the context with the lowest mean NLL:
96 observational rows and 96 rows from \(\doop{\mathrm{green}}\).
This design is then fixed for evaluation on real measurements.

\begin{table}[htbp]
\centering
\small
\caption{Candidate contexts for
\(\mathrm{red}\to\mathrm{vis}_3\mid\mathrm{green},\mathrm{blue}\),
each with 192 rows. Bold marks the context selected by simulator NLL.}
\label{tab:chambers-context-designs}
\begin{tabular}{@{}lrrr@{}}
\toprule
Context & Obs rows & \(\doop{\mathrm{green}}\) rows
& \(\doop{\mathrm{blue}}\) rows \\
\midrule
\(\doop{\mathrm{green}}\) & 0 & 192 & 0 \\
\(\doop{\mathrm{blue}}\) & 0 & 0 & 192 \\
\(\doop{\mathrm{green}}+\doop{\mathrm{blue}}\) & 0 & 96 & 96 \\
\textbf{Obs + \(\doop{\mathrm{green}}\)} & \textbf{96} & \textbf{96} & \textbf{0} \\
Obs + \(\doop{\mathrm{blue}}\) & 96 & 0 & 96 \\
Obs + \(\doop{\mathrm{green}}+\doop{\mathrm{blue}}\) & 64 & 64 & 64 \\
\bottomrule
\end{tabular}
\end{table}

\paragraph{Comparison methods.}
We use the pretrained ComplexMech CIDER-FM and CIDER-FM-Obs from
\cref{sec:general-random-graphs}, keeping their weights fixed.
All \texttt{obs\_total} methods receive the same 192 observational rows,
including the 96 used by CIDER-FM's fusion context.
The \texttt{obs\_fixed} comparison applies the same CIDER-FM-Obs checkpoint
to those 96 rows. We also evaluate the fusion checkpoint with the
192-row observational input.
TabPFN uses the local v2 checkpoint and its full predictive distribution
for NLL. Graph4CFM receives from no ancestry information,
\(\mathrm{red}\to Y\), to
\(\mathrm{red},\mathrm{green},\mathrm{blue}\to Y\),
with all other off-diagonal relations left unknown.
Baseline and metric definitions follow \cref{app:baselines-metrics}.

\paragraph{Evaluation and aggregation.}
We divide the real-data confirmation partition into nine disjoint blocks.
Each block provides 128 query rows and three context draws sampled without
replacement within each draw. The query rows are fixed across methods
and repeats, giving 27 episodes with 1,152 distinct query outcomes.
We evaluate \(Y_q\) in the original sensor units and first average scores
over context repeats within each block.
NLL, MSE, and mean \(R^2\) give equal weight to the nine blocks; their
standard errors are the sample standard deviation of block scores divided
by \(\sqrt{9}\).
Median \(R^2\) is taken across the repeat-averaged block scores, with a
block-bootstrap standard error. Pooled \(R^2\) uses repeat-averaged squared
errors and the variation of all distinct query outcomes around their pooled
mean; its standard error is estimated by a delete-one-block jackknife.

\section{Case study: surrogate experiments sharpen conditional interventional predictions.}
\label{app:cid_concentration}

This section provides the construction and quantitative evaluation behind
\cref{fig:case_converge}. We use two observationally equivalent SCMs whose
conditional interventional distributions differ, but which can be distinguished
using a surrogate intervention on $Z$. The models receive neither graph
information nor interventions on the target treatment $T$.

\subsection{Task construction}
\label{app:cid_concentration:construction}

\paragraph{Observational equivalence.}
Consider four observed variables $(Z,T,Y,X)$ and mutually independent
standard-normal exogenous noises. With $r=0.65$ and $s=0.9$, define
\begin{equation}
\begin{aligned}
\psi_{\rightarrow}:\quad
  &X=\varepsilon_X,\qquad Z=\varepsilon_Z,\\
  &T=rZ+\sqrt{1-r^2}\,\varepsilon_T,\qquad
   Y=sT+\sqrt{1-s^2}\,\varepsilon_Y,\\[3pt]
\psi_{\leftarrow}:\quad
  &X=\varepsilon_X,\qquad Y=\varepsilon_Y,\\
  &T=sY+\sqrt{1-s^2}\,\varepsilon_T,\qquad
   Z=rT+\sqrt{1-r^2}\,\varepsilon_Z.
\end{aligned}
\label{eq:cid-case-scms}
\end{equation}
The respective graphs are $Z\to T\to Y$ and $Y\to T\to Z$, with an
independent nuisance covariate $X$ and no hidden confounding. Both SCMs induce
the same zero-mean Gaussian observational distribution: $X$ is independent of
$(Z,T,Y)$, whose covariance matrix is
\begin{equation}
\Sigma_{ZTY}=
\begin{pmatrix}
1&r&rs\\
r&1&s\\
rs&s&1
\end{pmatrix}
=
\begin{pmatrix}
1&0.65&0.585\\
0.65&1&0.9\\
0.585&0.9&1
\end{pmatrix}.
\label{eq:cid-case-covariance}
\end{equation}
Consequently, no amount of observational data can distinguish these two SCMs
by their likelihood. We generate all empirical contexts from
$\psi_{\rightarrow}$; $\psi_{\leftarrow}$ explains the ambiguity rather than
serving as a second data-generating condition.

\paragraph{Fixed conditional query.}
Although our main experiments use row-specific randomised query interventions
$\doop{T=t_i}$, we fix the query here to
$p(y\mid\doop{T=2},Z=z_\star,X=0)$, so the conditioning vector
in the main-text notation is $\mathbf{x}_\star=(z_\star,0)$. We set
$z_\star=0.58335$, the root closest to zero of
\begin{equation}
\varphi(z_\star;0,1)=\varphi(z_\star;1.3,0.5775),
\label{eq:cid-case-neutral-query}
\end{equation}
where $\varphi(\cdot;\mu,v)$ denotes a Gaussian density with mean $\mu$ and
variance $v$. These are precisely the two densities of $Z$ under
$\doop{T=2}$. This choice makes the query covariates equally likely under
the two intervened SCMs, so conditioning on them does not by itself favour
one SCM. Nevertheless, their CIDs differ:
\begin{equation}
\begin{aligned}
q_\star(y)
  :=p_{\psi_{\rightarrow}}(y\mid\doop{T=2},z_\star,0)
  &=\mathcal{N}(y;1.8,0.19),\\
p_{\psi_{\leftarrow}}(y\mid\doop{T=2},z_\star,0)
  &=\mathcal{N}(y;0,1).
\end{aligned}
\label{eq:cid-case-targets}
\end{equation}
In the reverse SCM, intervening on $T$ removes $Y\to T$, leaving $Y$
independent of $Z$ and $X$. Under an illustrative equal prior over exactly
these two SCMs, observational data and the chosen query covariates therefore
leave an equal-weight mixture of these two CIDs.

\paragraph{Why a surrogate experiment helps.}
Under $\doop{Z=z}$, the two SCMs instead imply
\begin{equation}
\mathbb{E}_{\psi_{\rightarrow}}[Y\mid\doop{Z=z}]=rsz=0.585z,
\qquad
\mathbb{E}_{\psi_{\leftarrow}}[Y\mid\doop{Z=z}]=0.
\label{eq:cid-case-surrogate}
\end{equation}
Thus a randomised $Z$ experiment supplies information that further
observational samples cannot provide about this pair: The ambiguity here
comes from the unknown graph direction.

\subsection{Paired contexts and frozen models}
\label{app:cid_concentration:protocol}

For each replication we generate 256 observational rows from the forward SCM
and 64 independent surrogate-interventional rows. Each experimental row
sets $Z_i\sim\operatorname{Uniform}[-4,4]$, independently of the exogenous
noises, and then samples the remaining variables according to the SCM.
These rows form one $\doop{Z}$ regime with row-specific assignments. No row intervenes on $T$ or $Y$.

The primary comparison uses an \emph{extra-intervention} protocol: the same
256 observational rows are retained and the 64 experimental rows are added.
To distinguish source information from sample-count effects, we also
generate a nested 320-row observational context whose first 256 rows are
identical to the primary context. This yields an equal-total-budget
comparison between 320 observational rows and $256+64$ observational and
experimental rows.

We evaluate the following frozen models without training or fine-tuning:
\begin{itemize}[leftmargin=*,itemsep=2pt]
\item \textbf{CIDER-FM:} the same Linear--Gaussian checkpoint (training step
95,000, 128-bar output) receives 256 Obs, 320 Obs, or 256 Obs plus 64
$\doop{Z}$ rows. The observational controls are inference-only uses of the
same weights, not separately trained CIDER-FM-Obs.
\item \textbf{Graphs4CFM:} the released fully-conditioned checkpoint of
\citet{reuter2026use}, with a 1,000-bar output, receives either 256 or 320
observational rows. All graph-relation information is marked unknown.
\end{itemize}
The 256/320-row observational regimes exceed CIDER-FM's training allocation
of 64 rows per regime; its observational-only inputs also constitute
source-availability extrapolation. These runs therefore serve as
same-checkpoint source-input controls, not as separately trained,
in-distribution observational baselines.

We froze the SCM, query, weights and budgets before evaluating 16 new independent context/noise
seeds. All replications are retained. The first seed
was specified for \cref{fig:case_converge} before this confirmation; the
figure is not an average of densities across seeds or queries.

\subsection{Distributional scores and quantitative results}
\label{app:cid_concentration:results}

Let $\widehat q_d$ and $\widehat F_d$ denote the predictive density and CDF for context draw $d$, and let $F_\star$ be the CDF of $q_\star=\mathcal{N}(1.8,\sigma_\star^2)$, with $\sigma_\star^2=0.19$. The expected negative log-likelihood (NLL) and continuous ranked probability score (CRPS) under $q_\star$ are
\begin{equation}
\begin{aligned}
\operatorname{NLL}_d
  &=-\int_{\mathbb{R}}q_\star(y)\log\widehat q_d(y)\,\mathrm{d}y,\\
\operatorname{CRPS}_d
  &=\int_{\mathbb{R}}\bigl(\widehat F_d(u)-F_\star(u)\bigr)^2\,\mathrm{d}u
    +\frac{\sigma_\star}{\sqrt{\pi}}.
\end{aligned}
\label{eq:cid-case-scores}
\end{equation}
We also report the equal-tailed 90\% predictive-interval width and coverage:
\begin{equation}
\begin{gathered}
\ell_d=\widehat F_d^{-1}(0.05),\qquad u_d=\widehat F_d^{-1}(0.95),\\
W_d=u_d-\ell_d,\qquad C_d=F_\star(u_d)-F_\star(\ell_d).
\end{gathered}
\label{eq:cid-case-intervals}
\end{equation}
Here, $W_d$ measures concentration, while $C_d$ assesses coverage against the nominal level of 0.90.

All scores use the unsmoothed bar distributions in original outcome units, including their tails. We compute NLL analytically under the Gaussian truth and CRPS by numerical quadrature. \Cref{tab:cid-concentration} reports mean $\pm$ SE over 16 independent context draws, with $\operatorname{SE}=s_{\mathrm{draw}}/\sqrt{16}$ and $s_{\mathrm{draw}}$ the sample standard deviation. In this case, surrogate data improve NLL and CRPS and narrow predictive intervals while maintaining near-nominal coverage. The improvement exceeds that obtained from the same number of additional observational samples.

\begin{table}[t]
\centering
\caption{Fixed-query case study (mean $\pm$ SE over 16 context draws).
$O$ denotes observational rows and $I_Z$ randomised $\doop{Z}$ rows.
All CIDER-FM rows use the same checkpoint. Nominal coverage is 0.90;
the true CID provides an analytic reference.}
\label{tab:cid-concentration}
\begingroup
\footnotesize
\setlength{\tabcolsep}{3pt}
\renewcommand{\arraystretch}{1.15}
\begin{tabular}{@{}llrrrr@{}}
\toprule
Model & Context & NLL $\downarrow$ & CRPS $\downarrow$ & 90\% width & Coverage\\
\midrule
Graphs4CFM & $256O$ & $1.5301\pm .0619$ & $.5864\pm .0287$ & $3.0828\pm .0702$ & $.7903\pm .0298$\\
Graphs4CFM & $320O$ & $1.3926\pm .0500$ & $.5622\pm .0245$ & $3.1862\pm .0596$ & $.8539\pm .0223$\\
CIDER-FM & $256O$ & $1.0544\pm .0186$ & $.4404\pm .0113$ & $3.3288\pm .0233$ & $.9260\pm .0058$\\
CIDER-FM & $320O$ & $1.0383\pm .0142$ & $.4313\pm .0085$ & $3.3268\pm .0256$ & $.9287\pm .0046$\\
CIDER-FM & $256O+64I_Z$ & $\mathbf{.7852}\pm .0124$ & $\mathbf{.2977}\pm .0055$ & $2.5854\pm .0467$ & $.9277\pm .0049$\\
\midrule
True CID & --- & $.5886$ & $.2459$ & $1.4340$ & $.9000$\\
\bottomrule
\end{tabular}
\endgroup
\end{table}

\subsection{Visualisation and scope}
\label{app:cid_concentration:visualisation}

For \cref{fig:case_converge}, we smooth each model's bar distribution using a Gaussian KDE fitted to $n=16{,}384$ stratified inverse-CDF samples, including the tails. All models use Scott's bandwidth rule, $h=s_Y n^{-1/5}$, where $s_Y$ is the sample standard deviation of the predictive draws. The black dashed curve shows the analytic true CID. KDE smoothing is used only for visualisation. This case illustrates how surrogate experiments can sharpen conditional interventional predictions for a fixed SCM and query, with the results replicated across independent context draws.

\end{document}